%% file: main.tex
\documentclass[twocolumn]{IEEEtran}

\usepackage[pdftex]{graphicx} % 插图命令 \includegraphics
\usepackage{amsmath} % Math Packages
\usepackage{algorithmic}
\usepackage{algorithm}
\usepackage{array} % Alignment Packages
\usepackage{subfigure} % Subfigure Packages
\usepackage{dblfloatfix}
\usepackage{url} % simple URL typesetting

\usepackage{microtype}
\usepackage[utf8]{inputenc} % allow utf-8 input
\usepackage[T1]{fontenc}    % use 8-bit T1 fonts
\usepackage{paralist} % Compacted versions of enumerate and itemize.
\usepackage{marvosym} % 常用符号, 特别是打勾、打叉
\usepackage{bm} % 实现字体的斜体加粗
\usepackage{bbm} % fix \mathbb doesn't support digits
\usepackage{amssymb}   % 黑板粗体 \mathbb, 德文尖角体 \mathfrak, \backsim
\usepackage{mathrsfs} % 花体 \mathscr
\usepackage{tabularray} % tblr or talltblr or longtblr environment
\UseTblrLibrary{booktabs}       % professional-quality tables，也就是三线表
\usepackage{tabularx} % 自动计算列宽的表格包
\usepackage[flushleft]{threeparttable} % 给表格项添加注释
\usepackage{makecell} % 表格内容换行
\usepackage{multirow} % 多行表格
\usepackage{nicefrac} % 斜着的分号
\usepackage[dvipsnames]{xcolor} % 字体颜色
\usepackage{xpatch}
\usepackage{colortbl} % \rowcolor
\usepackage{soul} % highlight, strikeout, underline
\setul{0.5ex}{0.3ex}

\soulregister\cite7
\soulregister\ref7 
\soulregister\citep7 % 针对\citep命令
\soulregister\citet7 % 针对\citet命令
\soulregister\pageref7 % 针对\pageref命令

\definecolor{citecolor}{RGB}{34,139,34}
\usepackage[pagebackref=true,breaklinks=true,letterpaper=true,colorlinks,
    citecolor=citecolor,bookmarks=false]{hyperref}
\usepackage{cleveref}  % for \cref
\crefformat{figure}{Fig.~#2#1#3}
\crefformat{equation}{Eq.~(#2#1#3)}
\crefformat{table}{Table~#2#1#3}
\crefformat{section}{Section~#2#1#3}
\crefformat{algorithm}{Algorithm~#2#1#3}
\crefformat{appendix}{Appendix #2#1#3}
\crefformat{subsection}{subsection~#2#1#3}
\usepackage{cellspace}
\usepackage[numbers,sort&compress]{natbib} % 代替 cite.sty

\usepackage{xspace} % 用于在 \newcommand 中添加空格

\usepackage{pifont} % http://ctan.org/pkg/pifont

\begin{document}

\setlength{\abovedisplayskip}{.5\baselineskip} % 调整公式与正文间的段前距离
\setlength{\belowdisplayskip}{.5\baselineskip} % 调整公式与正文间的段后距离

% \title{Sparsity Is Not All You Need: Directional Selectivity Model for Infrared Small Target Detection}
\title{Sparse Prior Is Not All You Need: When Differential Directionality Meets Saliency Coherence for \\Infrared Small Target Detection}

\input{contents/authors.tex}

\maketitle

\input{contents/abstract.tex}
\input{contents/introduction.tex}
\input{contents/related-work.tex}
\input{contents/method.tex}

\input{contents/experiment.tex}

\input{contents/conclusion.tex}

\bibliographystyle{IEEEtran}
\bibliography{./reference.bib}

\input{contents/biography.tex} % 作者介绍

\end{document}

%% file: contents/authors.tex
% !TEX root = ../main.tex

\author{
  Fei~Zhou,
  Maixia~Fu,
  Yulei~Qian,
  Jian~Yang,
  Yimian~Dai
  \thanks{
    This work was supported by
      % 青基
      the Young Scientists Fund of the National Natural Science Foundation of China (62303165,      % 周飞
        % 我
        62301261,
        % 庞栋栋
        62301036,
        62361166670),
    % 博后面上
    China Postdoctoral Science Foundation (No. 
    2021M701727, % 我
    2022M720446 % 庞栋栋
    ),
    % 周飞 - 河南省科技攻关
    Key R\&D and Promotion Special Projects (Science and Technology Research, Soft Science) in Henan Province (No. 232102220032);
    Open Project of Key Laboratory of Grain Information Processing and Control (Henan University of Technology), Ministry of Education (KFJJ2023014);
    % 付老师 - 河南工业大学创新基金支持计划专项资助
    the Innovative Funds Plan of Henan University of Technology (No. 2022ZKCJ15),
    Research Foundation for University Key Teacher of Henan Province (No.2020GGJS084).
    \emph{(Corresponding author: Yimian Dai)}
    }

  % 河工大
  \thanks{Fei Zhou and Maixia Fu are with Key Laboratory of Grain Information Processing and Control (Henan University of Technology), Ministry of Education; Henan Key Laboratory of Grain Photoelectric Detection and Control, Henan University of Technology, Zhengzhou, China.
  (e-mail: \href{mailto:hellozf1990@163.com,fumaixia@126.com}{hellozf1990@163.com,fumaixia@126.com}).
  }

  % 724
  \thanks{Yulei Qian is with Nanjing Marine Radar Institute, Nanjing, China
  (e-mail: \href{mailto:yuleifly@126.com}{yuleifly@126.com}).
  }

  % % 广州第二师范
  % \thanks{Feifei Zhang is with School of Computer Science, Guangdong University of Education, Guangzhou 510303, China
  % (e-mail: \href{mailto:zhangfeifei2006@126.com}{zhangfeifei2006@126.com}).
  % }

  % % 光机所
  % \thanks{
  %   Yaohong Chen is with Xi'an Institute of Optics and Precision Mechanics, Chinese Academy of Sciences, Xi'an, China.
  %   (e-mail: \href{mailto:chenyaohong@opt.ac.cn}{chenyaohong@opt.ac.cn}).
  % }

  % 南理工
  \thanks{
    % Huan Wang, 
    Jian Yang and Yimian Dai are with PCA Lab, Key Lab of Intelligent Perception and Systems for High-Dimensional Information of Ministry of Education, and Jiangsu Key Lab of Image and Video Understanding for Social Security, School of Computer Science and Engineering, Nanjing University of Science and Technology, Nanjing, China.
    (e-mail:
    % \href{wanghuanphd@njust.edu.cn}{wanghuanphd@njust.edu.cn};
    \href{mailto:csjyang@mail.njust.edu.cn}{csjyang@mail.njust.edu.cn};     \href{mailto:yimian.dai@gmail.com}{yimian.dai@gmail.com}).
  }

  % 南开
  % \thanks{Xiang Li is with IMPlus@PCA Lab, College of Computer Science, Nankai University, Tianjin, China. (e-mail:\href{mailto:xiang.li.implus@nankai.edu.cn}{xiang.li.implus@nankai.edu.cn}).
  % }

  % 南邮
% \thanks{
%     Minrui Zou and Kang Ni are with Nanjing University of Posts and Telecommunications, Nanjing, China. Kang Ni is also with Key Laboratory of Radar Imaging and Microwave Photonics, Nanjing University of Aeronautics and Astronautics, Ministry of Education, Nanjing, China.
%     (e-mail:
%     \href{mailto:}{};
%     \href{mailto:tznikang@163.com}{tznikang@163.com}).
%   }

}

%% file: contents/abstract.tex
% !TEX root = ../main.tex

\begin{abstract}
Infrared small target detection is crucial for the efficacy of infrared search and tracking systems.
% The prevalent tensor decomposition methods are intensely focused on representing small targets constrained by sparsity. Nonetheless, sparsity alone falls short in effectively segregating the target from \hl{the complex background}, which is principally due to two factors: (1) the insufficient exploitation of intrinsic directional information present in infrared backgrounds, and (2) the significant reduction in target visibility during the decomposition phase. 
Current tensor decomposition methods emphasize representing small targets with sparsity but struggle to separate targets from complex backgrounds due to insufficient use of intrinsic directional information and reduced target visibility during decomposition.
% In addressing these limitations, our study introduces the concept of \emph{Sparse Differential Directionality prior (SDD)}, a framework that discerns small targets from background interference by leveraging their distinct directional characteristics. This approach utilizes mixed sparse constraints on the differential directional images of the spatial component and the continuity difference matrix of the temporal component, both extrapolated from Tucker decomposition. 
To address these challenges, this study introduces a \emph{Sparse Differential Directionality prior (SDD)} framework. SDD leverages the distinct directional characteristics of targets to differentiate them from the background, applying mixed sparse constraints on the differential directional images and continuity difference matrix of the temporal component, both derived from Tucker decomposition.
% Additionally, our method incorporates a saliency coherence enhancement strategy, which intensifies the detectability of small targets against the backdrop of hierarchical decomposition, thereby escalating the contrast with background interference. 
We further enhance target detectability with a saliency coherence strategy that intensifies target contrast against the background during hierarchical decomposition. 
% The proposed model is adeptly resolved using a Proximal Alternating Minimization-based (PAM) algorithm. 
A Proximal Alternating Minimization-based (PAM) algorithm efficiently solves our proposed model. 
% Extensive experiments on a variety of real-world datasets affirm the supremacy of our approach, surpassing ten state-of-the-art methods in terms of target detection and clutter suppression. 
Experimental results on several real-world datasets validate our method's effectiveness, outperforming ten state-of-the-art methods in target detection and clutter suppression.
Our code is available at \url{https://github.com/GrokCV/SDD}.
\end{abstract}

% \vspace{-1\baselineskip}
% \vspace{2\baselineskip}

%% file: contents/introduction.tex
% !TEX root = ../main.tex
% \bibliography{../reference.bib}
\section{Introduction} \label{sec:introduction}
% 1. 研究背景的意义
% 2. 国内外发展动态分析
% 3. 关键技术瓶颈/关键科学问题
% 4. 研究目标/文章动机
% 5. 技术路线
% 6. 本研究的意义
% 7. 章节安排
% \IEEEPARstart{I}{nfrared} small target detection is a pivotal task for the early identification and localization of objects of interest \cite{Deng2019A}, thereby greatly enhancing the effectiveness of subsequent search and tracking operations.
% This technique boasts an extensive array of \hl{practical applications}, including but not limited to, early warning systems \cite{bai2018derivative}, anti-missile defense mechanisms \cite{deng2016small}, marine intelligent defense \cite{Deng2024FMR}, and automated defect detection \cite{Ruben2018Automated}, among various other critical domains.
\IEEEPARstart{I}{nfrared} small target detection is crucial for the early identification and localization of objects of interest \cite{Deng2019A}, enhancing the effectiveness of subsequent search and tracking operations.
This technique has a wide range of practical applications, including early warning systems \cite{bai2018derivative}, anti-missile defense mechanisms \cite{deng2016small}, marine intelligent defense \cite{Deng2024FMR}, and automated defect detection \cite{Ruben2018Automated}, among others.

\begin{figure*}[!h]
\begin{minipage}[b]{0.10\linewidth}
\centering
	\subfigure[]{\label{fig:1(a)}
	\includegraphics[height=2.9in]{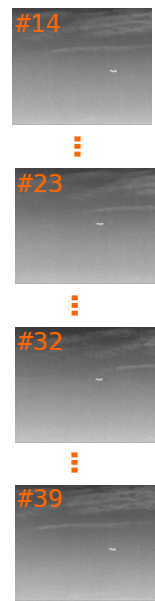}}
\end{minipage}
\begin{minipage}[b]{0.45\linewidth}
\centering
	\subfigure[]{\label{fig:1(b)}
	\includegraphics[width=2.8in]{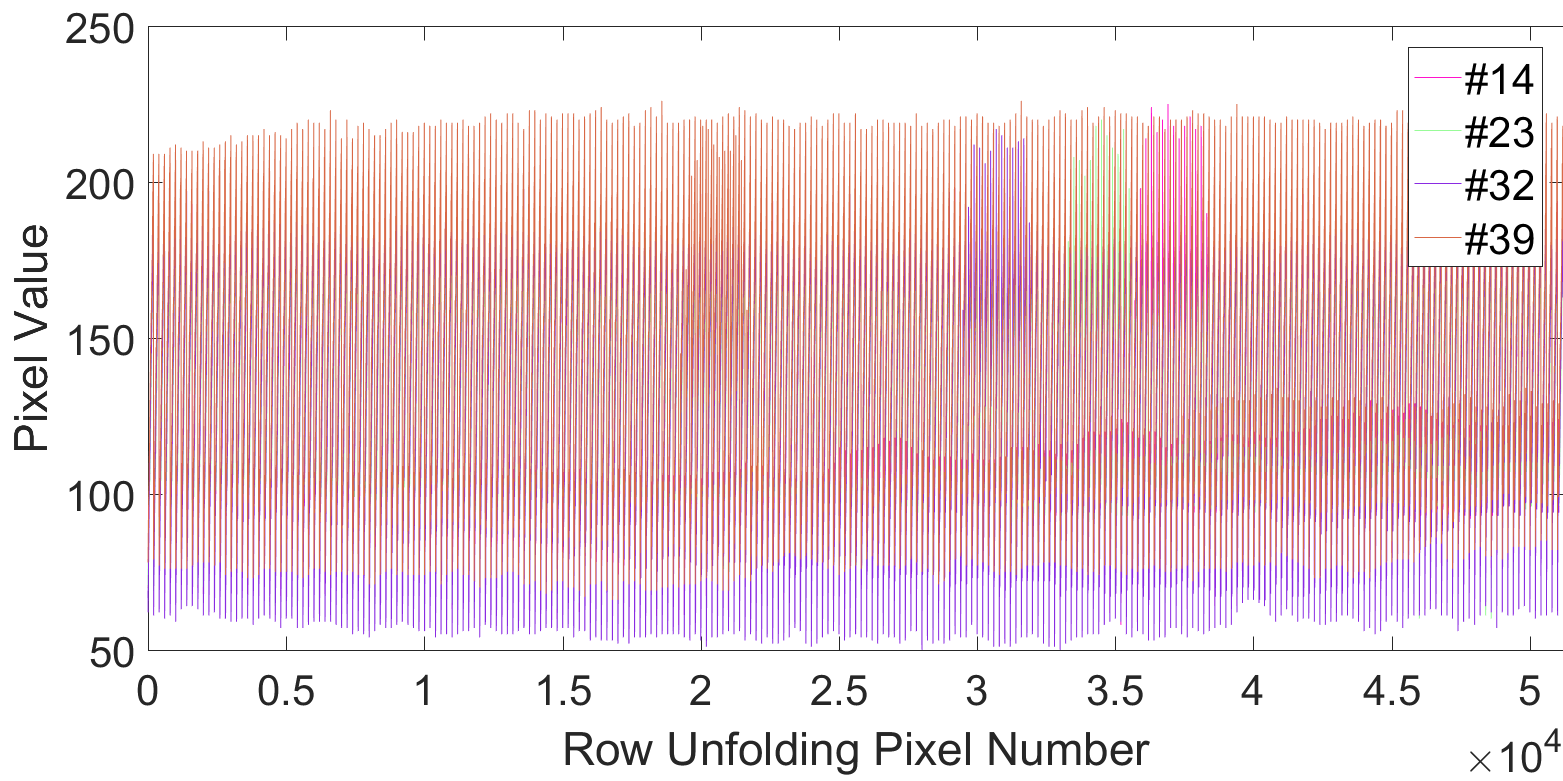}} \\
	\subfigure[]{\label{fig:1(c)}
	\includegraphics[width=2.8in]{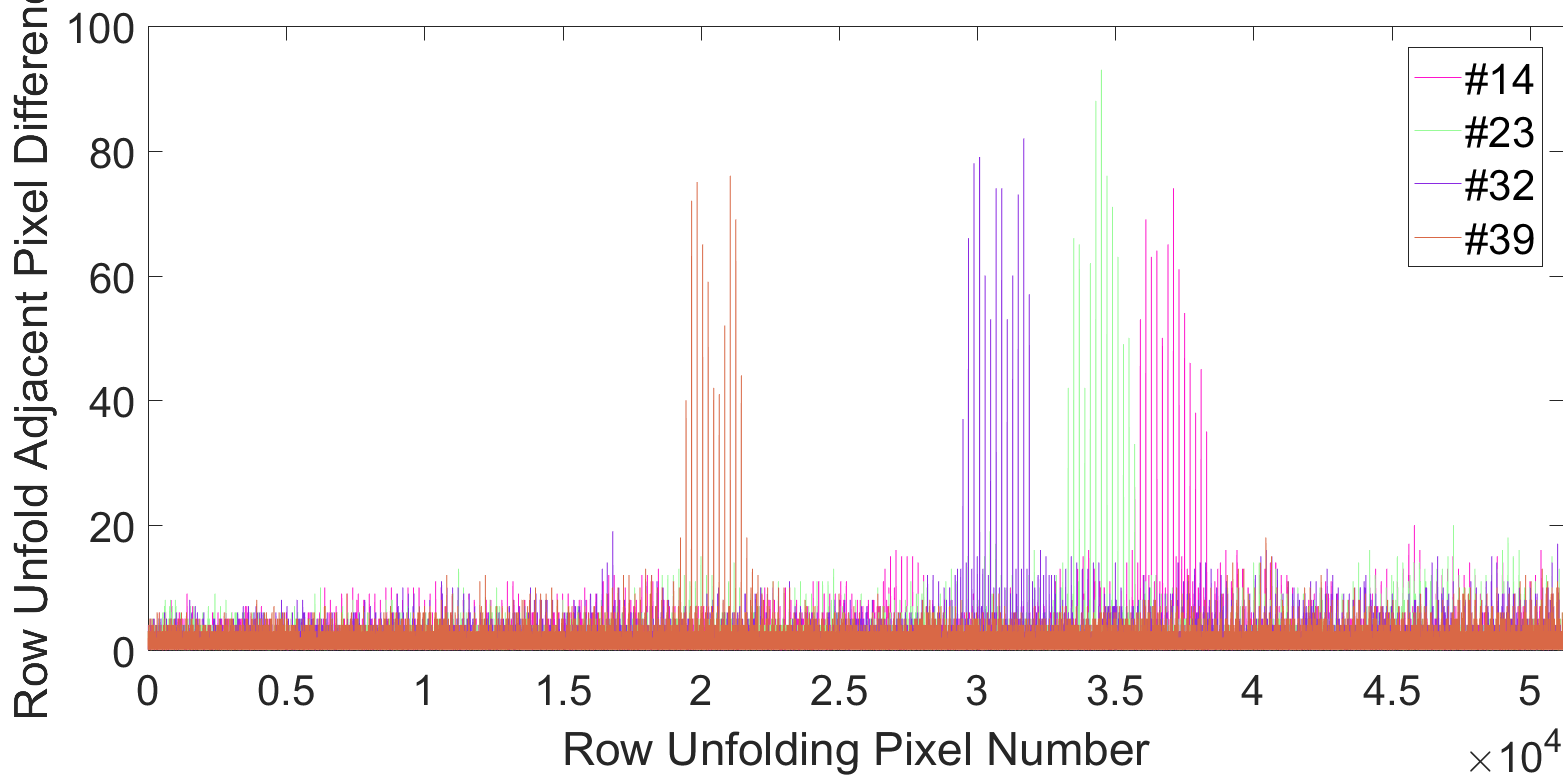}} \\% 换行
\end{minipage} % 中间不空行代表不换行
\begin{minipage}[b]{0.10\linewidth}
\centering
	\subfigure[]{\label{fig:1(e)}
	\includegraphics[width=2.8in]{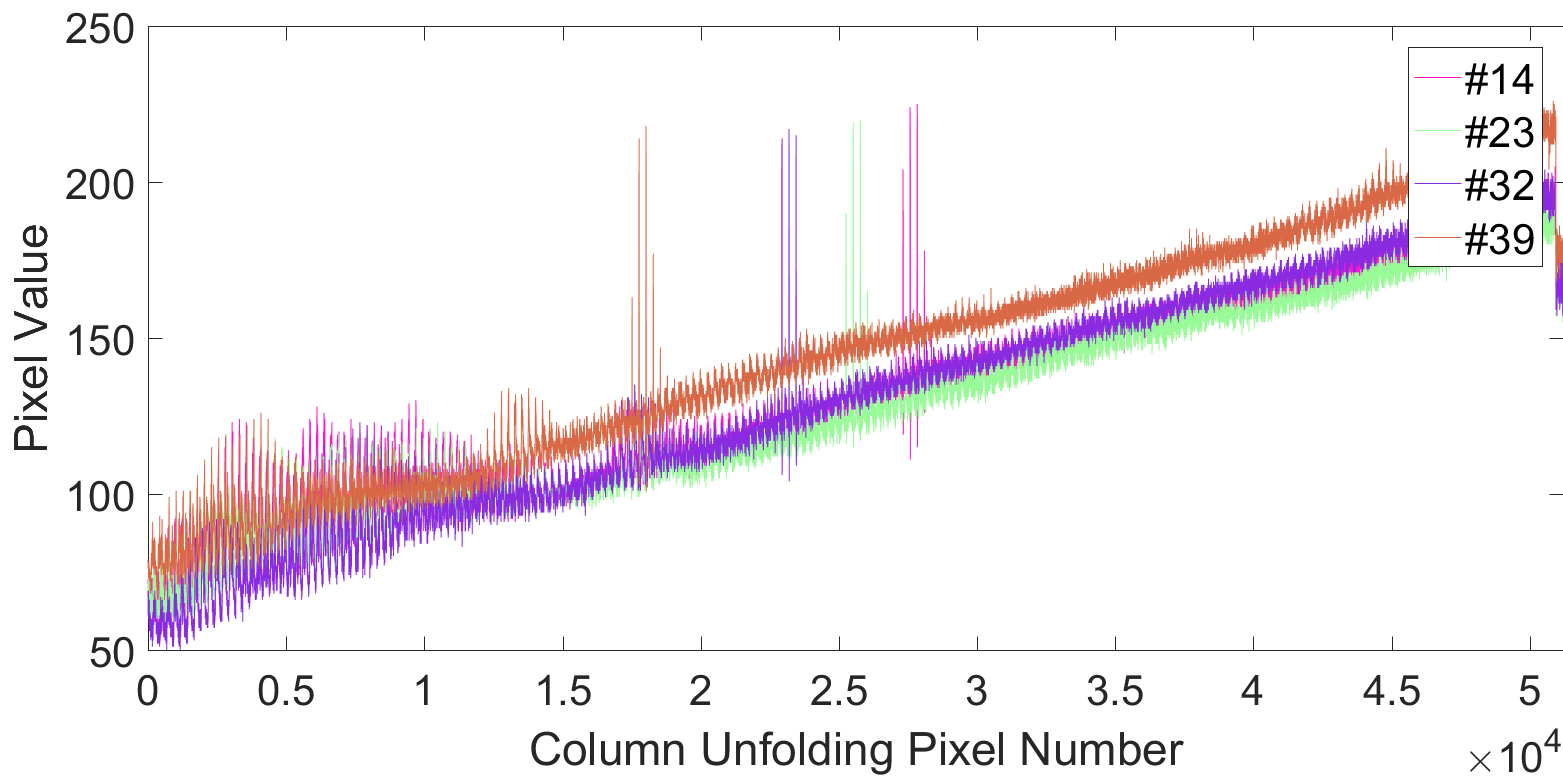}}\\
 \subfigure[]{\label{fig:1(e)}
	\includegraphics[width=2.8in]{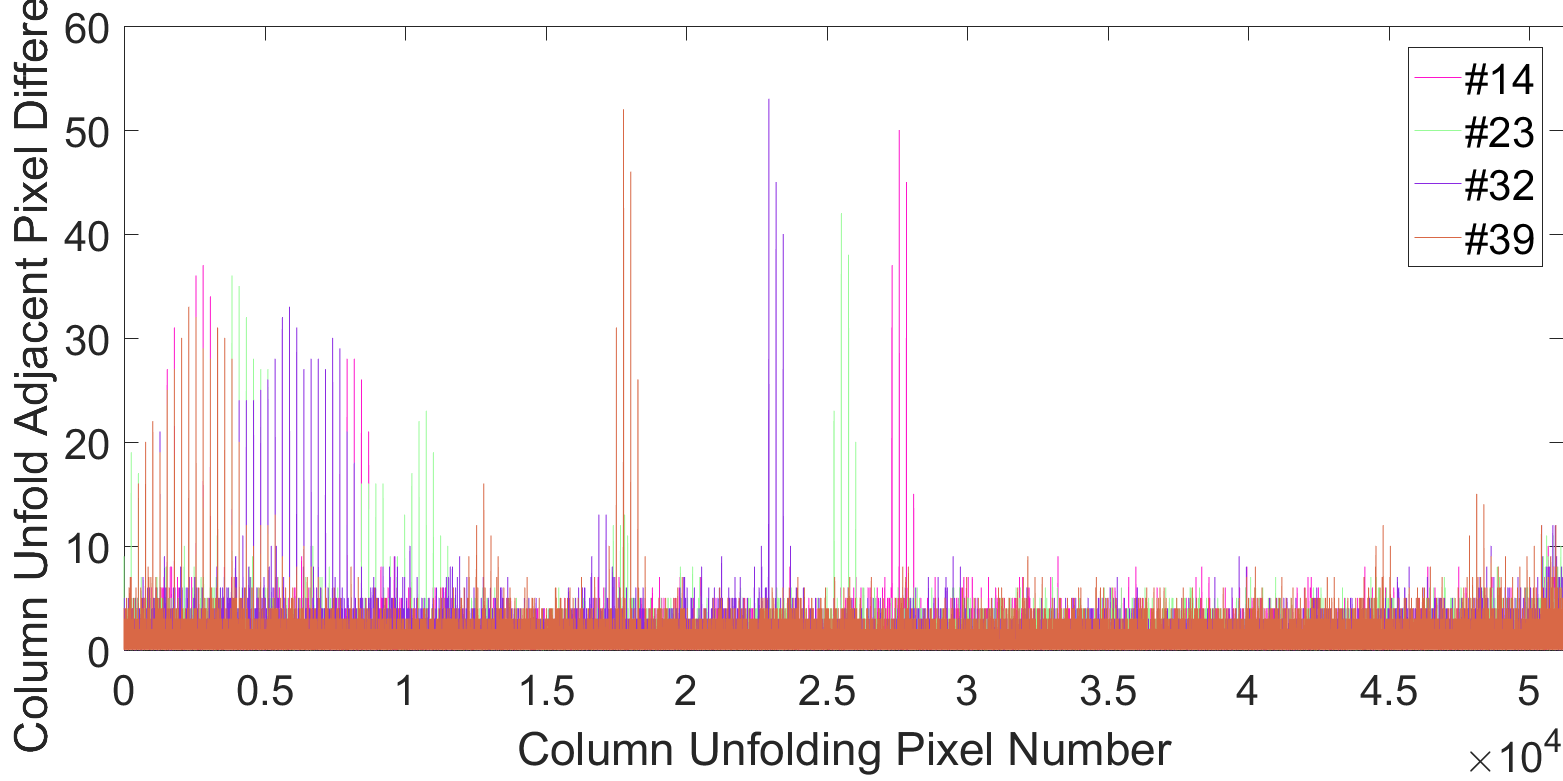}}\\
\end{minipage}
\caption{Illustration of directional information of background and target in infrared sequences. (a) Infrared Sequence. (b) Unfolding the four scenes in the sequence horizontally. (c) Difference between adjacent pixels during horizontal unfolding. (d) Unfolding the four scenes in the sequence vertically. (e) Difference between adjacent pixels during vertical unfolding.}
\label{Directional information}
\end{figure*}

\subsection{Prior Work on Infrared Small Target Detection}

Despite the remarkable advancements in infrared small target detection, two significant challenges persist:
\begin{enumerate}
    \item \textbf{Limited pixel representation}: Objects of interest often occupy only a few pixels in the scene, a consequence of the requirement for long-distance perception. The limited number of pixels fails to convey the intrinsic shape and texture characteristics needed for target identification. This paucity of information poses a significant challenge in accurately detecting and recognizing small targets.

    \item \textbf{Salient background interferences}: Infrared scenes frequently contain salient background elements, such as thick clouds or sea clutters, which can obscure small targets. These interferences weaken the correlation between backgrounds, making it difficult to distinguish targets from their surroundings. Effectively suppressing background interferences while preserving target information remains a critical challenge in infrared small target detection.
\end{enumerate}
These challenges continue to be the key factors impeding progress in downstream tasks and require solutions to enhance the performance of infrared small target detection.

In recent years, deep learning methods have made significant strides in small target detection \cite{Zhang2022ISNet}, with notable works such as YOLO series \cite{yang2024yolov8,huang2023improved}.
The advent of these methodologies have eclipsed the detection performance of traditional knowledge-driven approaches \cite{Dai2021Asymmetric,wang2019miss,Zhang2022ISNet,Fang2022Infrared}.
However, these data-driven approaches are constrained by limitations such as dataset scarcity, difficulty in representing intrinsic patterns, and poor interpretability, leading to a bottleneck in performance enhancement. 
To address these issues, researchers have explored deep unfolding methods that incorporate domain knowledge into data-driven approaches, leveraging the advantages of feature representation inherent in both domain knowledge-driven and data-driven methods \cite{dai2021attentional,wu2024rpcanet,Dai2023One,Yang2022DFFIR}, for example RPCANet \cite{wu2024rpcanet}, ALCNet \cite{dai2021attentional}.
% For instance, RPCANet \cite{wu2024rpcanet} combines the strengths of traditional RPCA and deep learning for infrared small target detection. Nevertheless, these methods often rely on basic prior characterizations and may overlook valuable domain-specific knowledge that could further enhance their performance.
From this, we observe that precise delineation of domain-specific knowledge can significantly contribute to the augmentation of data-driven models. Therefore, this paper endeavors to further explore the development of models underpinned by such specialized domain knowledge.

Over the past decades, numerous domain knowledge-driven detection methods have been proposed. These techniques focus on extracting small targets from infrared backgrounds by modeling target priors, such as sparsity \cite{Cao2013Infrared}, saliency \cite{Chen2014A,Wei2016Multiscale}, or constructing background knowledge \cite{Zhang2024A,Han2016An,Dai2017Reweighted,Wang2017Infrareddim,Pan2020Infrared,Li2016A}. Despite their effectiveness, early low-rank methods struggled to distinguish targets from background interference \cite{Cao2013Infrared,guo2017small,IPT17NIPPS} due to the presence of sparse distractors in the background. Consequently, improvement schemes have been proposed, with sequential detection offering an attractive alternative due to its ability to integrate inter-frame priors.

% In sequence detection, based on the inter-frame correlation and intra-frame non-local similarity of the sequence background, some methods enhance the background's low-rank nature by designing different feature structures, such as spatio-temporal patch images \cite{Pan2020Infrared,Gao2017Infrared}, tensor cube models \cite{Zhu2020Infrared,zhou2021infrared,Sun2020Infrared,Liu2020Small}, and connected multi-frame patch groups \cite{zhou2023infrared}. 
% Small targets can then be extracted by applying different tensor constraints \cite{deng2022generalized,Li2023Sparse} to decompose the low-rank component hidden in the infrared sequences and the sparse components representing small targets.
% Furthermore, researchers have incorporated additional domain expert priors, such as tophat regularization \cite{Zhu2020Infrared}, total variation \cite{Wang2017Infrared,Dan2024Dynamic}, and its variants \cite{Zhang2020Edge,liu2022Nonconvex}, to explore the spatial local correlation of the background or the local saliency of small targets.
% However, these methods are not only complex to implement and computationally expensive, but they also lack certain stability in dynamic scenes.
% \hl{It is because that in dynamic scenes, the inter-frame correlation and the non-local auto-correlation of the background often decrease over time with changes in the background.}

In sequence detection, methods enhance the low-rank nature of the background using different feature structures, such as spatio-temporal patch images \cite{Pan2020Infrared,Gao2017Infrared}, tensor cube models \cite{Zhu2020Infrared,zhou2021infrared,Sun2020Infrared,Liu2020Small}, and connected multi-frame patch groups \cite{zhou2023infrared}. These methods apply tensor constraints \cite{deng2022generalized,Li2023Sparse} to decompose the low-rank and sparse components representing small targets. Additionally, researchers have incorporated domain expert priors, such as tophat regularization \cite{Zhu2020Infrared}, total variation \cite{Wang2017Infrared,Dan2024Dynamic}, and its variants \cite{Zhang2020Edge,liu2022Nonconvex}. However, these methods are complex, computationally expensive, and lack stability in dynamic scenes due to the decrease in inter-frame correlation and non-local auto-correlation of the background over time.

From the above analysis, infrared small target detection methods are broadly classified into two groups: background characterization and target characterization methods.
Background characterization methods, such as low-rank sparse modeling \cite{Wu2023Infrared,Liu2023Single}, filtering \cite{Han2016An}, and transform-domain approaches \cite{Lu2021An}, leverage the intrinsic geometric structures and mathematical statistical properties of infrared images for target-background separation. While effective in suppressing globally repetitive sparse components, these methods struggle with rare sparse components in dynamic scenes due to the similar sparsity between disturbances and targets, and the weakened interframe correlation from scene changes.

% On the other hand, target characterization methods, such as local contrast \cite{Chen2014A,Xu2023Infrared}, gradient vector fields \cite{Liu2018Flux,liu2023infr}, random walker \cite{Qin2019Infrared}, and other methods, characterize target local saliency through the difference information of horizontal, vertical, or diagonal pixels in the local area. These methods have the disadvantage of being unable to fully exploit the intrinsic structural information of the background, making it difficult to extract dim targets and suppress repetitive sparse regions. This is due to the fact that high-brightness regions within the background can exhibit greater prominence relative to the targets.

Target characterization methods, for instance, local contrast \cite{Chen2014A,Xu2023Infrared}, gradient vector fields \cite{Liu2018Flux,liu2023infr}, and random walker \cite{Qin2019Infrared}, exploit the difference information of local pixels to highlight target saliency. However, their inability to fully utilize the background's structural information limits their effectiveness in extracting dim targets and suppressing repetitive sparse regions, especially when high-brightness background areas exhibit greater prominence than targets.

% In summary, existing methods consider either global structural priors or local directional differential priors unilaterally, resulting in performance limitations. Background characterization methods excel at suppressing globally repetitive sparse components but struggle with rare sparse components in dynamic scenes. Conversely, target characterization methods face difficulties in extracting dim targets and suppressing repetitive sparse regions due to their inability to fully utilize the background's intrinsic structural information. To address these limitations, a more effective approach should leverage the complementary strengths of both background and target characterization methods, enabling better handling of complex infrared scenes containing both rare and repetitive sparse components.

In summary, existing methods primarily consider either global structural priors or local directional differential priors, leading to performance limitations. A more efficient approach should merge the complementary strengths of both methods to better manage complex infrared scenes with both rare and repetitive sparse components.

\subsection{Motivation}

% \hl{It is evident that the primary issue in the aforementioned algorithms lies in the ineffective distinction between targets and background interference.}
% In the case of low-rank and sparse decomposition methods, the sparse disturbances present in the background manifest a sparsity akin to that of the targets themselves.
% Similarly, for local saliency methods, high-brightness regions within the background can exhibit greater prominence relative to the targets.
% \hl{However, beyond the incorporation of prior information by the aforementioned methods, we have identified an additional latent prior -- sparse targets exhibit non-directionality, while background interference possesses inherent directionality.
% Inspired by the local saliency method, the application of differential methods across various axes in infrared imagery can more precisely delineate the directionality inherent in the background as opposed to the nondirectionality of targets.
% This approach facilitates a more effective segregation of targets from background interference.}
In this study, we have identified a latent prior that augments the effectiveness of low-rank plus sparse decomposition methods: sparse targets exhibit non-directionality, while background interference possesses inherent directionality.
Background directionality pertains to the phenomenon wherein components of clutter and interference in the background exhibit varying sparse representations across different directions. In contrast, the non-directionality of the target implies that the target components maintain consistent sparse representations regardless of direction.
Our approach utilizes the contrast between the non-directionality of targets and the directionality of backgrounds. By applying differential methods across various axes in infrared imagery, we accurately delineate the inherent directionality of the background compared to the non-directionality of targets, enabling effective target-background segregation.
Different from previous works, our study is decicated to incorporate the concept of directionality as a prior to distinguish targets from the background in the low-rank sparse decomposition model.
As depicted in Fig. \ref{Directional information}, small targets maintain a significant sparse distribution, regardless of the unfolding direction. Conversely, the background displays notable differences across different directions. The differential information between adjacent pixels in both unfolding directions further accentuates the anisotropy of the background and the isotropy of the targets, as shown in Fig. \ref{Directional information} (c) and (e).

To validate the non-directionality of the target and the directionality of background clutter interference, we applied a multi-scale local contrast strategy \cite{Wei2016Multiscale} on the public dataset SIRST \cite{dai2021attentional}. A statistical analysis of the differential variance information in the horizontal, vertical, and two diagonal directions for the selected homogeneous background region, clutter edge region, and target region was conducted. Table \ref{tab:direc} presents the differential variances of different regions in various directions, with smaller variance values indicating weaker directionality. The statistical results show that 95.6\% of the dataset exhibits differential variances greater than 20 in the background clutter edge, 96.3\% of the dataset shows target differential variances between 5 and 20, and 97.2\% of the dataset has differential variances between 0 and 10 in homogeneous background regions.

\begin{table}
\tabcolsep=0.33cm
  \caption{Directional verification 
  % of infrared image components 
  on the SIRST dataset}
  \vspace{-1.0em}
 \begin{tabular} {cccc}% {\textwidth}{c@{\extracolsep{\fill}}cccc}
  \toprule
  % \rowcolor{yellow}
   Indicators & \makecell{Homogeneous\\ Region} &  \makecell{Small\\ Target Area} &  \makecell{Clutter\\ Edge} \\
  \midrule  
  % \rowcolor{yellow}
  Variance range & $0 \sim 10$ & $5 \sim 20$ &  $\ge {\rm{20}}$\\
  % \midrule
  % \rowcolor{yellow}
  Variance percentage & 97.2\%& 96.3\% & 95.6\% \\
  % \midrul
  \bottomrule
  \end{tabular}
  \label{tab:direc}
  \vspace{-1.0em}
  \end{table}
  
In summary, the main contributions can be summarized as follows:
\begin{enumerate}
  \item We introduce the SDD prior, which distinguishes sparse small targets from sparse background interference based on their differential directionality.
  % We observe that the sparse small target and sparse background interference can be distinguished via differential directionality, which is termed as \textbf{sparse differential directionality prior} (SDD).
  % We propose an infrared small target detection method that combines directional differential priors with mixed sparsity regularization, taking into account the latent directional prior in addition to the single sparsity.
  \item We propose an infrared small target detection method that incorporates the SDD prior into sparse and low-rank decomposition with mixed sparsity regularization, effectively exploiting the differential directionality information.
  % Driven by this prior, we propose an infrared small target detection method that incorporates the differential directionality into sparse and low-rank decomposition with mixed sparsity regularization.
  \item We design a saliency coherent map to mitigate the issues of brightness attenuation in small targets and weak saliency coherence among background clutters.
  % We design a saliency coherent map with a local enhancement function based on structure tensor indicators. This approach effectively mitigates the issues of brightness attenuation in small targets and weak saliency coherence among background clutters.
  \item We develop an efficient PAM algorithm to solve the proposed model, which has improved detection efficiency by at least 50\% compared to similar algorithms  \cite{Pang2022STTM,Zhu2020Infrared,Zhang2020Edge,liu2022Nonconvex}.
  % \hl{We develop an efficient PAM algorithm to solve the proposed model.}
\end{enumerate}

% The model's effectiveness is evaluated through a series of experiments, demonstrating its superiority compared to other state-of-the-art models in terms of target extraction and clutter suppression.

% The remainder of this paper is organized as follows. Section \uppercase\expandafter{\romannumeral2} introduces the tensor notations used in this paper and briefly reviews relevant works. Section \uppercase\expandafter{\romannumeral3} presents the construction of mixed sparse regularization with directional differential information and the saliency coherence map. Section \uppercase\expandafter{\romannumeral4} provides the solution for the proposed model. Section \uppercase\expandafter{\romannumeral5} reports experiments on extensive real datasets to demonstrate the superiority of the proposed model. The conclusion is presented in the final section.

%% file: contents/related-work.tex
% !TEX root = ../main.tex
% \bibliography{../reference.bib}

\section{Related Work} \label{sec:related}
\subsection{ Notations and Preliminaries}
In this section, we introduce several notations related to the proposed method, as listed in Table \ref{tab:notion}. Additionally, we provide definitions for the mode-$n$ tensor-matrix product, the Tucker decomposition, and the Tucker rank. For a more comprehensive understanding of tensors, we recommend consulting \cite{Kolda2009Tensor}.
\begin{table}[htb!]
  \tabcolsep=0.02cm
\caption{Summary of Mathematical Annotation}
\begin{tabular}{cc}
\toprule
Notation & Interpretation  \\
\midrule
$x$,$\bf x$, $X$, $\cal X$ & scalar, vector, matrix, $N$-dimensional tensor \\
${x_{{i_1}}}$,${x_{{i_2}}}$,...,${x_{{i_n}}}$ &the (${x_{{i_1}}}$,${x_{{i_2}}}$,...,${x_{{i_n}}}$) element of $\cal X $     \\
${x_{i:k}}$, ${x_{:jk}}$, ${x_{ij:}}$ & row, column and tube fibers of a 3-D tensor $\cal X $ \\
${X_{:j:}}$, ${X_{i::}}$, ${X_{::k}}$ & lateral, horizontal, and frontal slides of a 3-D tensor $\cal X $  \\
${X_{(n)}}  $ & \makecell{mode-n matricization of tensor  $\cal X$, obtained by \\arrangingthe mode-n fibers as the columns of the \\resulting matrix of size ${\mathbb{R}^{{I_n} \times \prod\nolimits_{k \ne n} {{I_k}} }}$ } \\
${\left\| {\cal X} \right\|_F} $ & \makecell{Frobenius norm of tensor $\cal X$,  \\defined as ${\left\| {\cal X} \right\|_F} = \sqrt {\sum\nolimits_{{i_1},{i_{2,}},...,{i_n}} {{{\left| {{x_{{i_1},{i_2},...,{i_N}}}} \right|}^2}} }$ }\\
${\left\| {\cal X} \right\|_1} $ & \makecell{ $l_1$ norm of tensor $\cal X$, defined as \\${\left\| {\cal X} \right\|_1} = \sum\nolimits_{{i_1},{i_{2,}},...,{i_n}} {\left| {{x_{{i_1},{i_2},...,{i_N}}}} \right|}$} \\
$\left\langle {{{\cal X}_1},{{\cal X}_2}} \right\rangle$ & \makecell{the inner product for two tensors, defined as\\ $\left\langle {{{\cal X}_1},{{\cal X}_2}} \right\rangle  = \sum\nolimits_{{i_1},{i_{2,}},...,{i_n}} {{x_1}_{{i_1},{i_2},...,{i_N}}{x_2}_{{i_1},{i_2},...,{i_N}}} $ }\\
$\otimes $ & Kronecker product   \\
$ \odot $  & component-wise multiplication \\
${ \times _n}$  & mode-$n$ tensor-matrix product\\
\bottomrule
\end{tabular}
\label{tab:notion}
\end{table}

{\textit {Definition 1 (Mode-$n$ tensor-matrix product)}}: the $n$-mode (matrix) product of a tensor ${\cal X} \in {\mathbb{R}^{{I_1} \times {I_2} \times {I_3} \times  \cdots  \times {I_N}}}$ with a matrix $U \in {\mathbb{R}^{J \times {I_N}}}$ is defined as ${\cal X}{ \times _n}U$ with size ${I_1} \times  \cdots  \times {I_{n - 1}} \times J \times {I_{n + 1}} \times  \cdots  \times {I_N}$ , which is written as:
\begin{equation}
  {\left( {{\cal X}{ \times _n}U} \right)_{{i_1}...{i_{n - 1}}j{i_{n + 1}}...{i_N}}} = \sum\limits_{{i_n}}^{{I_n}} {{x_{{i_1}{i_2}...{i_N}}}{u_{j{i_n}}}}
  \label{eq:1}
  \nonumber
  \end{equation}
According to the mode-$n$ fiber multiplication, it can also be rewritten as:
\begin{equation}
  {\cal Y} = {\cal X}{ \times _n}U \Leftrightarrow {Y_{(n)}} = U{X_{(n)}}
  \label{eq:2}
  \nonumber
  \end{equation}

\begin{figure}[htb!]
\centering
\includegraphics[width=6cm]{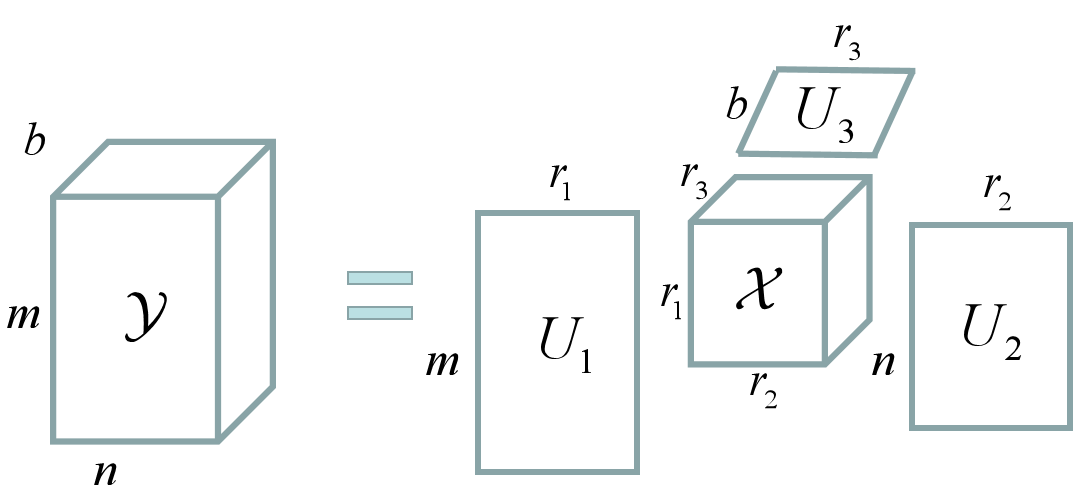} \\
\caption{Illustration of 3-D tensor Tucker decomposition.}
\label{Tucker decomposition}
\end{figure}

  {\textit {Definition 2 (Tucker decomposition and Tucker rank)}}: Tucker decomposition is formulated as a high order principal component analysis. 
It decomposes a tensor into a core tensor multiplied by a matrix along each mode, for example in a 3-D tensor, as shown in Fig. \ref{Tucker decomposition}. 
For a $N$-order tensor ${\cal X} \in {\mathbb{R}^{{I_1} \times {I_2} \times {I_3} \times  \cdots  \times {I_N}}}$, let ${U^{(n)}} \in {\mathbb{R}^{{J_n} \times {I_n}}}$ in any $n \in \left\{ {1,...,N} \right\}$, we obtain:
\begin{equation}
\begin{array}{l}
  {\cal Y} = {\cal X}{ \times _1}{U^{(1)}} \times {U^{(2)}} \cdots { \times _N}{U^{(N)}}\\
  {\rm{      }} \Leftrightarrow {Y_{(n)}} = {U^{(n)}}{X_{(n)}}{\left( {{U^{(N)}} \otimes  \cdots  \otimes {U^{(1)}}} \right)^{\rm T}}
  \end{array}
  \label{eq:3}
  \nonumber
  \end{equation}
And the tucker rank is defined as:
  \begin{equation}
   Ran{k_N}({\cal X}) = (Rank({U_{(1)}}),Rank({U_{(2)}}),...,Rank({U_{(N)}}))
  \label{eq:4}
  \nonumber
  \end{equation}

\subsection{Tensor Decomposition for Infrared Small Target Detection}
For an infrared small target image, it can be mathematically expressed as a linear superposition model by target, background, and noise components:
\begin{equation}
{f_Y} = {f_F} + {f_T} + {f_N}
  \label{eq:5}
\end{equation}

% A large number of methods have been devised for the purpose of discriminating targets from background elements. A comprehensive survey on these approaches can be found in \cite{Zhao2022single}. As the focus of our study lies in tensor decomposition-based methods, we will discuss this aspect in detail. One of the pioneering works on infrared small target detection leveraging tensor decomposition is presented in \cite{Dai2017Reweighted}. This study introduces the Infrared Patch Tensor (IPT) model by arranging overlapping patches extracted from a single image into a third-order tensor.
Numerous methods have been developed to distinguish targets from background elements, extensively reviewed in \cite{Zhao2022single}. Our focus is on tensor decomposition-based methods, particularly highlighted by the Infrared Patch Tensor (IPT) model introduced in \cite{Dai2017Reweighted}. This model arranges overlapping image patches into a third-order tensor and achieves target-background separation by applying optimally regularized constraints where ${J_1}({\cal F})$ and ${J_2}({\cal T})$ are constraint functions representing background and target priors, respectively, balanced by $\lambda$ and influenced by Gaussian noise density variance $\eta$.
This model is formulated as:
\begin{equation}
  {\cal Y} = {\cal F} + {\cal T} + {\cal N}
    \label{eq:6}
\end{equation}
where $\cal Y$, $\cal F$, $\cal T$, $\cal N$ denotes the tensor model of infrared image, background component, target component and noise component, respectively.
% According to the characteristics of different components, the purpose of target-background separation is achieved by introducing optimal regularized constraints.
The purpose of target-background separation is achieved by introducing optimal regularized constraints based on the characteristics of different components.
Then, the generalized form can be written as:
\begin{equation}
\mathop {\min }\limits_{{\cal F},{\cal T}} {J_1}\left( {\cal F} \right) + \lambda {J_2}\left( {\cal T} \right)  s.t.\left\| {{\cal Y} - {\cal F} - {\cal T}} \right\|_F^2 \le \eta
    \label{eq:7}
\end{equation}
where ${J_1}({\cal F})$ and ${J_2}({\cal T})$ represent constraint functions used to encode the background prior and target prior, respectively. $\lambda$ is a positive tradeoff between the two components, while $\eta$ denotes the Gaussian noise density variance. 
Consequently, numerous studies have explored methods to precisely extract small targets by investigating regularization techniques. Among these, the sum of the nuclear norm and reweighted $l_1$ norm are employed to constrain the background and small target components in the IPT model \cite{Dai2017Reweighted}. However, this approach is susceptible to sparse interference, similar to small targets, due to the approximation error of the nuclear norm constraint.

% To address this issue, several improvements have been developed by introducing non-convex surrogates, such as the partial sum of tensor nuclear norm \cite{zhang2019infrared}, log operation tensor fiber nuclear norm \cite{Kong2021Infrared}, or by incorporating well-designed local spatial priors, such as tree structure \cite{Zhao2021three} or background component measurement \cite{Yang2022infrared}. 
% Additionally, spatial-temporal tensor models have been constructed to exploit interframe information \cite{zhou2021infrared,Sun2020Infrared}. Under these models, various regularizations are employed to describe the underlying sequential priors, such as the nonlocal self-similarity and local structure prior in the spatial domain \cite{Sun2020Infrared,Liu2020Small}, as well as interframe correlation in the temporal domain \cite{zhou2021infrared, Pang2022STTM}.
% In particular, TV regularization and saliency regularization are designed to exploit spatial-temporal saliency information \cite{Kong2021Infrared,zhnag2022infrared, Pang2022STTM}.
To address this issue, improvements have been developed by introducing non-convex surrogates \cite{zhang2019infrared,Kong2021Infrared} or incorporating well-designed local spatial priors \cite{Zhao2021three,Yang2022infrared}. Spatial-temporal tensor models have also been constructed to exploit interframe information \cite{zhou2021infrared,Sun2020Infrared}, employing various regularizations to describe underlying sequential priors in the spatial \cite{Sun2020Infrared,Liu2020Small} and temporal domains \cite{zhou2021infrared, Pang2022STTM}. TV regularization and saliency regularization are designed to exploit spatial-temporal saliency information \cite{Kong2021Infrared,zhnag2022infrared, Pang2022STTM}.

% Nevertheless, these methods still exhibit two issues: inadequate utilization of background information and incomplete interference suppression. To address these shortcomings, the proposed model employs differential directionality and mixed sparse constraint to characterize more intrinsic knowledge. Furthermore, saliency coherence factors are devised to enhance differentiation between background interference and small targets.
Despite advancements, these methods still face challenges in fully utilizing background information and effectively suppressing interference. Our proposed model introduces differential directionality and mixed sparse constraints for a more intrinsic understanding, with saliency coherence factors to better differentiate between background interference and small targets.

%% file: contents/method.tex
% !TEX root = ../main.tex
% \bibliography{../reference.bib}

\section{Detection Model} \label{sec:method}
\subsection{Differential Directionality Mixed Sparse Regularization}
In dynamic scenes, background variations reduce correlation across frames, challenging models reliant on low-rank sparse priors for effective target separation \cite{wang2022Infrared,chen2019hypsectral}.
Analysis reveals a latent directionality in background components: homogeneous regions and targets show no directionality, while clutter exhibits directionality.
This directionality is effectively captured by first-order differential operators in horizontal and vertical directions \cite{moradi2020fast,Li2023Infrared,zheng2020double}.
Despite gradual background changes, strong continuity exists between frames.
Thus, we propose a spatio-temporal differential prior with hybrid sparse constraints to address both directionality and continuity.

\begin{figure*}[htb!]
  \centering
  \includegraphics[width=16cm]{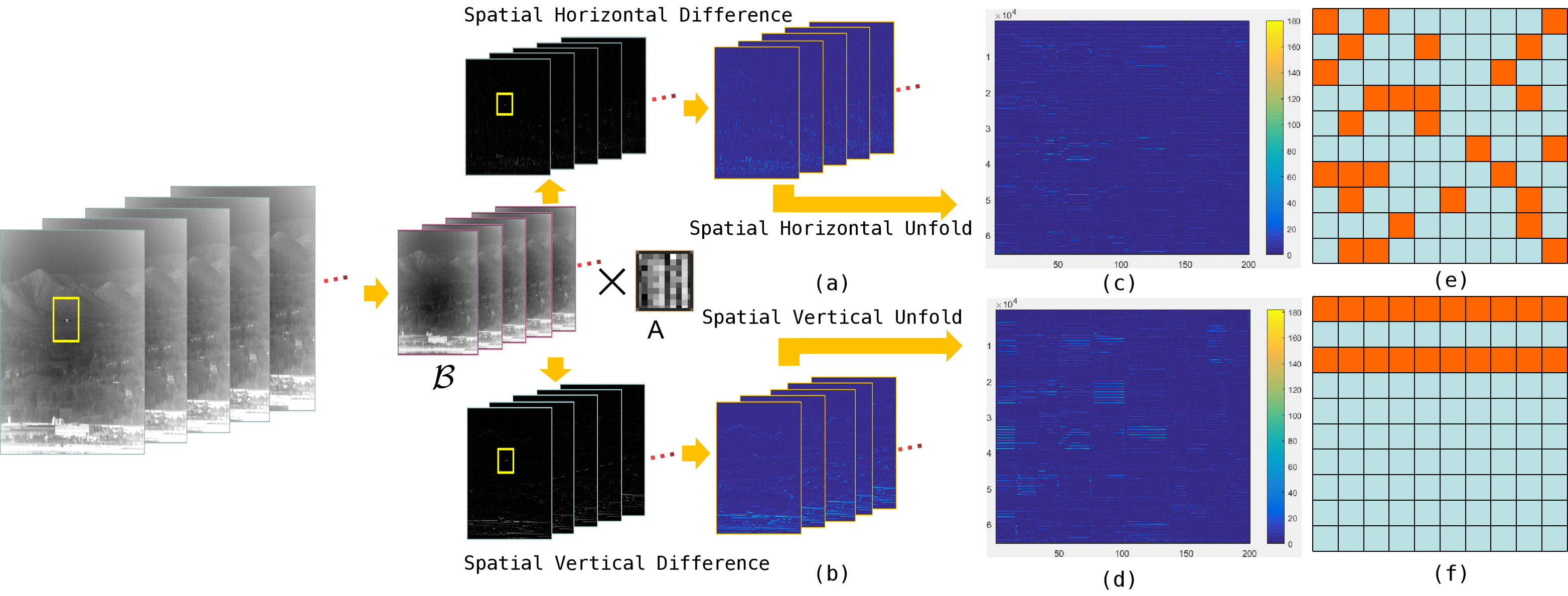} \\
  \caption{The difference images along the temporal mode for infrared sequence data in both spatial modes. (a) The heatmaps of the spatial horizontal difference, (b) The heatmaps of the spatial vertical difference,  (c) and (d) The matrices of corresponding difference images along the temporal mode, (e) and (f) Display of the shared and unshared sparsity.}
  \label{fig:difference images}
  \end{figure*}

% In decomposition, the redundant low-rank component $\cal F$ can be further decomposed into the mode-3 tensor-matrix product:
The low-rank component $\cal F$ is further decomposed into a mode-3 tensor-matrix product:
\begin{equation}
{\cal F} = {\cal B}{ \times _3}A
  \label{eq:tensorproduct}
  \end{equation}
% where ${\cal B} \in {\mathbb{R}^{{n_1} \times {n_2} \times r}}$($r \ll {n_3}$) is the spatial factor, and $A \in {\mathbb{R}^{{n_3} \times r}}$ is the temporal factor and satisfies ${A^{\rm T}}A = I$.
where ${\cal B} \in {\mathbb{R}^{{n_1} \times {n_2} \times r}}$ represents the spatial factor, and $A \in {\mathbb{R}^{{n_3} \times r}}$ is the temporal factor with ${A^{\rm T}}A = I$.
We construct spatial-temporal differential images to explore background priors, using first-order difference operators as shown in Fig.~\ref{fig:difference images}, which displays heatmaps of differential images for spatial factors. The residual distribution shows notable disparity in horizontal and vertical images, with unfolding matrices presented for clarity. Distinct sparse patterns, namely unshared and shared, are illustrated, indicating significant non-zero element distribution differences.
% To thoroughly investigate the potential prior of the background cube, we construct spatial-temporal differential images based on the two factors, utilizing a first-order difference operator, as depicted in Fig. \ref{fig:difference images}.

% In Fig. \ref{fig:difference images}, subfigures (a) and (b) display the heatmaps of the horizontal and vertical differential images for each slice in the spatial factors, respectively. As evidenced by these subfigures, the residual distribution of the horizontal and vertical differential images exhibits a notable disparity. For clearer visualization, the unfolding matrices along the temporal dimension for the two differential components are presented in subfigures (c) and (d). From these matrices, it is evident that despite both exhibiting sparsity, their sparse patterns differ.

% \hl{Specifically, examples of distinct sparse patterns, namely unshared and shared patterns, are illustrated in Fig. \ref{fig:difference images}(e) and (f).} Within these images, cyan and orange blocks represent zero and non-zero elements in the unfolding matrices, respectively. This observation indicates that the number of non-zero elements is significantly lower than that of zero elements in both grids. However, the key distinction between the two sparse patterns lies in the distribution of non-zero elements. Specifically, the non-zero elements in Fig. \ref{fig:difference images}(e) exhibit a haphazard distribution, corresponding to the unshared sparse pattern, while those in Fig. \ref{fig:difference images}(f) are arranged regularly in rows, manifesting a shared row sparse pattern.

To address this issue, a mixed sparse constraint approach is employed, with the $l_1$ norm constraint applied to the spatial horizontal mode and the $l_{2,1}$ norm constraint utilized for the spatial vertical mode. This enables the full exploitation of the sparse prior of different patterns, enhancing the algorithm's robustness across varied dynamic conditions. Additionally, differential information is constructed for the temporal factor and constrained by the $l_2$ norm to capture local temporal coherence in dynamic scenes.

% Employing a single sparsity regularization, such as the $l_1$ norm or its non-convex relaxations, to characterize both sparse patterns would neglect common features within the shared sparse pattern, such as group sparsity. Moreover, utilizing the same group sparsity constraint, such as the mixed $l_{2,1}$ norm, for both sparse patterns could hinder the effective differentiation of components lacking shared features \cite{zhang2018infrared}. Consequently, a single sparse constraint is inappropriate for characterizing both the spatial horizontal and vertical differential priors.

% To address this issue, a mixed sparse constraint approach is employed to regularize the differential images along the two spatial dimensions, with the $l_1$ norm constraint applied to the spatial horizontal mode and the $l_{2,1}$ norm constraint utilized for the spatial vertical mode. This mixed sparse constraint enables the full exploitation of the sparse prior of different patterns, thereby enhancing the algorithm's robustness across varied dynamic conditions. Additionally, as the temporal factor $A$ represents key information denoting the continuity between interframes of a sequence, differential information is also constructed for the temporal factor and constrained by the $l_2$ norm.
% By incorporating this continuity constraint, the algorithm can efficiently capture the local temporal coherence in dynamic scenes, yielding more precise and consistent results compared to approaches that lack a continuity constraint.

The regularization on the low-rank background component can be modeled as follows:
\begin{equation}
{J_{\rm{1}}}\left( {\cal F} \right){\rm{ = }} {\left\| {{\cal B}{ \times _1}{D_1}} \right\|_{2,1}} + {\left\| {{\cal B}{ \times _2}{D_2}} \right\|_1} + \lambda \left\| {{D_3}A} \right\|_2^2
    \label{eq:lowrankconstraint}
    \end{equation}
where $\lambda$ is the tradeoff between different items; ${D_k}(k = 1,2,3)$ are first-order difference matrices in three dimensions of infrared sequence.
And the differential operation of each term can be obtained:
\begin{equation}
\begin{array}{l}
  {\cal B}{ \times _1}{D_1} = {\cal B}\left( {i + 1,j,t} \right) - {\cal B}\left( {i,j,t} \right)\\
  {\cal B}{ \times _2}{D_2} = {\cal B}\left( {i,j + 1,t} \right) - {\cal B}\left( {i,j,t} \right)\\
  {D_3}A = A\left( {:,t + 1} \right) - A\left( {:,t} \right)
  \end{array}
  \label{eq:difference factor}
\end{equation}
To enhance the sparsity of the difference images, a reweighted strategy is introduced in the mixed sparse regularization. 
The regularization on the low-rank component is rewritten as follows:
\begin{equation}
\begin{array}{l}
  {J_1}\left( {\cal F} \right) = {\left\| {  {\cal B}{ \times _1}{D_1}} \right\|_{2,1,{{\cal W}_1}}} + {\left\| { {\cal B}{ \times _2}{D_2}} \right\|_{1,{{\cal W}_2} }}\\
  {\rm{ ~~~~~~~~~~~~          }} + \lambda \left\| {{D_3}A} \right\|_2^2
  \end{array}
    \label{eq:lowrankregularization}
  \end{equation}
  where ${\cal W}_1$ and ${\cal W}_2$ are weight tensors.

  \subsection{Adaptive Saliency Coherence Map}
  % Infrared sensing scenes frequently encounter various interferences, such as cloud formations, sea waves, and surface structures, among others. 

  % Such interference frequently occurs in infrared sensing scenarios, including cloud clutter, sea waves, and ground structures. These interfering structures exhibit more significant inter-frame changes in dynamic scenes and display sparsity similar to the target, leading to misclassification during target-background separation.
  Infrared sensing often experiences interference from elements like cloud clutter, sea waves, and ground structures. These structures exhibit significant inter-frame changes and similar sparsity to the target, causing misclassification during separation.
  % To suppress these interferences, some studies \cite{Dai2017Reweighted, zhou2021infrared, Zhang2020Edge, Pang2022STTM} have meticulously designed clutter suppression factors based on the structure tensor and incorporated them into the target-background separation process.
  To mitigate these interferences, previous studies have designed clutter suppression factors \cite{Dai2017Reweighted, zhou2021infrared, Zhang2020Edge, Pang2022STTM}, but these methods can still miss targets or leave residual salient points.
  % Although substantial progress has been achieved in eliminating edge clutter, the potential for missed targets or residual salient corner points persists. The primary reason for this is that previous methods employing single-scale structure tensors exhibit limited capabilities in characterizing the significant coherence between different background components, and overlook the decay of target energy during the decomposition process. 
  This is primarily due to the limited capabilities of single-scale structure tensors in characterizing coherence between background components, and the overlooked decay of target energy during decomposition.
  % To address this challenge, we propose an adaptive saliency coherence exponent (ASCE) map, which is devised based on the pointwise structure tensor computed from the image.
  To tackle this, we propose an adaptive saliency coherence exponent (ASCE) map, based on the pointwise structure tensor computed from the image.

 \begin{figure*}[htb!]
  \centering
  \subfigure[]{
     \includegraphics[width=3.0cm]{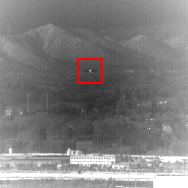}
   % \label{1a}
   }
   \hspace{-4mm}
  \subfigure[]{
      \includegraphics[width=3.0cm]{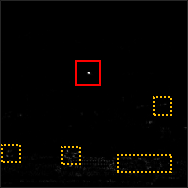}
  % \label{1b}
  }
   \hspace{-4mm}
%  \vskip -5pt
  \subfigure[]{
      \includegraphics[width=3.0cm]{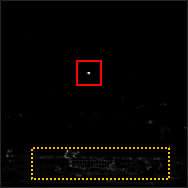}
  % \label{1c}
  }
   \hspace{-4mm}
    \subfigure[]{
      \includegraphics[width=3.0cm]{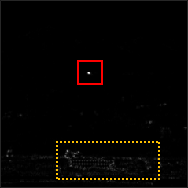}
   % \label{1d}
  }
   \hspace{-4mm}
  \subfigure[]{
     \includegraphics[width=3.0cm]{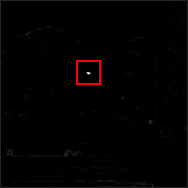}
  % \label{1a}
  }
        \vskip -5pt
  \caption{ Illustration of single and multiscale saliency coherence maps. (a) Original image (b)-(d) The saliency map obtained by the first to three saliency coherence items. (e) The saliency map obtained by the transformed enhancement factor. }
\label{fig:itemeffect}
\end{figure*}

Given an infrared image $Y \in {\mathbb{R}^2}$, the smoothed structure tensor (SST) \cite{Prasath2015multiscale} can be computed at every pixel:
\begin{equation}
{K_\sigma }\left( {I\left( {x,y} \right)} \right) = {G_\sigma } \star \left( {\nabla I\nabla {I^{\rm T}}} \right) = \left[ {\begin{array}{*{20}{c}}
  {{G_\sigma } \star I_x^2}&{{G_\sigma } \star {I_x}{I_y}}\\
  {{G_\sigma } \star {I_y}{I_x}}&{{G_\sigma } \star I_y^2}
  \end{array}} \right]
  \label{eq:structuretensor}
  \end{equation}
where $ {G_\sigma } = {\left( {\sigma \sqrt {2\pi } } \right)^{ - 1}}\exp \left( { - {{\left| {\rm{x}} \right|}^2}/2{\sigma ^2}} \right)$ denotes the Gaussian kernel with the scale parameter $\sigma > 0$, and $ \star $ is the 2D convolution operation.
In general, ${K_\sigma }$ contains a maximum-minimum eigenvalue pair $\left( {{e_ + }(x,y,\sigma ),{e_ - }(x,y,\sigma )} \right)$ and ${{e_ + } \ge {e_ - }}$.
The eigenvalues characterize the local background regions within a size $\sigma$ neighborhood with the range of their values, specifically, indicating ${e_ + } \ge {e_ - } \gg 0$ for corner region, ${e_ + } \gg {e_ - } \approx 0$ for edge region, ${e_ + } \approx {e_ - } \approx 0$ for the flat region.
Based on this, the adaptive saliency coherence exponent function can be formulated as follows:
 %  \begin{equation}
 %   \begin{array}{l}
 %  ASCE(x,y,\sigma ) = 1 - \exp \left( { - {\alpha _1}\sqrt {\left| {{e_ + }{e_ - }} \right|} } \right)\\
 %  {\rm{~~~~~~~~~~~~~~~~~~~                      }} \times {\rm{exp}}\left( { - {\alpha _2}\left| {\frac{{{e_ + } + {e_ - }}}{{{{({e_ + } - {e_ - } + \delta )}^{1/2}}}}} \right|} \right)\\
 %  {\rm{ ~~~~~~~~~~~~~~~~~~~                 }} \times \exp \left( { - {\alpha _3}\frac{{{e_2}{e_ - }}}{{{e_ + } + {e_ - } + \delta }}} \right)
 %  \end{array}
 %    \label{eq:multiscalestructure}
 % \end{equation}
 % where $\delta  = 0.001$ is a smoothed parameter avoiding division by zero.
 % The $ASCE(x,y,\sigma )$ represents the saliency coherence factor for portraying various background components at the pixel $(x,y)$.
% To analyze the effect of the factor, we rewrite the $ASCE(x,y,\sigma)$ as follows:
  \begin{equation}
ASCE(x,y,\sigma ) = 1 - \exp \left( { - \sum\limits_{i = 1}^3 {{\alpha _i}{C_i}({e_ + },{e_ - },\sigma )} } \right)
  \label{eq:remultiscalestructure}
  \end{equation}
where ${\alpha _i} \ge 0(i = 1,2,3)$ are the balance parameters.$\delta  = 0.001$ is a smoothed parameter avoiding division by zero.
The $ASCE(x,y,\sigma )$ represents the saliency coherence factor for portraying various background components at the pixel $(x,y)$.

  \begin{enumerate}[]
  \item The first item is represented as:
         \begin{equation}
         {C_1}({e_ + },{e_ - },\sigma ) = \sqrt{\left| {{e_ + }{e_ - }} \right|}
         \label{eq:item1}
         \end{equation}
The design is a corner measure \cite{vicas2015detecting, akl2015texture}, used to capture the structure of non-target points with high spatial frequencies along two oriented directions, as shown in Fig. \ref{fig:itemeffect}(b).

  \item The second item is defined as:
        \begin{equation}
      {C_2}({e_ + },{e_ - },\sigma ) = \left| {\frac{{{e_ + } + {e_ - }}}{{{{({e_ + } - {e_ - } + \delta )}^{1/2}}}}} \right|
         \label{eq:item2}
         \end{equation}
It is deemed as coherence measure, where its value is close to one in edge pixels and close to zero in homogeneous or noisy regions, as presented in Fig. \ref{fig:itemeffect}(c).

  \item The third item is written as:
        \begin{equation}
        {C_3}({e_ + },{e_ - },\sigma ) = \frac{{{e_ + }{e_ - } }}{{{e_ + } + {e_ - } + \delta }}
          \label{eq:item3}
        \end{equation}
The item can be seen as edge indicator since its values encode the background components with strong gradient, such as the edge region tending to zero, the corner component tending to one, as presented in Fig. \ref{fig:itemeffect}(d).
  \end{enumerate}

Furthermore, it can be observed that the $ASCE(x,y,\sigma)$ indicator ranges between 0 and 1. The greater the contrast of a background component, the closer its corresponding value approaches 1, while the values of other components gravitate toward 0. Generally, small targets exhibit higher local contrast areas, causing their corresponding $ASCE(x,y,\sigma)$ values to be nearer to 1 than those of other components, as illustrated in Fig. \ref{fig:itemeffect}(e). During tensor decomposition, components with heightened contrast typically possess increased sparsity and can be more easily separated. However, if $ASCE(x,y,\sigma)$ is directly employed as the penalty factor, target saliency may diminish gradually throughout the decomposition iteration. Consequently, we propose a variant of $ASCE(x,y,\sigma)$:
\begin{equation}
{W_{ASCE}} = \left\{ {\begin{array}{*{20}{c}}
  {1 + ASCE(x,y,\sigma ),}\\
  {ASCE(x,y,\sigma ),}
  \end{array}} \right.\begin{array}{*{20}{c}}
  {ASCE(x,y,\sigma ) \ge {T_S}}\\
  {otherwise}
  \end{array}
    \label{eq:enhancefactor}
  \end{equation}
  where, ${T_s}$ is a threshold defined as $mean(ASCE(x,y,\sigma )) + 5{\mathop{\rm var}} (ASCE(x,y,\sigma ))$.

  The target enhancement factor of an infrared sequence cube can be obtained by stacking the improved $ASCE(x,y,\sigma)$ maps of each frame. 
We incorporate this factor and a reweighted strategy into the constraint of small target components:
\begin{equation}
% \mathcolorbox{yellow}{\left\| {{{\cal W}_{ASCE}} \odot {\cal T}} \right\|_{1,{\cal W}{_S}}}
\left\| {{{\cal W}_{ASCE}} \odot {\cal T}} \right\|_{1,{\cal W}{_S}}
\label{eq:sparsereg}
\end{equation}
where ${{\cal W}_S}$ denotes the weight of each element in the sparse component.

Finally, the proposed model is constructed by integrating the spatial-temporal difference prior with the saliency coherence exponent map, as follows:
\begin{equation}
\begin{array}{l}
  \mathop {\min }\limits_{A,{\cal B},{\cal T}} {\left\| {{\cal B}{ \times _1}{D_1}} \right\|_{2,1,{{\cal W}_1}}} + {\left\| {{\cal B}{ \times _2}{D_2}} \right\|_{1,{{\cal W}_2}}}\\
  {\rm{~~~~~~~~~~  }} + \lambda \left\| {{D_3}A} \right\|_F^2 + \gamma {\left\| {{{\cal W}_{ASCE}} \odot {\cal T}} \right\|_{1,{{\cal W}_S}}}\\
  s.t.\left\| {{\cal Y} - {\cal B}{ \times _3}A - {\cal T}} \right\|_F^2 \le \varepsilon
  \end{array}
  \label{eq:model}
\end{equation}

\section{Model solution based on PAM}
We solve the proposed model based on the PAM-based iterative algorithm by alternately updating each variable with the others fixed. 
First, the objective function is transformed into an equivalent form:
\begin{equation}
\begin{array}{l}
  \mathop {\min }\limits_{A,{\cal B},{\cal T}} \frac{1}{2}\left\| {{\cal Y} - {\cal B}{ \times _3}A - {\cal T}} \right\|_F^2 + \gamma {\left\| {{{\cal W}_{ASCE}} \odot T} \right\|_{1,{{\cal W}_S}}}\\
  ~~~~~ + \lambda \left\| {{D_3}A} \right\|_F^2 + ({\left\| {{\cal B}{ \times _1}{D_1}} \right\|_{2,1,{{\cal W}_1}}} + {\left\| {{\cal B}{ \times _2}{D_2}} \right\|_{1,{{\cal W}_2}}})
  \end{array}
  \label{eq:tranfunction}
\end{equation}

By separating each variable, we can update them by minimizing the following sub-problems:
\begin{equation}
\begin{array}{l}
  {A^{k + 1}} = \arg {\min _A}f\left( {A,{{\cal B}^k},{{\cal T}^k}} \right) + \frac{\rho }{2}\left\| {A - {A^k}} \right\|_F^2\\
  {{\cal B}^{k + 1}} = \arg {\min _{\cal B}}f\left( {{A^{k + 1}},{\cal B},{{\cal T}^k}} \right) + \frac{\rho }{2}\left\| {{\cal B} - {{\cal B}^k}} \right\|_F^2\\
  {{\cal T}^{k + 1}} = \arg {\min _{\cal T}}f\left( {{A^{k + 1}},{{\cal B}^{k + 1}},{\cal T}} \right) + \frac{\rho }{2}\left\| {{\cal T} - {{\cal T}^k}} \right\|_F^2
  \end{array}
  \label{eq:subproblems}
\end{equation}
where $f(A,{\cal B},{\cal T})$ is the equivalent objective function and $\rho > 0$ is a proximal parameter.

 1) {\bf Updating} $A$: by eliminating other items unrelated to $A$, the subproblem is defined as:
  \begin{equation}
\begin{array}{l}
  \mathop {\arg \min }\limits_A \frac{1}{2}\left\| {{\cal Y} - {\cal B}{ \times _3}A - {\cal T}} \right\|_F^2\\
  {\rm{ ~~~~~~~~          }} + \lambda \left\| {{D_3}A} \right\|_F^2 + \frac{\rho }{2}\left\| {A - {A^k}} \right\|_F^2
  \end{array}
    \label{eq:solutionA3}
  \end{equation}
  whose solver can be directly obtained by solving the following Sylvester matrix equation:
  \begin{equation}
\begin{array}{l}
  AB_{(3)}^k{(B_{(3)}^k)^{\rm T}} + 2\lambda D_3^{\rm T}{D_3}A + \rho A\\
  {\rm{ ~~~~~~~~~~      }} = ({Y_{(3)}} - T_{(3)}^k){(B_{(3)}^k)^{\rm T}} + \rho {A^k}
  \end{array}
    \label{eq:solutionA1}
  \end{equation}

   A fast solution of the above equation can be obtained by diagonalizing the circulant matrix $D_3^T{D_3}$ and a symmetric matrix  $B_{(3)}^k{(B_{(3)}^k)^{\rm T}}$ through the 1-D fast Fourier transformation (FFT) and SVD, and its specific process can be referred to in Ref. \cite{kirrinnis2001fast}.
   % separately expressing as:
   % \begin{equation}
% D_3^{\rm T}{D_3} = F_1^{\rm T}{\Psi _1}{F_1}{\rm{ ~~}}and{\rm{ ~~ }}B_{(3)}^k{(B_{(3)}^k)^{\rm T}} = {U_1}\sum {}_1U_1^{\rm T}
    % \label{eq:solutonA2}
   % \end{equation}
   % where $F_1$ is 1-D discrete Fourier transformation (DFT) matrix. The subproblem can then be solved as:
   % \begin{equation}
% {A^{k + 1}} = F_1^T((1 \oslash {M_1}) \odot ({F_1}G{U_1}))U_1^T
    % \label{eq:solutionA3}
   % \end{equation}
   % where $G = ({Y_{(3)}} - T_{(3)}^k){(B_{(3)}^k)^{\rm T}} + \rho {A^k}$ and ${M_1} = 2\lambda (diag({\Psi _1}),diag({\Psi _1}),...,diag({\Psi _1}) + \rho one{s^3}({n_3},r)) + {(diag({\Sigma _1}),diag({\Sigma _1}),...,diag({\Sigma _1}))^{\rm T}}$.

     2) {\bf Updating} $\cal B$: the subproblem is obtained by picking out the items related to  $\cal B$:
   \begin{equation}
   \begin{array}{l}
  \mathop {\min }\limits_{\cal B} \frac{1}{2}\left\| {{\cal Y} - {\cal B}{ \times _3}A - {\cal T}} \right\|_F^2 + \frac{\rho }{2}\left\| {{\cal B} - {{\cal B}^k}} \right\|_F^2\\
   ~~~~~+ ({\left\| {{\cal B}{ \times _1}{D_1}} \right\|_{2,1,{{\cal W}_1}}} + {\left\| {{\cal B}{ \times _2}{D_2}} \right\|_{1,{{\cal W}_2}}})
  \end{array}
    \label{eq:solutionB}
   \end{equation}
   It can be solved by alternating direction method of multipliers(ADMM) \cite{boyd2011distributed}.
   By introducing two auxiliary valuables, the above equation is rewritten as:
   \begin{equation}
   \begin{array}{l}
  \mathop {\min }\limits_{{\cal B},{{\cal Z}_1},{{\cal Z}_2}} \frac{1}{2}\left\| {{\cal Y} - {\cal B}{ \times _3}A - {\cal T}} \right\|_F^2 + \frac{\rho }{2}\left\| {{\cal B} - {{\cal B}^k}} \right\|_F^2\\
  ~~~~~~~~~~ + ({\left\| {{{\cal Z}_1}} \right\|_{2,1,{{\cal W}_1}}} + {\left\| {{{\cal Z}_2}} \right\|_{1,{{\cal W}_2}}}){\rm{          }}\\
  s.t.{\rm{  }}{\cal B}{ \times _1}{D_1} = {{\cal Z}_1},{\rm{ }}{\cal B}{ \times _2}{D_2} = {{\cal Z}_2}
  \end{array}
    \label{eq:solutionB1}
   \end{equation}
   Its augmented Lagrangian function is:
   \begin{equation}
   \begin{array}{l}
  {{\cal L}_\beta }\left( {{\cal B},{{\cal Z}_1},{{\cal Z}_2},{{\cal P}_1},{{\cal P}_2}} \right) = \frac{1}{2}\left\| {{\cal Y} - {\cal B}{ \times _3}A - {\cal T}} \right\|_F^2\\
  {\rm{~~~~~~~~         }} + {\left\| {{{\cal Z}_1}} \right\|_{2,1,{{\cal W}_1}}} + {\left\| {{{\cal Z}_2}} \right\|_{1,{{\cal W}_2}}} + \frac{\rho }{2}\left\| {{\cal B} - {{\cal B}^k}} \right\|_F^2\\
  {\rm{~~~~~~~~           + }}\left\langle {{\cal B}{ \times _1}{D_1} - {{\cal Z}_1},{{\cal P}_1}} \right\rangle  + \frac{\beta }{2}\left\| {{\cal B}{ \times _1}{D_1} - {{\cal Z}_1}} \right\|_F^2\\
  {\rm{~~~~~~~~          + }}\left\langle {{\cal B}{ \times _2}{D_2} - {{\cal Z}_2},{{\cal P}_2}} \right\rangle  + \frac{\beta }{2}\left\| {{\cal B}{ \times _2}{D_2} - {{\cal Z}_2}} \right\|_F^2
  \end{array}
    \label{eq:solutionBaux}
   \end{equation}
where $\beta  > 0$ is penalty factor.
The solution of the function can be obtained by alternately updating each variable:
\begin{equation}
\begin{array}{l}
  {{\cal B}^{k + 1,l + 1}} = \arg {\min _{\cal B}}{{\cal L}_\beta }({\cal B},{\cal Z}_1^l,{\cal Z}_2^l,{\cal P}_1^l,{\cal P}_2^l)\\
  {\cal Z}_1^{l + 1} = \arg {\min _{{{\cal Z}_1}}}{{\cal L}_\beta }({{\cal B}^{k + 1,l + 1}},{{\cal Z}_1},{\cal Z}_2^l,{\cal P}_1^l,{\cal P}_2^l)\\
  {\cal Z}_2^{l + 1} = \arg {\min _{{{\cal Z}_2}}}{{\cal L}_\beta }({{\cal B}^{k + 1,l + 1}},{\cal Z}_1^{l + 1},{{\cal Z}_2},{\cal P}_1^l,{\cal P}_2^l)\\
  {\cal P}_1^{l + 1} = \arg {\min _{{{\cal P}_1}}}{{\cal L}_\beta }({{\cal B}^{k + 1,l + 1}},{\cal Z}_1^{l + 1},{\cal Z}_2^{l + 1},{{\cal P}_1},{\cal P}_2^l)\\
  {\cal P}_2^{l + 1} = \arg {\min _{{{\cal P}_2}}}{{\cal L}_\beta }({{\cal B}^{k + 1,l + 1}},{\cal Z}_1^{l + 1},{\cal Z}_2^{l + 1},{\cal P}_1^{l + 1},{{\cal P}_2})
  \end{array}
  \label{eq:solutionB2}
\end{equation}

For ${{\cal B}^{k + 1,l + 1}}$, its updating is achieved by the following function:
\begin{equation}
\begin{array}{l}
  \arg \mathop {\min }\limits_{\cal B} \frac{1}{2}\left\| {{\cal Y} - {\cal B}{ \times _3}{A^{k + 1}} - {{\cal T}^k}} \right\|_F^2 + \frac{\rho }{2}\left\| {{\cal B} - {{\cal B}^k}} \right\|_F^2\\
  {\rm{    }} + \frac{\beta }{2}\left\| {{\cal B}{ \times _1}{D_1} - {{\cal Z}_1} + \frac{{{\cal P}_1^l}}{\beta }} \right\|_F^2 + \frac{\beta }{2}\left\| {{\cal B}{ \times _2}{D_2} - {{\cal Z}_2} + \frac{{{\cal P}_2^l}}{\beta }} \right\|_F^2
  \end{array}
  \label{eq:solutionB7}
\end{equation}
By taking the derivative of variable $\cal B$ and making it equal to zero, its solution can be computed by the following equation:
\begin{equation}
{\cal B}{ \times _3}({({A^{k + 1}})^{\rm T}}{A^{k + 1}} + \beta ({\cal B}{ \times _1}(D_1^{\rm T}{D_1}) + {\cal B}{ \times _2}(D_2^{\rm T}{D_2}) = {\cal K}
  \label{eq:solutionB4}
\end{equation}
where ${\cal K} = ({\cal Y} - {{\cal T}^k}){ \times _3}{({A^{k + 1}})^{\rm T}} + \beta (({\cal Z}_1^l - \frac{{{\cal P}_1^l}}{\beta }) + ({\cal Z}_2^l - \frac{{{\cal P}_2^l}}{\beta }) + \rho {{\cal B}^k}$.

The equation is further rewritten as:
\begin{equation}
B_{(3)}^{\rm T}({({A^{k + 1}})^{\rm T}}{A^{k + 1}}) + \rho B_{(3)}^{\rm T} + CB_{(3)}^{\rm T} = K_{(3)}^{\rm T}
  \label{eq:solutionB5}
\end{equation}
where $C = \beta [({I_{{n_2}}} \otimes D_1^{\rm T}{D_1}) + (D_2^{\rm T}{D_2} \otimes {I_{{n_1}}})]$ has a structure with circulant blocks.
 And the 2-D FFT and SVD are utilized to diagonalize the symmetric matrix ${({A^{k + 1}})^{\rm T}}{A^{k + 1}}$ and $C$, respectively, and its specific process can be referred to in Ref. \cite{kirrinnis2001fast}.
 % representing as:
 % \begin{equation}
  % C = F_2^{\rm T}{\Psi _2}{F_2}{\rm{~~ }}and{\rm{ ~~}}{({A^{k + 1}})^{\rm T}}{A^{k + 1}} = {U_2}\sum {}_2U_2^{\rm T}
  % \label{eq:solutionB6}
 % \end{equation}
 % where $F_2$ is 2-D DFT matrix.
 % It can also be renewed by the fast solution of Sylvester matrix equation:
 % \begin{equation}
% {B^{k + 1,l + 1}} = Fol{d_3}\left( {{{\left[ {F_2^{\rm T}\left( {\left( {1 \oslash {T_2}} \right) \odot \left( {{F_2}K_{(3)}^{\rm T}{U_2}} \right)} \right)U_2^{\rm T}} \right]}^{\rm T}}} \right)
  % \label{eq:solutionB7}
 % \end{equation}
 % where ${M_2} = 2\lambda (diag({\Psi _2}),diag({\Psi _2}),...,diag({\Psi _2}) + \rho ones({n_1}{n_2},r)) + {(diag({\Sigma _2}),diag({\Sigma _2}),...,diag({\Sigma _2}))^{\rm T}}$.

 For ${\cal Z}_1^{l + 1}$, ${\cal Z}_2^{l + 1}$, we solve the following problem:
\begin{equation}
\begin{array}{l}
  \arg \mathop {\min }\limits_{{{\cal Z}_1},{{\cal Z}_2}} {\left\| {{{\cal Z}_1}} \right\|_{2,1,{{\cal W}_1}}} + \frac{\beta }{2}\left\| {{\cal B}{ \times _1}{D_1} - {{\cal Z}_1} + \frac{{{\cal P}_1^l}}{\beta }} \right\|_F^2\\
  {\rm{~~~~~ ~~~          + }} {\left\| {{{\cal Z}_2}} \right\|_{1,{{\cal W}_2}}} + \frac{\beta }{2}\left\| {{\cal B}{ \times _2}{D_2} - {{\cal Z}_2} + \frac{{{\cal P}_2^l}}{\beta }} \right\|_F^2
  \end{array}
  \label{eq:solutiionZ}
\end{equation}
which can be directly solved by the following closed solutions:
\begin{equation}
\begin{array}{l}
  {\cal Z}_1^{l + 1}(i,j,:) = shrin{k_{2,1}}\left( {{{\hat {\cal Z}}_1}(i,j,:),\left| {{W_1}(i,j)} \right|\cdot\frac{1 }{\beta }} \right)\\
  {\cal Z}_2^{l + 1}(i,j,v) = shrin{k_1}\left( {{{\hat {\cal Z}}_2}(i,j,v),\left| {{{\cal W}_2}(i,j,v)} \right|\cdot\frac{1 }{\beta }} \right)
  \end{array}
  \label{eq:solutionZ1}
\end{equation}
And these threshold operators is defined as:
\begin{equation}
\begin{array}{l}
  shrin{k_{2,1}}(x,\xi ) = \left\{ {\begin{array}{*{20}{c}}
  {\frac{{{{\left\| x \right\|}_2} - \xi }}{{{{\left\| x \right\|}_2}}},if{\rm{~~ }}\xi {\rm{ < }}{{\left\| x \right\|}_2}}\\
  {0,otherwise}
  \end{array}} \right.,\\
  {\left[ {shrin{k_1}({\cal X},\xi )} \right]_{i,j,v}} = sign({x_{i,j,v}})\max (\left| {{x_{i,j,v}}} \right| - \xi ,0)
  \end{array}
  \label{eq:solutionthreshold}
\end{equation}
where ${{\hat {\cal Z}}_1} = {{\cal B}^{k + 1,l + 1}}{ \times _1}{D_1} + \frac{{{\cal P}_1^l}}{\beta }$, ${{\hat {\cal Z}}_2} = {{\cal B}^{k + 1,l + 1}}{ \times _2}{D_2} + \frac{{{\cal P}_2^l}}{\beta }$, ${W_1}(i,j) = \frac{1}{{{{\left\| {{{\hat {\cal Z}}_1}(i,j,:)} \right\|}_2} + \varepsilon }}$, ${{\cal W}_2}(i,j,v) = \frac{1}{{\left| {{{\hat {\cal Z}}_2}(i,j,v)} \right| + \varepsilon }}$. $\varepsilon$ is a small constant for avoiding the appearance of singularities.

For ${\cal P}_1^{l + 1}$, ${\cal P}_2^{l + 1}$, they can be solved by:
\begin{equation}
{\cal P}_1^{l + 1} = {\cal P}_1^l + \beta \left( {{{\cal B}^{k + 1,l + 1}}{ \times _1}{D_1} - {\cal Z}_1^{l + 1}} \right)
  \label{eq:solutionp1}
\end{equation}
\begin{equation}
  {\cal P}_2^{l + 1} = {\cal P}_2^l + \beta \left( {{{\cal B}^{k + 1,l + 1}}{ \times _2}{D_2} - {\cal Z}_2^{l + 1}} \right)
    \label{eq:solutionp2}
  \end{equation}

  3) {\bf Updating} ${\cal {T}}$:the ${\cal {T}}$-subproblem is as follows:
\begin{equation}
\begin{array}{l}
  \mathop {\min }\limits_{\cal T} \frac{1}{2}\left\| {{\cal Y} - {\cal B}{ \times _3}A - {\cal T}} \right\|_F^2 + \gamma {\left\| {{{\cal W}_{ASCE}} \odot T} \right\|_{1,{{\cal W}_S}}}\\
  {\rm{~~~~~~~~~    }} + \frac{\rho }{2}\left\| {{\cal T} - {{\cal T}^k}} \right\|_F^2
  \end{array}
  \label{eq:solutionT}
\end{equation}
which has the following solution:
\begin{equation}
{{\cal T}^{k + 1}} = shrin{k_1}({{\cal W}_{ASCE}} \odot \hat {\cal T},{{\cal W}_S} \odot \frac{\gamma }{{1 + \varepsilon }})
  \label{eq:solutionT1}
\end{equation}
where $\hat {\cal T} = \frac{{\left( {{\cal Y} - {{\cal B}^{k + 1}}{ \times _3}{A^{l + 1}} + \varepsilon {{\cal T}^k}} \right)}}{{1 + \varepsilon }}$, ${{\cal W}_S} = \frac{1}{{\left| {\hat {\cal T}(i,j,v)} \right| + \varepsilon }}$.

The pseudocode of the proposed PAM-based algorithm is summarized in Algorithm \ref{alg:modelsolutionPAM} to optimize the proposed model for detecting infrared small target.
\begin{algorithm}[htb!]
      \caption{Solution of the SDD Model Based on PAM.}
      \label{alg:modelsolutionPAM}
      \hspace*{0.02in}{\bf Input:}
      The acquired infrared sequence ${\cal Y} \in {\mathbb{R}^{{n_1} \times {n_2} \times {n_3}}}$, rank $r$, parameters ${\lambda}$, ${\beta}$, $\gamma$, $\varepsilon  = 0.01$;
\begin{algorithmic}[1] 
     \STATE {\bf Initialize}: $k=0$, $k_{max}=30$, ${l_{\max }} = 10$, ${A^0} = rand\left( {{n_3},r} \right)$, ${{\cal B}^0} = rand\left( {{n_1},{n_2},r} \right)$, ${{\cal T}^0} = {\cal P}_1^0 = {\cal P}_2^0 = 0$.\\
     %\varphi_n\bm{W}\vec{y}_n
      \WHILE{not converged and $k < {k_{\max }}$}
       \STATE Solving ${A^{k + 1}}$ by (\ref{eq:solutionA3})  \\
       \STATE {\bf Initialize}: $l=0$, ${\cal P}_1^0 = {\cal P}_2^0 = {\cal Z}_1^0 = {\cal Z}_2^0 = 0$.\\
        \WHILE{ $l < {l_{\max }}$}
        \STATE Solving ${{\cal B}^{k + 1,l + 1}}$ through (\ref{eq:solutionB7})\\
       \STATE Solving ${\cal Z}_1^{l + 1}$, ${\cal Z}_2^{l + 1}$  through (\ref{eq:solutionZ1})\\
       \STATE Solving ${\cal P}_1^{l + 1}$, ${\cal P}_2^{l + 1}$  through (\ref{eq:solutionp1},\ref{eq:solutionp2})\\
       \STATE Updating $l$:$l=l+1$\\
       \STATE Check the convergence condition:\\
        ~~~${\left\| {{{\cal B}^{k + 1,l}} - {{\cal B}^{k + 1,l - 1}}} \right\|_F}/{\left\| {{{\cal B}^{k + 1,l - 1}}} \right\|_F} < {10^{ - 5}}$ \\
      % \STATE  Updating $m$: $m  = m + 1.$\\
        \ENDWHILE
      \STATE Under ${{\cal B}^{k + 1}} = {{\cal B}^{k + 1,l}}$ \\
      \STATE Solving ${{\cal T}^{k + 1}}$ through (\ref{eq:solutionT1})\\
     \STATE  Updating $k$:$k=k+1$\\
     \STATE  Check the convergence condition:\\
       ${\left\| {{{\cal B}^k}{ \times _3}{A^k} - {{\cal B}^{k - 1}}{ \times _3}{A^{k - 1}}} \right\|_F}/{\left\| {{{\cal B}^{k - 1}}{ \times _3}{A^{k - 1}}} \right\|_F} < {10^{ - 5}}$\\
      \ENDWHILE
  \end{algorithmic} 
      \hspace*{0.02in}{\bf Output:} 
      Infrared background ${\cal F}{\rm{ = }}{\cal B}{ \times _3}A$ and small target component ${{\cal{T}}}$
      %\end{algorithmic}
  \end{algorithm}

\subsection{Detection Procedure}
\begin{figure*}[htb!]
  \centering
  \includegraphics[width=14cm]{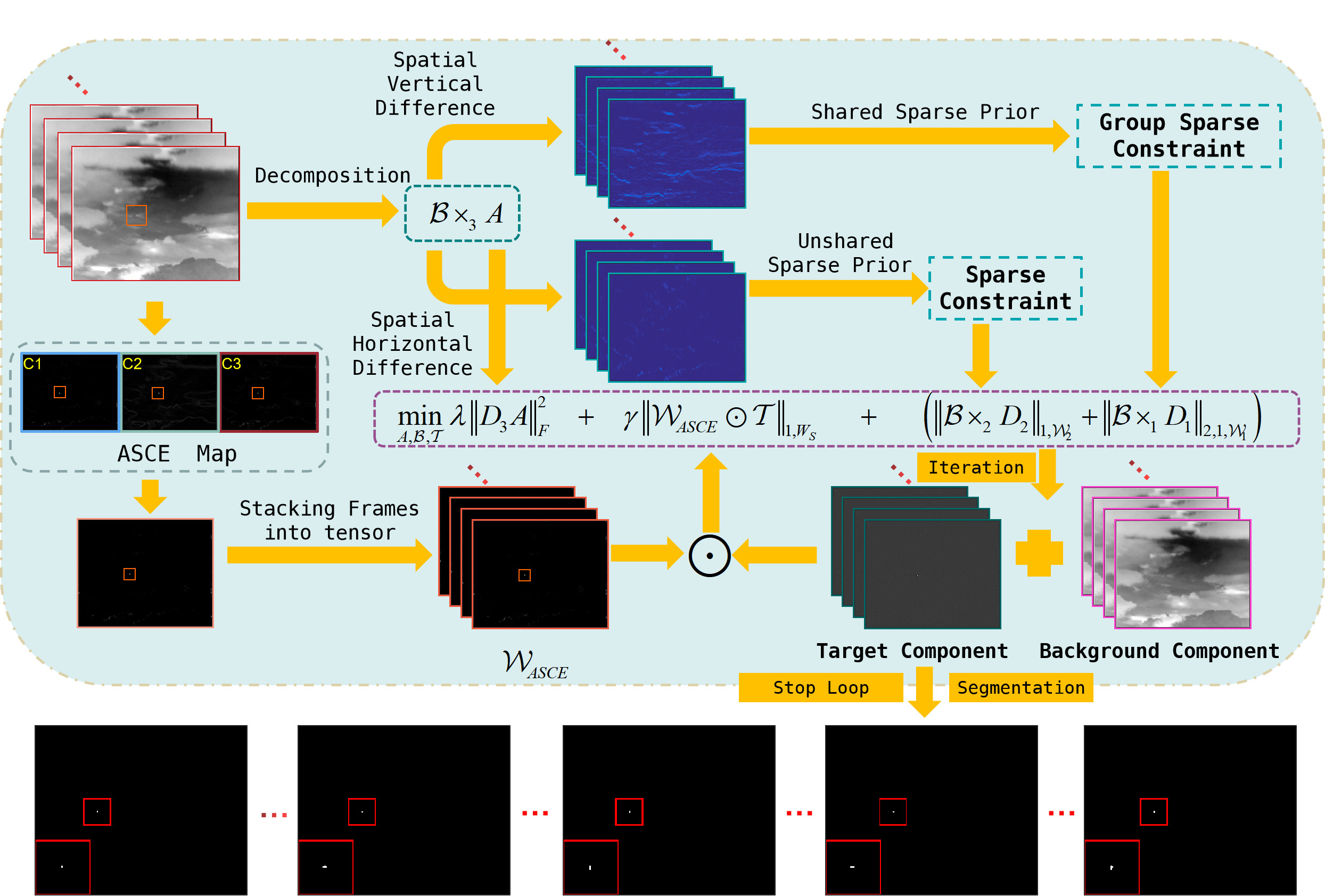} \\
  \caption{
  The overall flow-process diagram of the proposed infrared small target detection model. During detection, the target saliency enhancement factor ${{\cal W}_{ASCE}}$, obtained by stacking single frame ASCE mapping, is first constructed and then integrated into the SDD model based on PAM optimization. During decomposition, the target components from each layer are multiplied by the enhancement factor to highlight targets and suppress background clutter.
  }
\label{fig:schemtaic}
\end{figure*}

Each step of the proposed model for detecting small targets is introduced in the overall schematic, as displayed in Fig. \ref{fig:schemtaic}.
\par 1) Intercepting the infrared sequence with a fixed number of frames and taking it as an input tensor cube ${\cal Y}$.
\par 2) Calculating the $ASCE$ map of each frame of the clipping sequence cube and constructing the saliency enhancement factor ${{\cal W}_{ASCE}}$.
\par 3) Separating the small target $\cal T$ and the background components $\cal F$ via the Algorithm \ref{alg:modelsolutionPAM}.
\par 4) Employing a simple threshold operator to extracting small target from each slice of the small target component, as following:
\begin{equation}
{T_S}{\rm{ = }}\max \left( {{c_{\min }},\mu  + d\sigma } \right)
\label{threshold}
\end{equation}
where ${c_{\min }}$ and $d$ denote the given constants empirically. $\mu$ and $\sigma$ are the mean value and standard variance of each slice.

%% file: contents/experiment.tex
% !TEX root = ../main.tex
% \bibliography{../reference.bib}

\section{Experimental analysis} \label{sec:experiment}

\subsection{Experimental Settings} \label{subsec:setting}
\begin{figure*}[htb!]
  \centering
  \includegraphics[width=16cm]{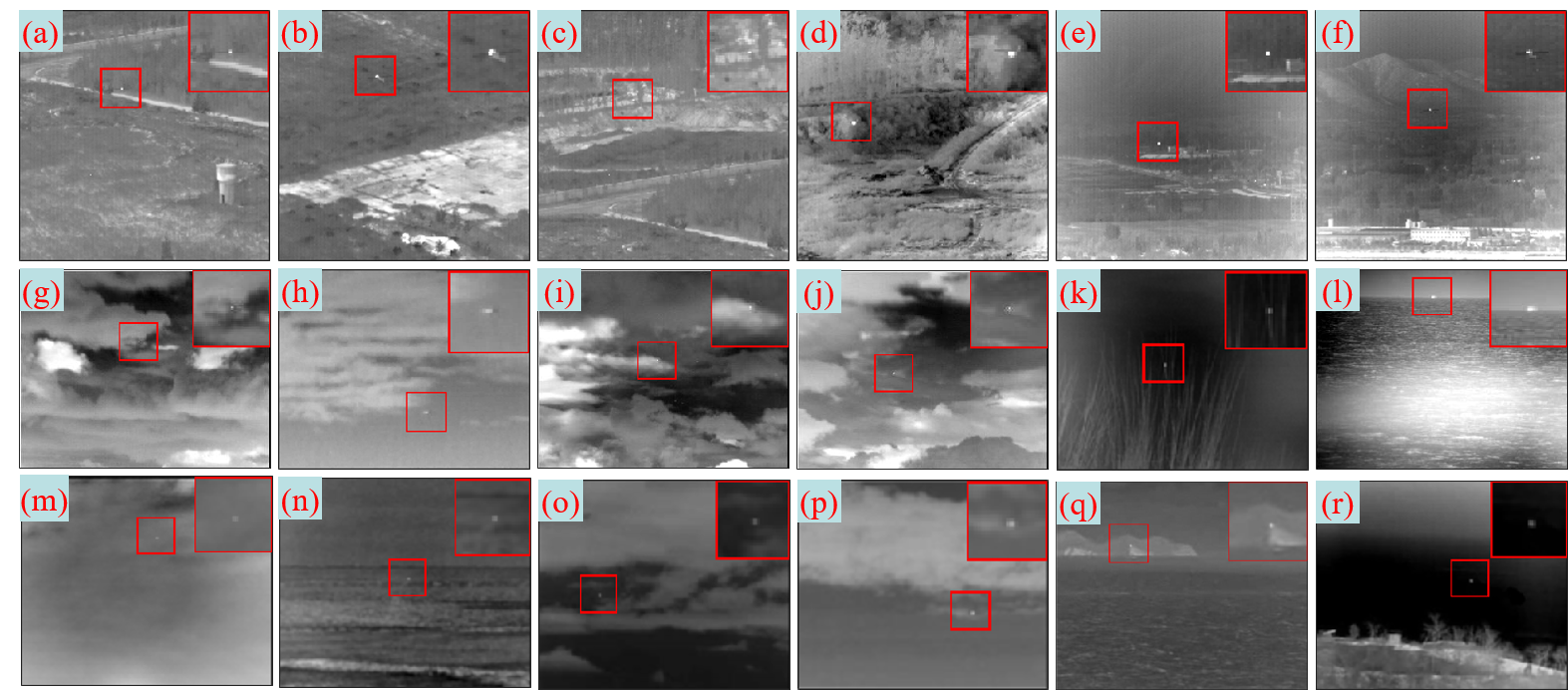} \\
  \caption{Exhibition of the 18 representative scenes.}
\label{fig:rrealscenes}
\end{figure*}

{\textit {Datasets}}:
The proposed model undergoes evaluation on an extensive array of real infrared sequences, encompassing both public and private datasets that span diverse scenes. 
Due to space constraints, we showcase only 18 representative sequences in Fig. \ref{fig:rrealscenes}. These sequences cover various scenarios, with backgrounds ranging from flat to complex and targets from prominent to dim. Considering that the greatest challenge for current detection algorithms is detecting dim targets from strong clutter, detection performance in extremely complex scenarios will be more convincing than in simple and uniform scenarios. Sequences $a$-$f$, furnished by Hui et al. \cite{Hui2020A}, feature air-ground scenes. Sequences $g$-$j$, $m$-$r$, collected by our research group, display sky-cloud, deep-space and sea-air scenes. Sequence $k$, provided by Wang et al. \cite{Wang2017Infrareddim}, presents a ground-air scene, while sequence $l$, contributed by our collaborator, exhibits a scene with sea clutter and fish scale light. A summary of the detailed information pertaining to the showcased scenes can be found in Table \ref{tab:realsce}.

\begin{table*}
\tabcolsep=0.1cm
  \caption{Detailed features of these presented infrared sequences}
  \begin{tabular*}{\textwidth}{c@{\extracolsep{\fill}}ccccc}
  % \vspace{-1.0em}
  \toprule
   NO. & Frames & Resolution & Target Features & Background Features \\
  \midrule Sequence $a$-$f$ & 400 per sequence & $256\times256$ & \makecell{ Large changes in brightness and scale,\\ relatively high contrast} & \makecell{ With numerous surface disturbances and \\nature noise, quite uneven} \\
  % \midrule
  Sequence $g$-$k$ & 50 per sequence & $320\times256$, $200\times256$, & \makecell{Punctate structure with tiny size,\\ extremely dim with low contrast}& \makecell{With numerous cloud clutters, brightness \\cloud edges, atmospheric interference} \\
  % \midrule
  Sequence $l$ & 200 & $256\times320$ & \makecell{ Saliency and size varying greatly,\\ target edge blur}  & \makecell{With brighter sea surface reflected \\light and fish scale light, strong waves}\\
  % \rowcolor{yellow}
   Sequence $m$-$r$ & 100 per sequence & \makecell{$256\times200$, $300 \times 256$, \\ $520\times 430$ }& \makecell{Weak saliency and irregular size,\\ low signal to clutter ratio}  & \makecell{Significant background changes \\uneven ground surface, bright glitters.}\\
  \bottomrule
  \end{tabular*}
  \label{tab:realsce}
  \end{table*}

  {\textit {Baselines and Parameter Settings}}:
 In the experiments, we juxtapose the proposed method with 14 state-of-the-art solutions, assessing both objective indicators and visual effects. These approaches encompass 6 single-frame detection methods: Weighted Local Difference Measure (WLDM) \cite{deng2016small}, Multiscale Patch-based Contrast Measure (MPCM) \cite{Wei2016Multiscale}, Feature Kernel-based Random Walk (FKRW) \cite{Qin2019Infrared}, Infrared Patch-Image model (IPI) \cite{Cao2013Infrared}, Reweighted Infrared Patch-Tensor model (RIPT) \cite{Dai2017Reweighted}, and Dynamic Weight-guided Smooth-Sparse Decomposition (DWSSD) \cite{Dan2024Dynamic}, and 8 multi-frame detection methods: Spatial-Temporal Saliency Map (STSM) \cite{Li2016A}, spatial-temporal tensor modeling with Saliency Filter Regularization (SFR) \cite{Pang2022STTM}, Tensor Completion with Top-Hat model (TCTH) \cite{Zhu2020Infrared}, Edge and Corner Awareness-based spatial-temporal tensor model (ECA) \cite{Zhang2020Edge}, Non-convex Tensor Low-rank Approximation (NTLA) \cite{liu2022Nonconvex}, Tensor Spectral k-support model (TSPK) \cite{zhou2023infrared}, Sparse Regularization-based Spatial–Temporal Twist
tensor (SRSTT) \cite{Li2023Sparse} and 4-D Tensor Ring (4DTR) \cite{Wu2023Infrared}. To ensure equitable comparisons, we employ the original codes provided by the authors for FKRW, IPI, RIPT, ECA, NTLA, TSPK, 4DTR, SRSTT and DWSSD, adhering to the recommended parameter settings delineated in their respective papers. For the remaining methods, we re-implement them based on the corresponding references. A summary of all methods and their parameter settings can be found in Table \ref{tab:parset}.

  \newcommand{\tabincell}[2]{\begin{tabular}{@{}#1@{}}#2\end{tabular}}
\begin{table*}[htb!]
  \tabcolsep=0.01cm
\caption{Parameter setting summary by finely adjusting the tested methods}
\begin{tabular*}{\textwidth}{c@{\extracolsep{\fill}}lll}
\toprule
No. & Methods & Parameter Setting \\
\midrule
1 & WLDM \cite{deng2016small}  &$L=4,m=2,n=2$. \\
2 & MPCM \cite{Wei2016Multiscale} & Mean filter size: $3\times3$, local window size:$N = 3,5,7,9$. \\
3 & FKRW \cite{Qin2019Infrared}    & $K = 4, p = 6$, $\beta = 200$, window size: $11\times11$. \\
4 & IPI \cite{Cao2013Infrared}  &Patch size: 50$\times$50,sliding size: 10,$\lambda {\rm{ = }}L{\rm{/min(m,n}}{{\rm{)}}^{1/2}},{\rm{ }}L \in \left[ {1,{\rm{5}}} \right]$, $\varepsilon {\rm{ = 1}}{{\rm{0}}^{{\rm{ - 7}}}}$.    \\
5 & RIPT \cite{Dai2017Reweighted}  &Patch size: 50$\times$50, sliding size: 10, $\lambda {\rm{ = }}L{\rm{/min(}}I{\rm{,}}J{\rm{,}}P{{\rm{)}}^{1/2}},{\rm{ }}L \in \left[ {{\rm{0}}{\rm{.1}},{\rm{2}}} \right]$, $h = 10$, $\epsilon$=0.01, $\varepsilon {\rm{ = 1}}{{\rm{0}}^{{\rm{ - 7}}}}$.   \\
6 & STSM \cite{Li2016A} &Patch size:4, overlap:2, local neighboring samples:8, $\lambda  = {10^7}$, frame number: 5 \\
7 & SFR \cite{Pang2022STTM}  & Filter size:$5\times5$, ${\sigma _1} = 0.8$, ${\sigma _2} = 2.0$, $t=5$, $p=0.9$, $\epsilon=10^{-8}$, $\lambda  = L/\left( {\min {{\left( {m,n} \right)}^{1/2}} \times t} \right)$, $L \in \left[ {2,5} \right]$. \\
8 & TCTH \cite{Zhu2020Infrared}   & $p = 0.1$, $\varepsilon  = {10^{ - 7}}$, $s2B = s{B_O} = 5 \times 5$, $s1B = 3 \times 3$, frame number: 10, $\beta {\rm{ = 1/}}\sqrt {\min (I,J,P)}$ , $L \in \left[ {{\rm{2,}}10} \right]$, $\alpha=0.001$. \\
9 & ECA \cite{Zhang2020Edge} & $\beta  = 0.5$, $t=3$, ${\lambda _1} = 0.005$, ${\lambda _2} = L/\left( {\min {{\left( {m,n} \right)}^{1/2}} \times t} \right)$, $L \in \left[ {0.1,2} \right]$.  \\
10 & NLTA \cite{liu2022Nonconvex} & $L=3$, $H \in \left[ {5,10} \right]$, ${\lambda _{TV}} = 0.005$, ${\lambda _S} = H/\left( {\max {{\left( {m,n} \right)}^{1/2}} \times L} \right)$, ${\lambda _3} = 100$.  \\
% \rowcolor{yellow}
11 & TSPK \cite{zhou2023infrared} & frame number: 10, patch size: 10$\times$10, spectral factor: 5, $\lambda \in \left[1.5,2 \right]$, $\beta \in \left[0.5,5 \right]$, $\varepsilon {\rm{ = 5\times 1}}{{\rm{0}}^{{\rm{ - 7}}}}$. \\
% \rowcolor{yellow}
12 & 4DTR \cite{Wu2023Infrared} & Patch size:${N_1} \times {N_2}:70 \times 70$, temporal size:${N_3} = 15$,${\lambda _1} = \sum\nolimits_{i = 1}^l {{2 \mathord{\left/
 {\vphantom {2 {\sqrt {\max \left( {\prod _{i = n}^{n + l - 1}{N_i},\prod _{i = n + l}^{n - 1}{N_i}} \right)} }}} \right.
 \kern-\nulldelimiterspace} {\sqrt {\max \left( {\prod _{i = n}^{n + l - 1}{N_i},\prod _{i = n + l}^{n - 1}{N_i}} \right)} }}}$, $\tau  = 1 \times {10^{ - 7}}$.  \\
 % \rowcolor{yellow}
13 & SRSTT \cite{Li2023Sparse} & $L=30$, ${\lambda _1} = 0.05$, ${\lambda _2} = 0.1$, ${\lambda _3} = 100$,$\varepsilon {\rm{ = 1}}{{\rm{0}}^{{\rm{ - 7}}}}$,$\mu = 0.01$.  \\
% \rowcolor{yellow}
14 & DWSSD \cite{Dan2024Dynamic} & $L=0.015$, $\beta = 100\lambda$, ${\lambda} = L/\left( {\max {{\left( {m,n} \right)}^{1/2}} } \right)$, $k = 5$.  \\
% \rowcolor{yellow}
15 & SDD   & \makecell{Frame number: 30, $R=10$, $\lambda  \in \left[ {0.5,5} \right]$, $\gamma  \in \left[ {0.1,1} \right]$, outer iteration:50, inner iteration:10, $\beta  = 15000$, $\mu = 0.05$, $\varepsilon = 0.01$.} \\
\bottomrule
\end{tabular*}
\label{tab:parset}
\end{table*}

{\textit {Evaluation Indicators}}:
To provide an objective comparison of the performance among the tested methods, we utilize four evaluation indices. These indices include background suppression factor ($BSF$), signal to clutter ratio gain ($G_{SCR}$), contrast gain ($CG$), and receiver operating characteristic curve (ROC). The $BSF$ quantifies the suppression effect in the vicinity of the small target's neighboring area and is defined as follows:
\begin{equation}
  BSF = \frac{{{\sigma _{in}}}}{{\left( {{\sigma _{out}}{\rm{ + }}\omega } \right)}}
  \label{exp:2}
  \end{equation}
where ${\sigma _{in}}$ and ${\sigma _{out}}$ are the standard deviation of target neighboring region of original and suppressed images, separately; $\omega {\rm{ = 0}}.{\rm{01}}$ is a smoothing factor to avoid division zero.
$G_{SCR}$ measures the improvement in signal-to-clutter ratio before and after target separation and can be defined as:
  \begin{equation}
   {G_{SCR}} = \frac{{ou{t_{SCR}}}}{{i{n_{SCR}}}}
    \label{exp:1}
    \end{equation}
  where $in_{SCR}$ and $out_{SCR}$ are the $SCR$ values before and after target separation, separately, and $SCR = \left| {{M_t} - {\mu _b}} \right|/\left( {{\sigma _b}{\rm{ + }}\omega } \right)$.
  $M_t$ denotes the maximum intensity of the target area. 
  ${\mu _b}$ and ${\sigma _b}$ are the average grayscale and standard deviation of the neighboring area around small targets.
  $CG$ measures the contrast between the separated target and its local neighborhood and can be defined as:
  \begin{equation}
   CG = \frac{{CO{N_{out}}}}{{CO{N_{in}}}}
    \label{exp:3}
  \end{equation}
  where $CO{N_{in}}$ and $CO{N_{out}}$ represent the target local contrast before and after small target separation, respectively, and $CON = \left| {{M_t} - {\mu _b}} \right|$.

  Herein, we assume that the target size is $a\times b$, and the neighboring area is defined as $\left( {a + 2d} \right) \times \left( {a + 2d} \right)$, with $d$ set to 30 as the neighborhood width. 
  % The performance of the examined methods is assessed using three evaluation indices, namely $BSF$, $G_{SCR}$, and $CG$. 
The larger values of the evaluation indicators corresponds to superior method performance.

  The ROC curve depicts the dynamic range between the probability of detection ($P_b$) and the false alarm rate ($F_a$). $P_b$ and $F_a$ can be defined as follows:
  \begin{equation}
    {P_d} = \frac{{{\rm{number~ of~ true~ detections}}}}{{{\rm{number~ of~ actual~ targets}}}}
    \label{exp:4}
    \nonumber
    \end{equation}
  \begin{equation}
    {F_a} = \frac{{{\rm{number~ of~ false~ detections}}}}{{{\rm{number~ of~ images}}}}\nonumber
    \label{exp:5}
    \end{equation}
In ROC curves, we regard small targets as detected if pixels exist within a $5\times 5$ window surrounding the ground truth. 
Moreover, we compute the area under the curve (AUC) of the ROC curve to further appraise the detection performance. A larger AUC signifies superior detection performance.

\subsection{Model Discussion} \label{subsec:ablation}
\subsubsection{\textit {Inner Iteration Analysis}} 
The number of inner iterations, denoted as $l$, influences not only the detection performance but also the computational complexity of the proposed model. To evaluate its impact, we vary $l$ from 2 to 20 under scenes $a$ and $g$, presenting the AUC and average computational time at a specific $l$ in Table \ref{tab:inneriter}. It is observed that the proposed model attains stable and superior performance when $l \ge 12$. However, as the number of iterations escalates, the computational time increases significantly, while the performance remains unaltered. Consequently, we set the number of inner iterations to 12 in subsequent experiments to strike a balance between detection performance and computational complexity.
\begin{table*}[htb!]
\caption{Detailed Information of The Candidate Target Sets and The Original Real Backgrounds}
\begin{tabular*}{\textwidth}{c@{\extracolsep{\fill}}cccccccccccc}
\toprule
Datasets & $l$ & 2 & 4	& 6 &	8	& 10	& 12	& 14	& 16	& 18	& 20       \\
\midrule
\multirow{2}*{\shortstack{scene $a$}} & AUC  &4.68	& 4.74	& 4.78	& 4.80	& 4.82	& 4.83	& 4.83	& 4.83	& 4.83	& 4.83  \\
% \midrule[2-5]
&Times  & 0.33	& 0.60	& 0.94	& 1.42	& 1.82	& 2.74	& 3.18	& 3.57	& 4.00	& 4.46   \\
\midrule
\multirow{2}*{\shortstack{scene $b$}}& AUC  &1.24	&1.24	&1.26	&1.28	&1.30	&1.31	&1.30	&1.31	&1.31	&1.31 \\
&Times  & 0.29	& 0.61	&1.12	&1.95	&2.40	&2.71	&3.29	&3.54	&4.03	&4.39   \\
\bottomrule
\end{tabular*}
\label{tab:inneriter}
\end{table*}

\subsubsection{\textit {Model Parameter Analysis}} 
\begin{figure*}[htb!]
  \centering
  \subfigure[]{
  \begin{minipage}[b]{0.28\textwidth}
     \includegraphics[width=4.90cm]{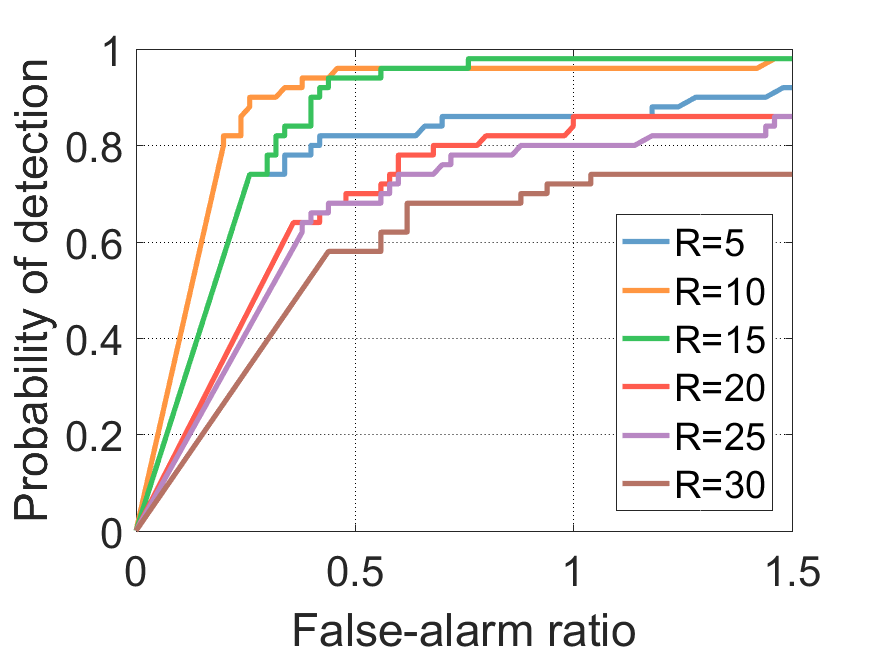}
  % \label{1a}
  \end{minipage}}
  \subfigure[]{
\begin{minipage}[b]{0.28\textwidth}
      \includegraphics[width=4.90cm]{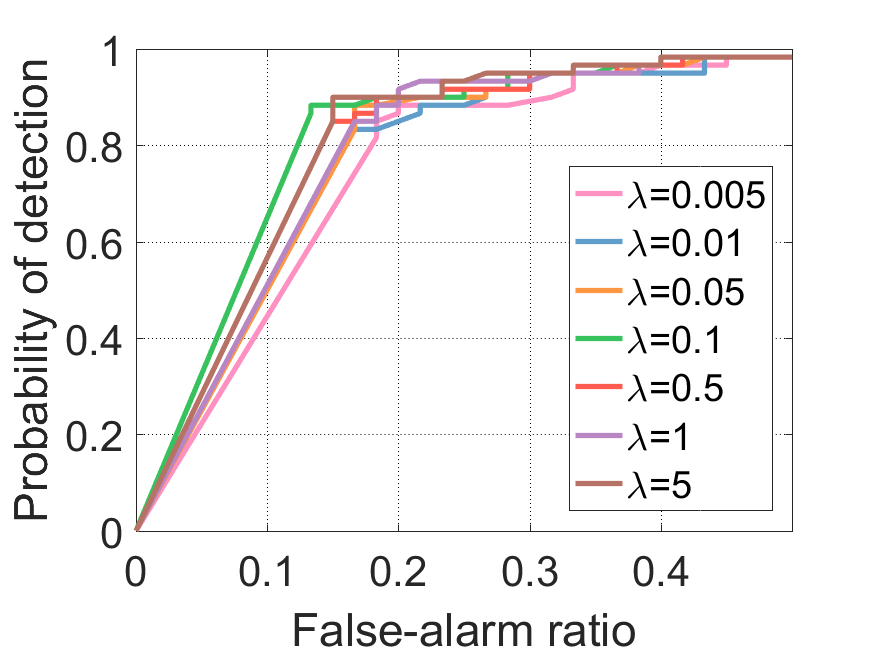}
  % \label{1c}
  \end{minipage}}
  \subfigure[]{
\begin{minipage}[b]{0.28\textwidth}
      \includegraphics[width=4.90cm]{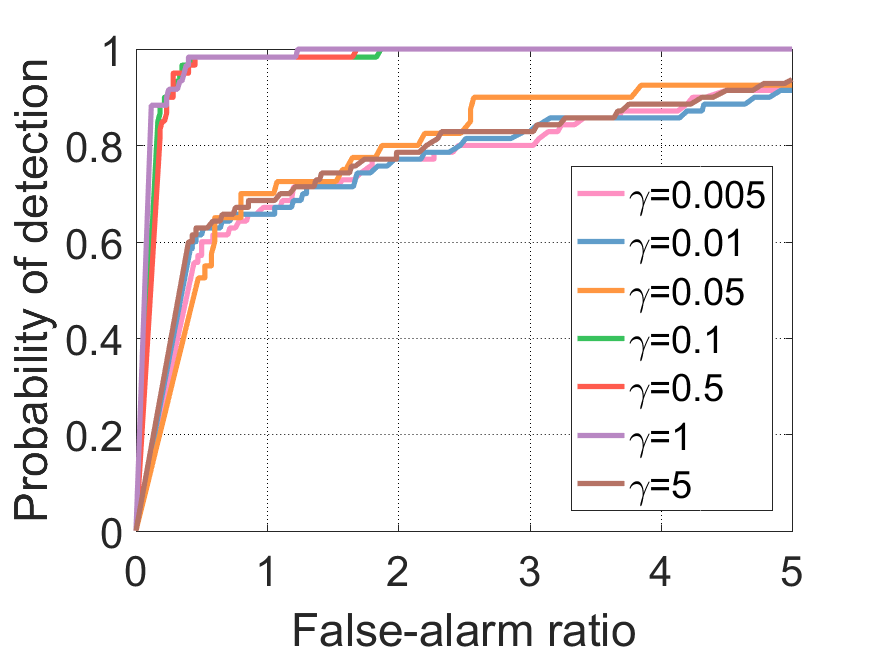}
      % \label{1d}
        \end{minipage}
        }
   \vskip -10pt
  \subfigure[]{
\begin{minipage}[b]{0.28\textwidth}
     \includegraphics[width=4.90cm]{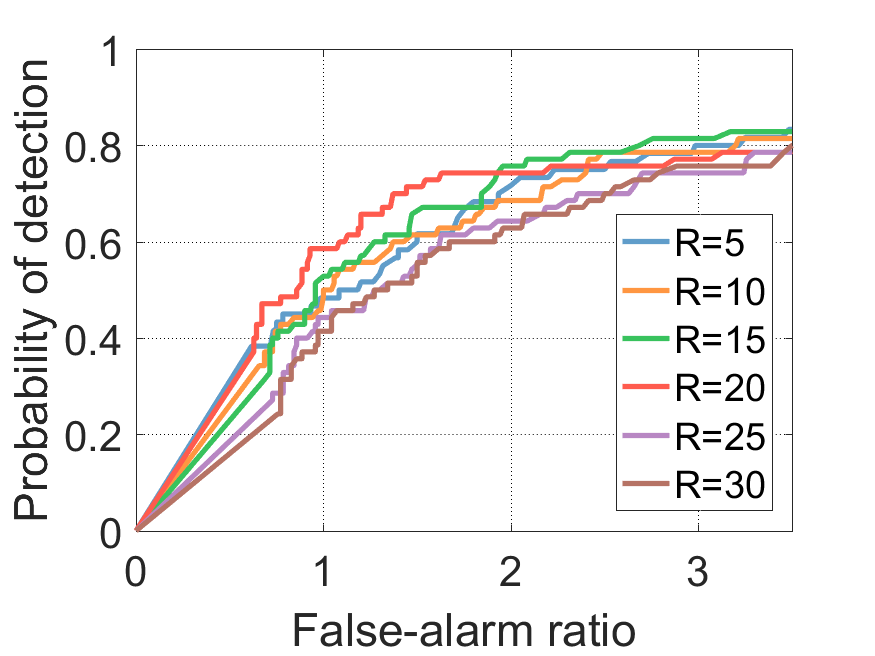}
  % \label{1a}
  \end{minipage}}
%   \subfigure{
% \begin{minipage}[b]{0.23\textwidth}
%       \includegraphics[width=4.60cm]{./figures/Set40_tau.png}
%   \label{1b}
%   \end{minipage}}
  \subfigure[]{
\begin{minipage}[b]{0.28\textwidth}
      \includegraphics[width=4.90cm]{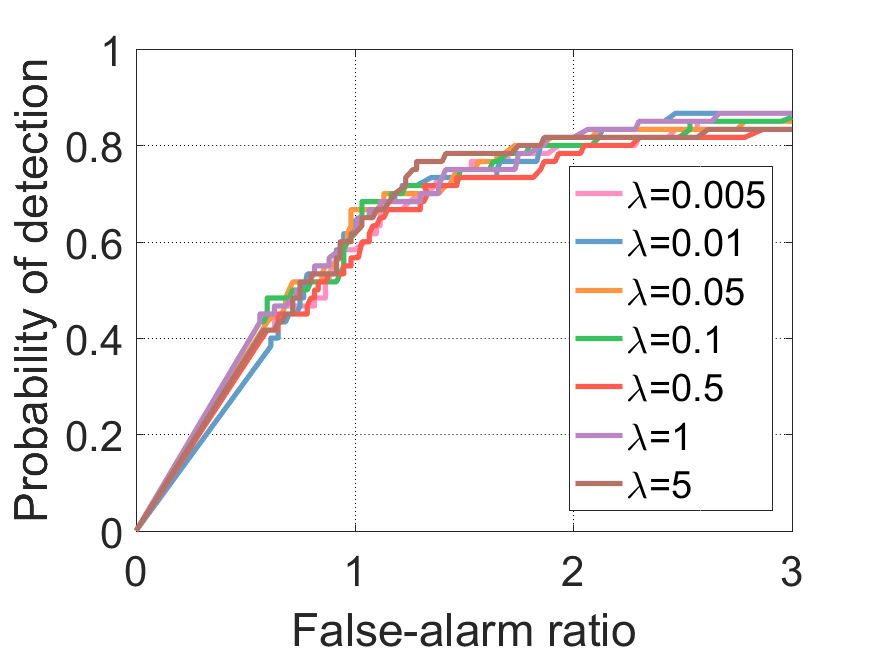}
  % \label{1c}
  \end{minipage}}
  \subfigure[]{
\begin{minipage}[b]{0.28\textwidth}
      \includegraphics[width=4.90cm]{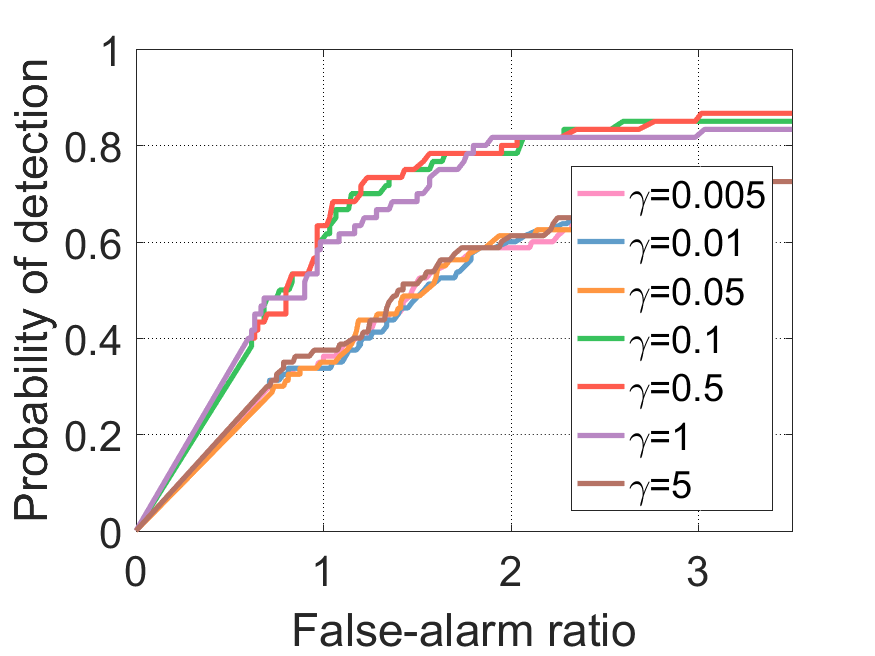}
      % \label{1d}
        \end{minipage}
        }
   \vskip -5pt
  \caption{Sensitivity analysis of the involved parameters by using the changing ROC curves under scenes (a) and (g).}
\label{fig:paraana}
\end{figure*}

The objective function of the proposed model involves four key parameters: the rank parameter $R$, continuity constraint factor $\lambda$, and sparse penalty $\mu$. The selection of these parameters significantly impacts the performance of the proposed model. We subsequently investigate the influence of these parameters on typically complex scenes $a$ and $g$. Fig. \ref{fig:paraana} illustrates the changes in the ROC curves of the two scenes when varying the parameter values.

The rank factor $R$ primarily serves to measure the inter-frame correlation of infrared sequence cubes. The sensitivity of the proposed model's performance to changes in $R$ is displayed in the first column of Fig. \ref{fig:paraana}. It is evident that when $R$ is set to 10 or 15, the proposed model achieves stable and superior performance. However, the model's performance diminishes when $R$ assumes other values. Furthermore, a larger value of $R$ engenders higher computational complexity. Considering these factors, we set $R$ to 10 in our experiments.

The parameter $\lambda$ governs the temporal continuity constraint. The second column of Fig. \ref{fig:paraana} demonstrates the changes in the ROC curves induced by varying the factor. The proposed model's behavior exhibits robustness to the parameters, with the ROC curves displaying a similar changing tendency under different scenes and only minor performance differences observed. Consequently, we set $\lambda$ to 1.

The sparse penalty $\gamma$ plays a pivotal role in controlling target sparsity due to its threshold attribute. A smaller value of $\gamma$ results in the retention of some non-target components, while a larger value may lead to the loss of small targets. Therefore, striking an appropriate balance between detection probability and false alarm necessitates the careful selection of $\gamma$. Fig. \ref{fig:paraana}'s third column presents the ROC curves when varying $\gamma$ from 0.005 to 5. It becomes apparent that the detection performance of the model deteriorates when the value of $\gamma$ is excessively large or small. This outcome arises from the prevalence of high false alarms or low detection rates in these cases. When the value range is set between 0.1 and 1, the proposed model achieves stable and superior detection performance.

\subsubsection{\textit {Ablation Experiment}}
\begin{figure*}[htb!]
  \centering
  \subfigure[]{
  \begin{minipage}[b]{0.28\textwidth}
     \includegraphics[width=4.90cm]{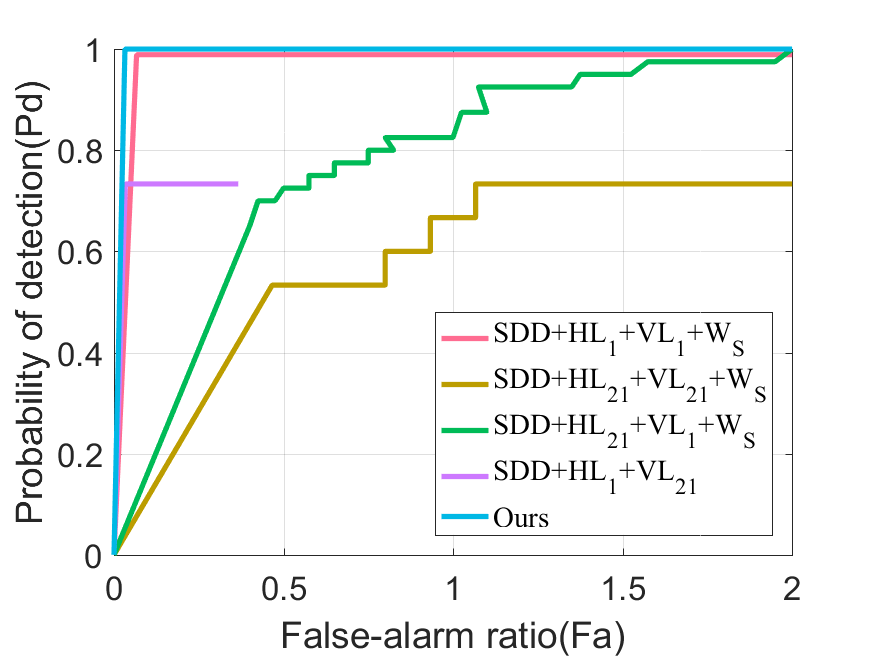}
  % \label{1a}
  \end{minipage}}
  \subfigure[]{
\begin{minipage}[b]{0.28\textwidth}
      \includegraphics[width=4.90cm]{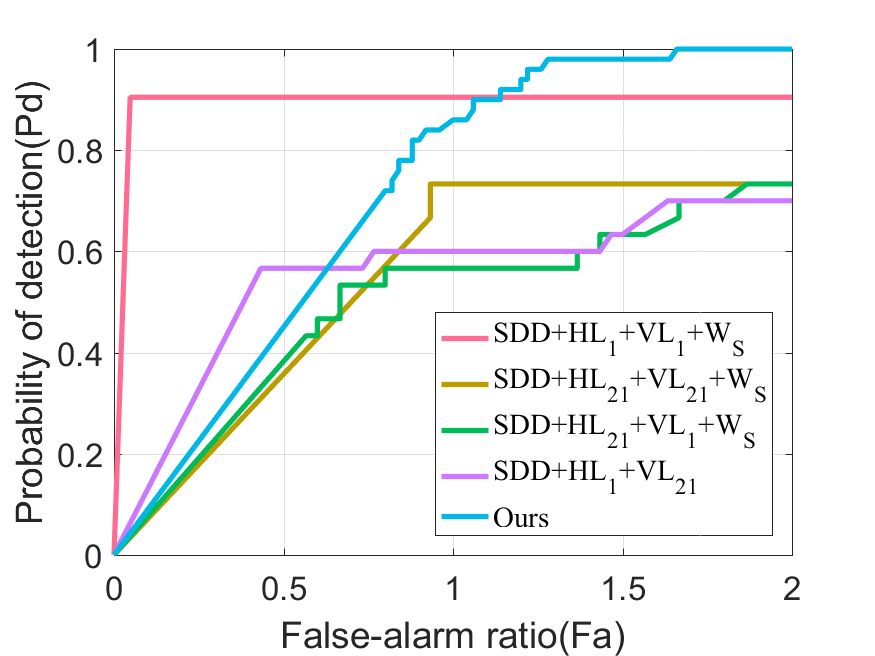}
  % \label{1c}
  \end{minipage}}
  \subfigure[]{
\begin{minipage}[b]{0.28\textwidth}
      \includegraphics[width=4.90cm]{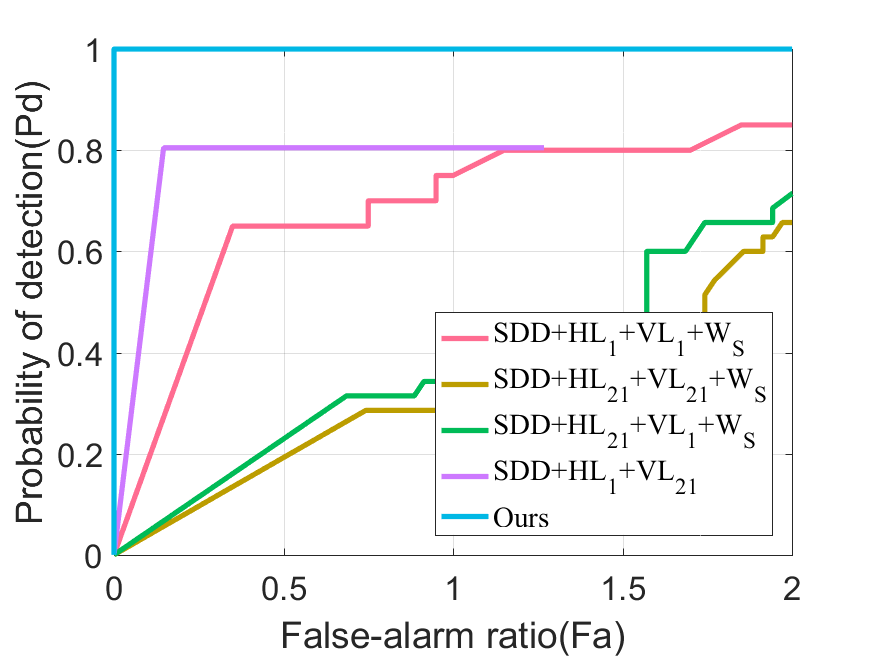}
      % \label{1d}
        \end{minipage}
        }
   \vskip -5pt
  \caption{Ablation studies with different assembles settings under Scenes $b$, $c$ and $f$ by the changing ROC curves. SDD+H$L_1$+V$L_1$+$W_S$: Nonshared $l_1$ constraints are used for both spatial horizontal and vertical modes. SDD+H$L_{21}$+V$L_{21}$+$W_S$: Shared $l_{21}$ constraints are used for both spatial horizontal and vertical modes. SDD+H$L_{21}$+V$L_1$+$W_S$: Shared $l_{21}$ constraint is used for the spatial horizontal mode while nonshared $l_1$ constraint for the vertical mode. SDD+H$L_1$+V$L_{21}$: Using the proposed constraint strategy without significant enhancement factors.}
\label{fig:ablexp}
\end{figure*}

In order to analyze the effectiveness of different spatial constraint configurations, we conduct a set of ablation experiments, as shown in Fig. \ref{fig:ablexp}. From the subfigures, it can be seen that using appropriate constraints for differential priors in different spatial directions can significantly improve the detection performance of the model. Especially, ignoring differential prior sparse patterns in different spatial directions and using the same constraints will affect the detection performance of the model. For example, in scene $b$ and $f$, the $P_d$ of SDD+H$L_1$+V$L_1$+$W_S$ and SDD+H$L_{21}$+V$L_{21}$+$W_S$) significantly decrease when the false alarms are the same. Furthermore, it can be observed that the model's performance is better when sparse constraints are employed concurrently, compared to the simultaneous application of group sparse constraints. This is primarily attributed to the inability of group sparse constraints to effectively suppress interference that resembles the target point. The stability of the SDD+H$L_{21}$+V$L_1$+$W_S$ model is inadequate, as evidenced by its strong performance in scenario b and poor performance in the other two scenarios. Owing to the absence of saliency enhancement factors, the target components of the SDD+H$L_1$+V$L_{21}$ model are prone to being lost during the decomposition process, thereby impeding further improvements in the target detection rate.

The experiment demonstrates that the proposed model can effectively suppress background clutter, highlight targets, and enhance robustness by appropriately allocating sparse and group sparse constraints, while integrating elaborately designed saliency enhancement factors.

\subsubsection{\textit {Convergence Analysis}}
\begin{figure}[htb!]
  \centering
  \subfigure[]{
%  \begin{minipage}[b]{0.23\textwidth}
     \includegraphics[width=4.4cm]{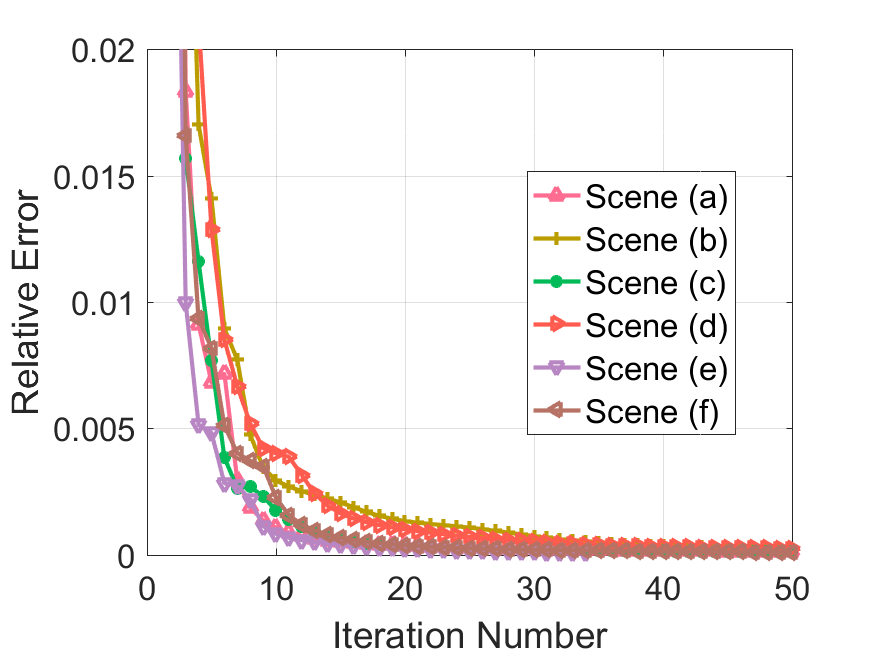}%}
  % \label{1a}
%  \end{minipage}
}\hspace{-5mm}
  \subfigure[]{
% \begin{minipage}[b]{0.23\textwidth}
      \includegraphics[width=4.4cm]{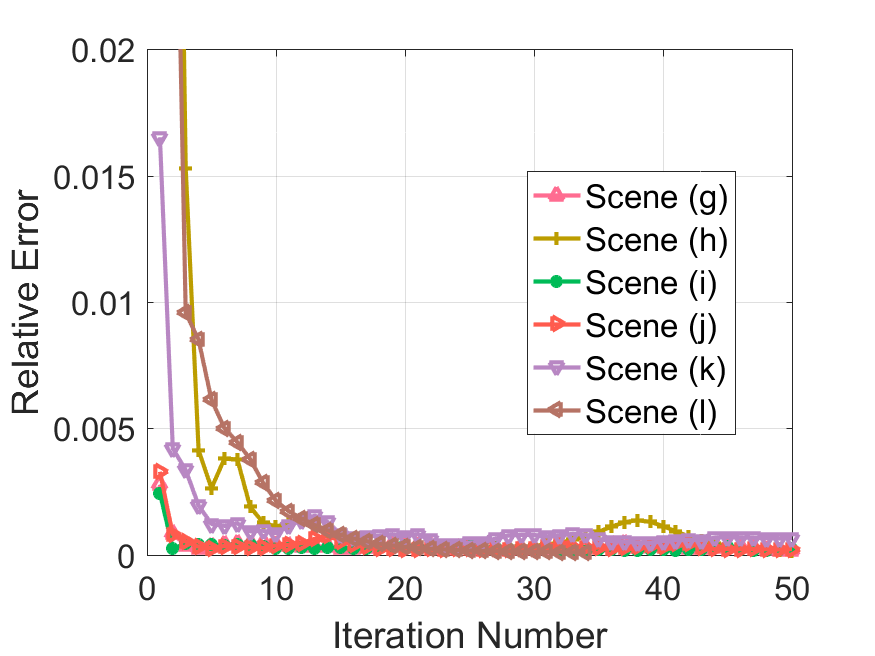}%}
  % \label{1b}
%    \end{minipage}
}
   \label{1d}
   \vskip -5pt
\caption{Relative error change curves of the proposed model along with the number of iteration under different scenes.}
\label{fig:convergence}
\end{figure}

Owing to the non-convexity of the proposed model, existing convex solvers are rendered inapplicable. Consequently, the ADMM and reweighted strategy are incorporated into the PAM-based optimization framework. 
% Although the solution of the proposed model can be attained by alternately updating each variable, establishing its convergence theoretically remains a challenge.
To empirically verify the convergence of the proposed algorithm, we employ the relative error change as a criterion. Fig. \ref{fig:convergence} presents the relative error change curves of the proposed algorithm concerning the iteration number across twelve distinct scenes. Upon examining the curves, it becomes apparent that the decrement values of the relative error gradually approach the convergence criterion after a sufficient number of iterations for all tested scenes. This observation numerically demonstrates the effective convergence of the solution.

\subsection{Qualitative Evaluation of The Proposed Method} \label{subsec:sota}
\subsubsection{Evaluation on Different Scenes}
\begin{figure*}[htb!]
  \centering
  \includegraphics[width=16cm]{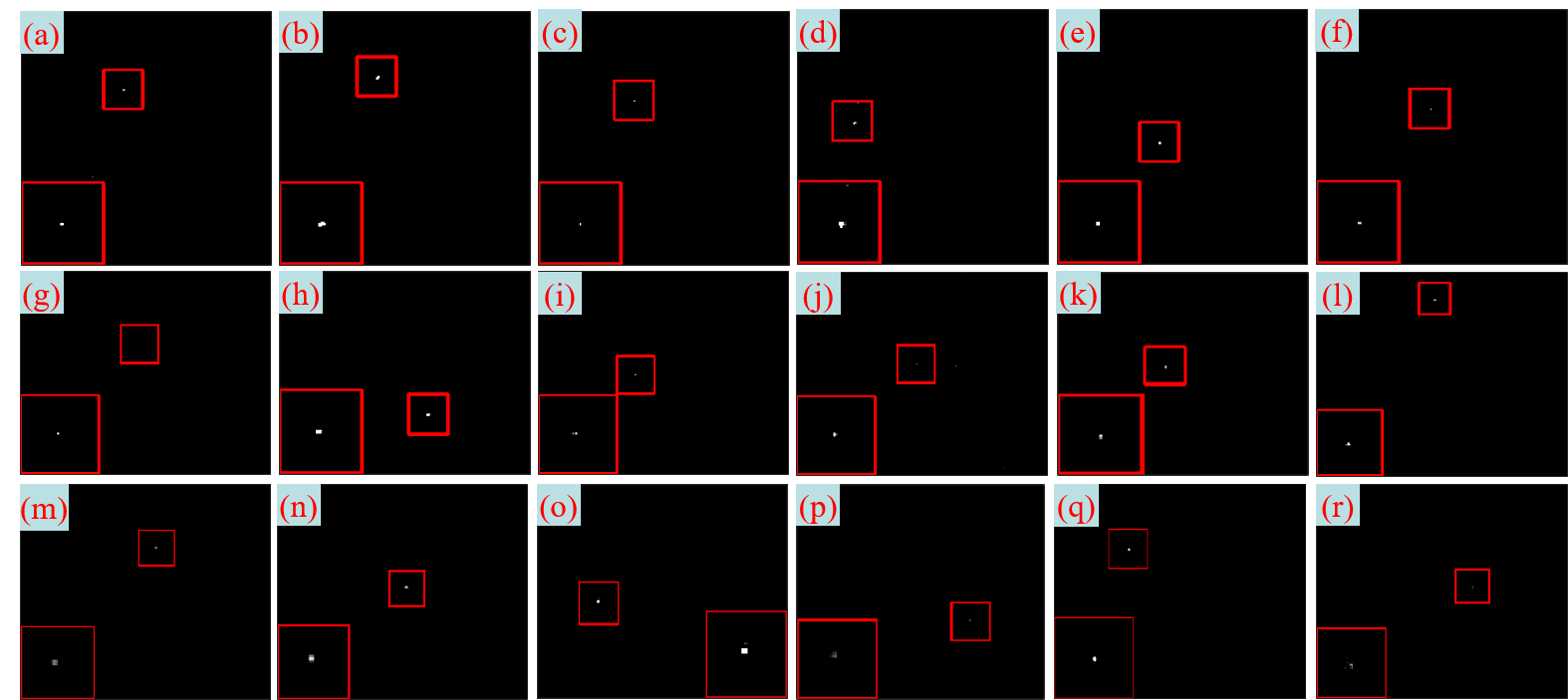} \\
  \caption{Detection results obtained by the proposed model under twelve displayed scenes.}
\label{fig:results}
\end{figure*}

\begin{figure*}[htb!]
  \centering
  \includegraphics[width=16cm]{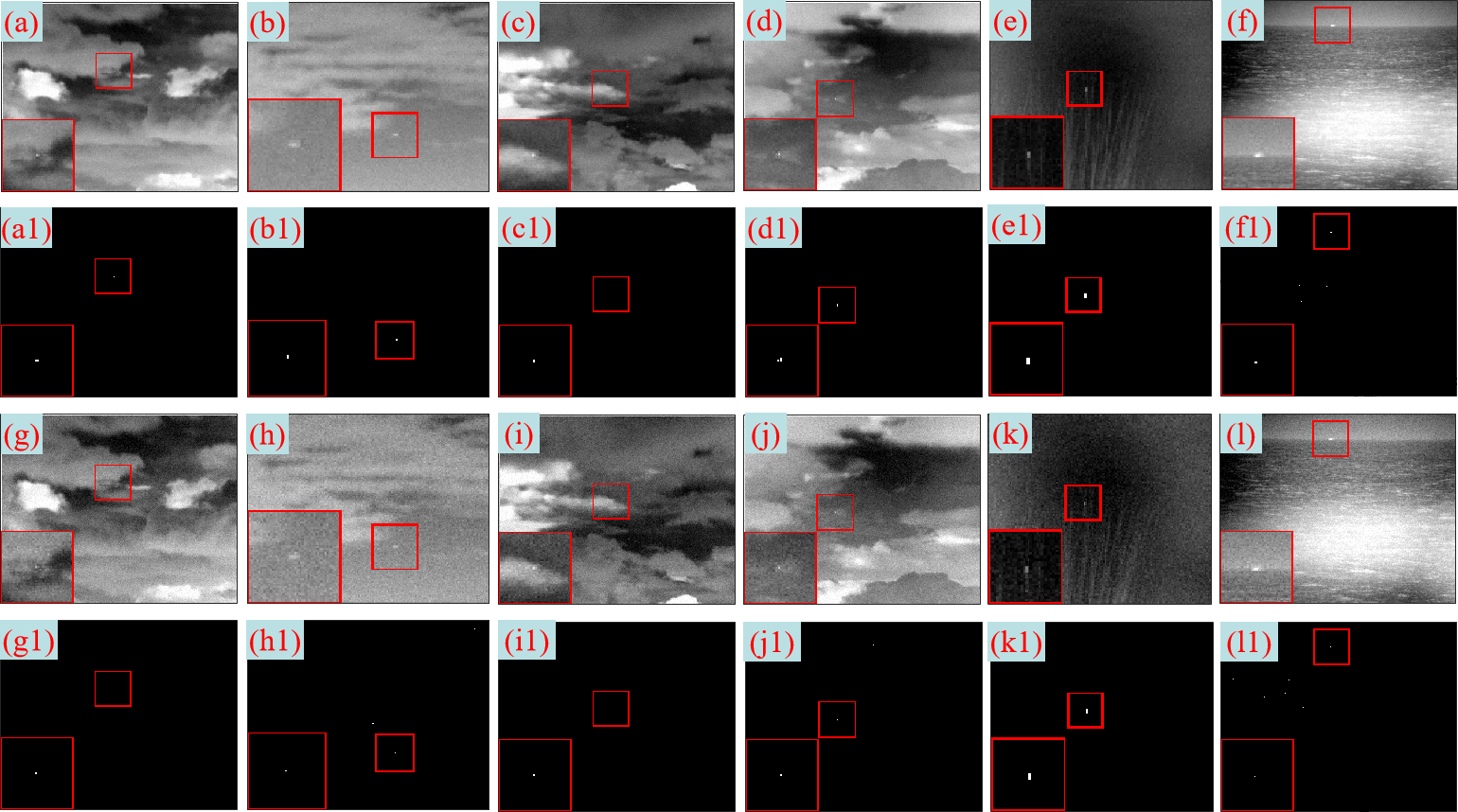} \\
  \caption{Evaluation of the proposed method on polluted scenes with additive white Gaussian noise. The first and third rows show the noise images with standard deviation of 5 and 10. The second and fourth rows show the detecting results by the proposed method.}
\label{fig:resultsnoise}
\end{figure*}
To assess the robustness of the proposed model, we examine its performance on different scenes depicted in Fig. \ref{fig:rrealscenes}. 
% These representative scenes cover diverse complex backgrounds, and the embedded targets range from high brightness and large size to dim and exceedingly small dimensions. 
%including air-ground backgrounds with surface interference, deep-space backgrounds with cloud clutter, and sea surface backgrounds with intense reflected light. 
In Fig. \ref{fig:results}, we showcase the detection results of the proposed model on these scenes. It is evident that small targets are entirely separated without any residual background clutter. These findings demonstrate that the proposed model can consistently and proficiently tackle various scenarios.

% Infrared images are susceptible to various noise sources, originating from sensors and natural factors, which can further impede target detection by reducing target brightness and contrast. 
To evaluate the noise-robust performance of the proposed method, we conduct experiments on scenes affected by different noise levels, as illustrated in Fig. \ref{fig:resultsnoise}. The first and third rows of the figure display six scenes with Gaussian noise, featuring a zero-mean and standard deviations of 5 and 10, respectively. The second and fourth rows present the detection results obtained using the proposed method.

A comparison of the detection results in the second and fourth rows reveals a substantial improvement in the former. For example, the small target in the results of noise scene $l$ is nearly indiscernible, and a false alarm point emerges in noise scene $h$. These observations suggest that detection becomes increasingly challenging as the noise level escalates. Nevertheless, the proposed method can accurately locate the target, provided it is not entirely submerged in noise.

\subsubsection{Visual Comparison with Other Competitors}
\begin{figure*}[htb!]
  \centering
  \includegraphics[width=16cm]{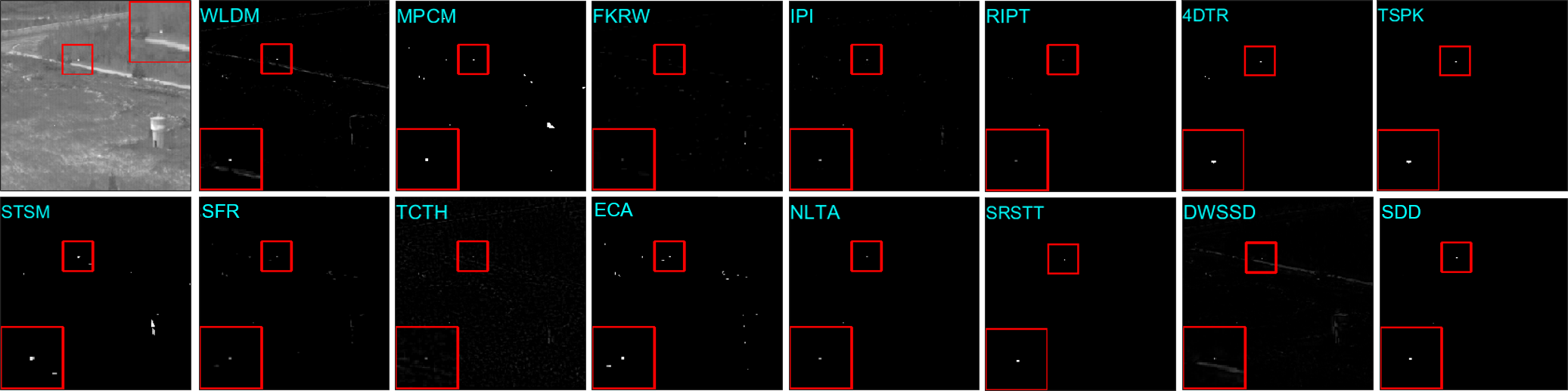} \\
  \caption{Detection results of ten comparison methods under scene $a$. The target area is zoomed in the lower left corner for better observation.}
\label{fig:com_a}
\end{figure*}

\begin{figure*}[htb!]
  \centering
  \includegraphics[width=16cm]{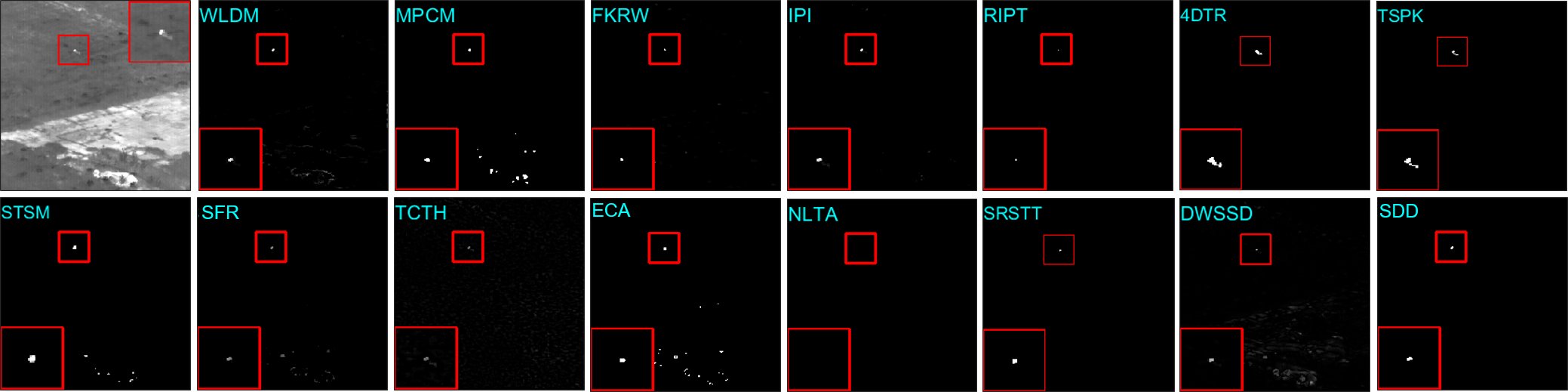} \\
  \caption{Detection results of ten comparison methods under scene $b$. The target area is zoomed in the lower left corner for better observation.}
\label{fig:com_b}
\end{figure*}

\begin{figure*}[htb!]
  \centering
  \includegraphics[width=16cm]{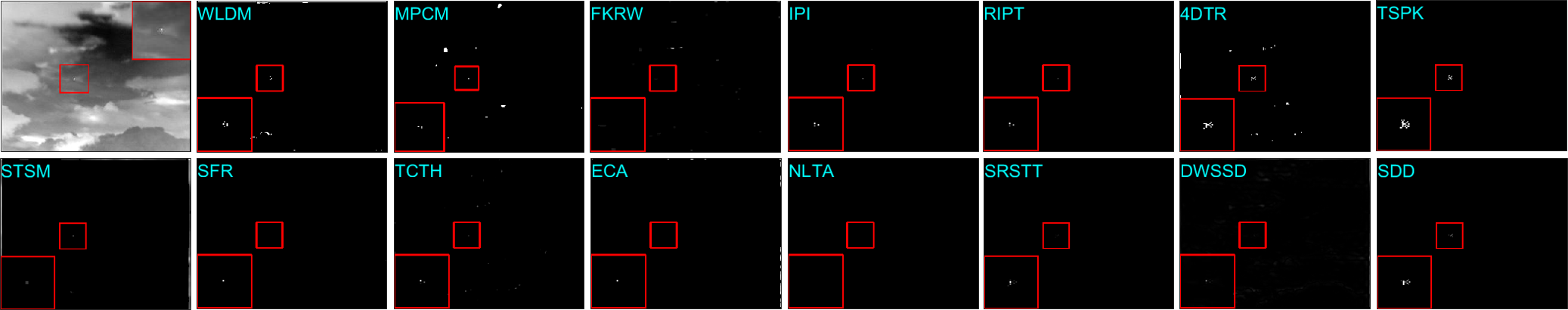} \\
  \caption{Detection results of ten comparison methods under scene $j$. The target area is zoomed in the lower left corner for better observation.}
\label{fig:com_c}
\end{figure*}

\begin{figure*}[htb!]
  \centering
  \includegraphics[width=16cm]{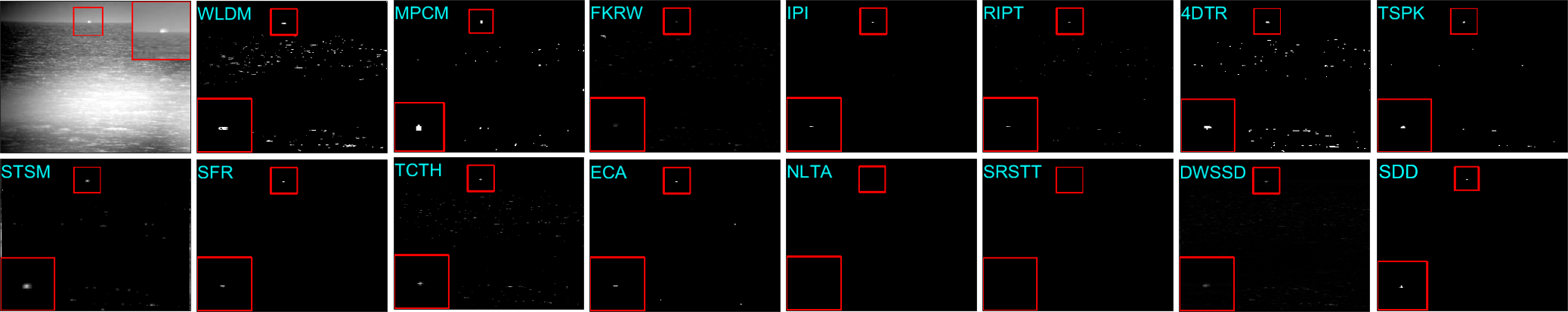} \\
  \caption{Detection results of ten comparison methods under scene $l$. The target area is zoomed in the lower left corner for better observation.}
\label{fig:com_d}
\end{figure*}

\begin{figure*}[htb!]
  \centering
  \includegraphics[width=16cm]{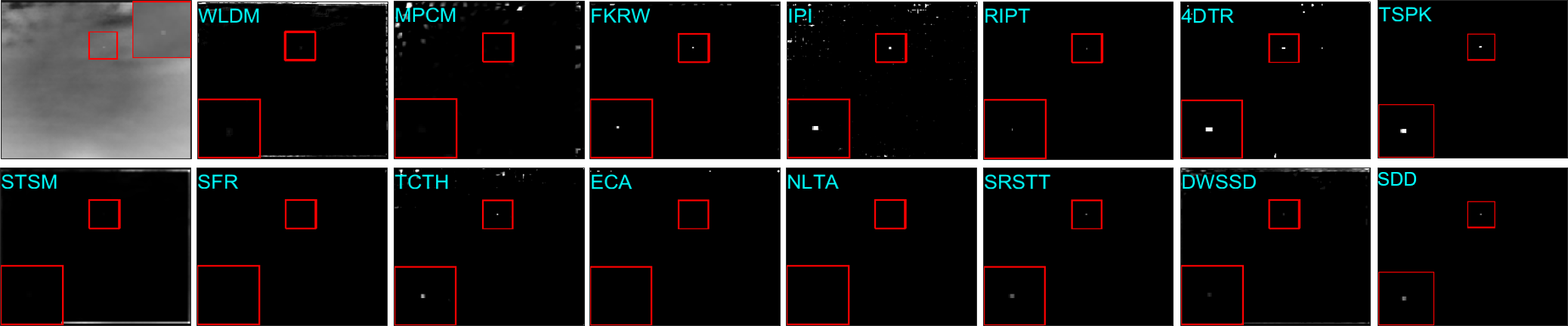} \\
  \caption{Detection results of ten comparison methods under scene $m$. The target area is zoomed in the lower left corner for better observation.}
\label{fig:com_e}
\end{figure*}

\begin{figure*}[htb!]
  \centering
  \includegraphics[width=16cm]{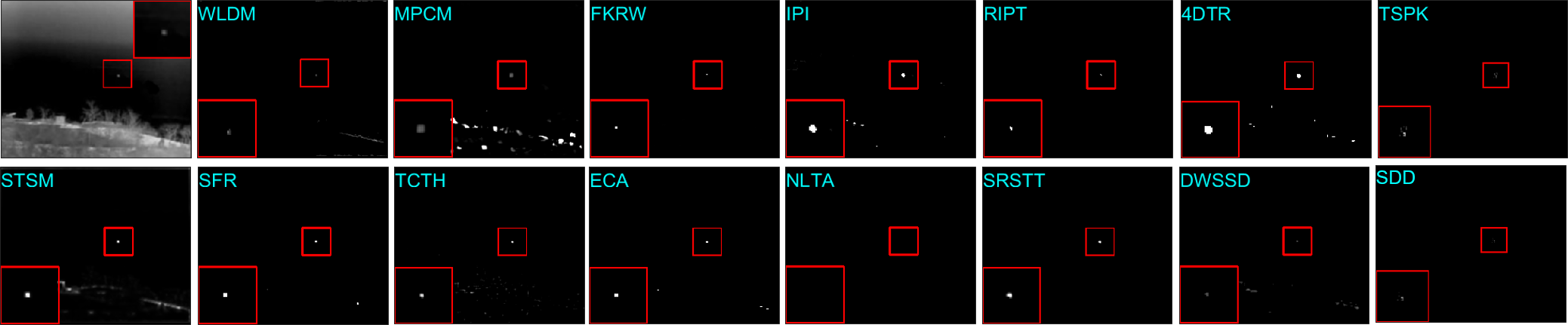} \\
  \caption{Detection results of ten comparison methods under scene $r$. The target area is zoomed in the lower left corner for better observation.}
\label{fig:com_f}
\end{figure*}

\begin{table*}[htb!]
  \tabcolsep=0.1cm
  \scriptsize
  \centering
  \caption{The quantitative indicators of all compared methods with respect to average $BSF$, $G_{SCR}$ and $CG$ per frame under six tested scenes.}
  \begin{tabular}{ccccccccccccccccccc}
  \toprule
  \multirow{2}{*}{Methods} & \multicolumn{3}{c}{Scene $a$} & \multicolumn{3}{c}{Scene $b$} & \multicolumn{3}{c}{Scene $c$} & \multicolumn{3}{c}{Scene $d$} & \multicolumn{3}{c}{Scene $e$} & \multicolumn{3}{c}{Scene $f$}\\
  \cmidrule(r){2-4}
  \cmidrule(r){5-7}
  \cmidrule(r){8-10}
  \cmidrule(r){11-13}
  \cmidrule(r){14-16}
  \cmidrule(r){17-19} 
      & $ {BSF}$ & $ G_{SCR}$ & $CG$
      & $ {BSF}$ & $ G_{SCR}$ & $CG$
      & $ {BSF}$ & $ G_{SCR}$ & $CG$
      & $ {BSF}$ & $ G_{SCR}$ & $CG$
      & $ {BSF}$ & $ G_{SCR}$ & $CG$
      & $ {BSF}$ & $ G_{SCR}$ & $CG$ \\
    \midrule
   WLDM\cite{deng2016small} & 4.66	&2.96	&1.42	&6.16	&8.55	&1.31	&6.14	&2.29	&1.25	&6.71	&3.82	&1.93	&5.03	&3.22	&1.15	&7.97	&13.38	&1.34 \\
   MPCM\cite{Wei2016Multiscale} & 9.00	&7.55	&2.81	&15.62	&11.94	&5.64	&6.29	&5.91	&2.85	&2.49	&1.82	&2.24	&14.58	&6.53	&1.13	&16.18	&8.05	& 2.41 \\
   FKRW\cite{Qin2019Infrared} &  5.55	&4.79	&0.96	&6.89	&1.77	&0.69	&7.12	&3.46	&0.88	&9.29	&2.37	&1.66	&3.49	&1.71	&0.98	&7.86	&3.44	& 0.74  \\
   IPI\cite{Cao2013Infrared} & 111.92	&39.38	&11.29	&72.32	&24.01	&12.07	&95.89	&89.31	&16.93	&121.56	&55.89	&18.19	&49.79	&16.74	&12.18	&73.65	&\textbf{76.54}	&12.9 \\
   RIPT\cite{Dai2017Reweighted} & 121.92	&\underline{76.48}	&\underline{29.16}	&82.39	&28.69	&22.07	&\underline{102.88}	&\underline{91.37}	&\underline{26.49}	&102.43	&\underline{65.47}	&23.46	&58.96	&26.62	&16.28	&79.61	&\underline{73.66}	&22.16 \\
   STSM\cite{Li2016A} &5.667	&11.30	&7.668	&6.965	&17.13	&1.334	&8.640	&7.384	&3.360	&13.22	&13.95	&1.944	&5.117	&9.508	&1.239	&7.985	&20.71	&1.34 \\
   SFR\cite{Pang2022STTM} & 39.222	&32.185	&19.357	&41.289	&28.649	&10.48	&45.127	&40.624	&10.189	&47.596	&39.146	&16.22	&26.174	&18.213	&\textbf{18.956}	&42.897	&34.212	&22.020    \\
   TCTH\cite{Zhu2020Infrared} & 71.568	&31.013	&15.143	&72.329	&36.103	&12.078	&96.040	&74.027	&16.820	&82.149	&62.300	&18.195	&49.79	&19.405	&12.176	&73.656	&70.063	&12.96 \\
   ECA\cite{Zhang2020Edge} & 37.937	&15.884	&16.286	&71.817	&43.005	&12.213	&23.408	&48.658	&14.584	& \textbf{145.96}	&58.474	&\underline{37.955}	&37.517	&17.665	&11.342	&69.21	&32.74	&22.56   \\
  NLTA\cite{Liu2020Small} &	60.712	&32.575	&14.967	&56.513	&27.291	&\underline{24.960}	&41.311	&29.003	&22.110	&59.540	&27.58	&24.19	&27.419	&9.953	&11.565	&29.447	&31.945	&\underline{24.226}   \\
  % \rowcolor{yellow}
  TSPK\cite{zhou2023infrared} & \bf{183.441}	&76.191	&12.867	&\textbf{106.866} &\underline{46.151} &14.282 &87.332	&55.796	&12.011	&132.720	&36.601	&14.182	&\textbf{126.091}	&\underline{49.429} 	&12.211	&\underline{80.068}	&41.458	&14.126   \\
  % \rowcolor{yellow}
  4DTR\cite{Wu2023Infrared} &	3.909	&1.278	&4.468	&13.372	&14.453	&5.820	&6.611	&1.617	&4.225	&6.292	&1.311	&6.912	&10.765	&1.263	&3.626	&14.446	&3.232	&4.116   \\
  % \rowcolor{yellow}
  SRSTT\cite{Li2023Sparse} &	80.912	&54.081	&9.823	&66.243	&24.355	&10.260	&62.214	&26.994	&12.102	&68.958	&28.979	&11.221	&64.352	&27.541	&10.612	&68.312	&36.975	&14.121   \\
  % \rowcolor{yellow}
  DWSSD\cite{Dan2024Dynamic} &	9.347	&1.244	&3.016	&8.999	&1.989	&2.936	&6.3725	&4.110	&2.949	&10.753	&1.799	&2.473	&4.576	&2.283	&2.251	&7.122	&3.460	&1.121   \\
  SDD &	\underline{123.733}	& \textbf{84.66}	& \textbf{35.092}	& \underline{90.222}	& \textbf{49.428}	& \textbf{26.917}	& \textbf{106.679}	& \textbf{97.322}	& \textbf{34.781}	& \underline{136.174}	& \textbf{70.430}	& \textbf{39.026}	& \underline{79.391}	& \textbf{49.946}	& \underline{14.827}	& \textbf{88.234}	& 68.673	& \textbf{42.436}   \\

    \bottomrule
    \end{tabular}
    \label{tbl:commetr}
  \end{table*}
  
We utilize six representative scenes $a$, $b$, $j$, $m$, $l$, $r$ to compare the visual impact of the proposed model with 14 state-of-the-art competitors. 
% We utilize six representative scenes a-f to compare the visual impact of the proposed model with 14 state-of-the-art competitors. 
These sequences not only include sea-air, land-sea, and deep-sea scenarios but also exhibit significant variations in target characteristics and background complexity. Fig. \ref{fig:com_a}-\ref{fig:com_f} display the detection results for all tested methods across the six scenes. 
% To facilitate a more effective visual comparison, the areas surrounding small targets are magnified and positioned in the lower-left corner of the result images.
Among the competitors, WLDM, MPCM, FKRW and STSM belong to the saliency-based methods, which emphasize extracting small targets by eliminating background clutter. As evident from the results obtained by WLDM and MPCM in Fig. \ref{fig:com_a} - \ref{fig:com_d}, \ref{fig:com_f}, small targets are accentuated, but structures resembling target saliency are also preserved, leading to high false alarms, and in Fig. \ref{fig:com_e}, low-contrast targets are not only unenhanced but also lost, leading to false detections. FKRW and STSM exhibit superior clutter suppression compared to WLDM and MPCM, with the exception of the detection results featuring either missing targets in Fig. \ref{fig:com_a}, \ref{fig:com_c} and \ref{fig:com_e} or residual backgrounds in Fig. \ref{fig:com_b} and \ref{fig:com_f}.
The remaining methods are founded on the low-rank and sparse decomposition framework. From their results, it is apparent that although fewer background residues are present in the target images compared to saliency-based methods, performance disparities still persist.

As illustrated in Fig. \ref{fig:com_a} and \ref{fig:com_b}, IPI, RIPT, 4DTR, TSPK, NLTA and SRSTT exhibit no background residues, while SFR, TCTH,  ECA and DWSSD contain some residues.
The detection results in Fig. \ref{fig:com_c} reveal that NLTA and 4DTR lead to the omission of real targets and confusion with false ones, but the rest of these methods display relatively stable performance. 
% The primary cause is the various ground interferences in scene $c$ that partially obscure small targets.
% Examining Fig. \ref{fig:com_d} and \ref{fig:com_f}, it is clear that ECA and NLTA fail to detect small targets, potentially resulting in a low detection probability. 
Examining Fig. \ref{fig:com_d} and \ref{fig:com_f}, it is clear that the performance of these methods in scene $r$ is better than in scene $l$, mainly due to the interference of bright sea surface reflections in scene $l$.
% RIPT, SFR, and TCTH successfully identify small targets, but some non-target components remain.
% In Fig.\ref{fig:com_e}, bright background interferences are not entirely eliminated in the target images of IPI, RIPT, SFR, 4DTR, TSPK, ECA and SRSTT.
In Fig.\ref{fig:com_e}, dim targets are easily lost in RIPT, ECA, NLTA and DWSSD.
In contrast, our proposed model achieves a significant performance improvement in eradicating background disturbances while accurately extracting small targets across different scenes, which is attributed to the integration of the spatial-temporal difference directionality prior and the saliency enhancement factor.

\subsubsection{Quantitative Comparison with Other Competitors}
In addition to visual illustrations, we offer objective evaluation indicators to thoroughly validate the proposed method. Three evaluation metrics are employed, namely the averages of $BSF$, $G_{SCR}$, and $CG$ for each frame. The average results for all tested methods across six different scenes, with respect to the three indices, are presented in Table \ref{tbl:commetr}. 
The highest values are marked in bold, and the second highest values are underlined.
The results clearly demonstrate that the proposed method surpasses its counterparts.
In scene $a, b, d, e$, the BSF of our proposed method is slightly lower than the highest value, ranking second.
Furthermore, the proposed method's $G_{SCR}$ value is best in scenes $a$-$e$, but lower than RIPT and TCTH in scene $f$.
Specifically, our approach attains the highest values in scenes $a$-$d$, $f$ for $CG$, but slightly lower than SFR in scene $e$.
This result indicates the model's efficacy in preserving the brightness of small targets during extraction.

\begin{figure*}[htb!]
        \centering
        \subfigure[]{
           \includegraphics[width=5.0cm]{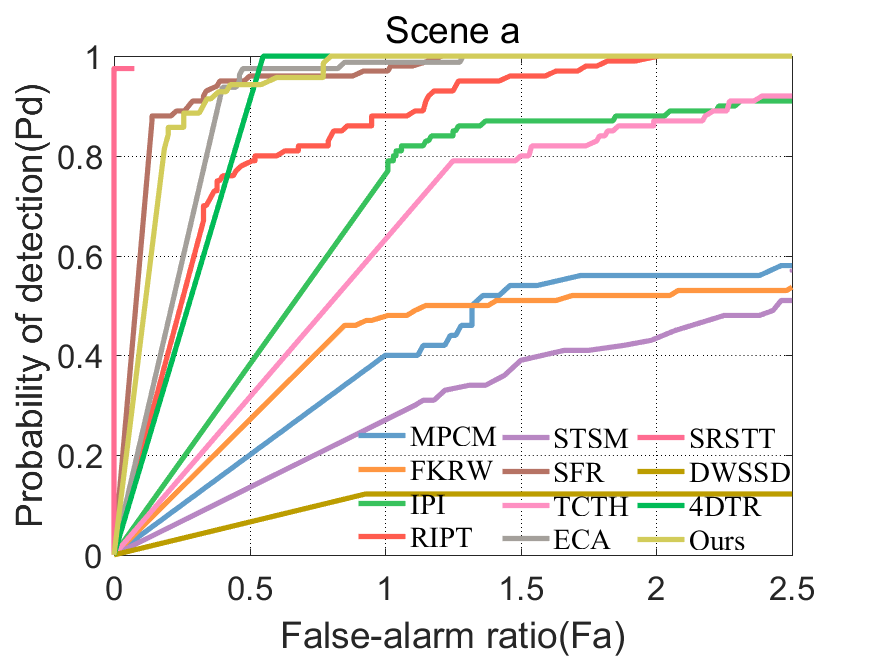}
    }%\hspace{-6mm}
       \subfigure[]{
            \includegraphics[width=5.0cm]{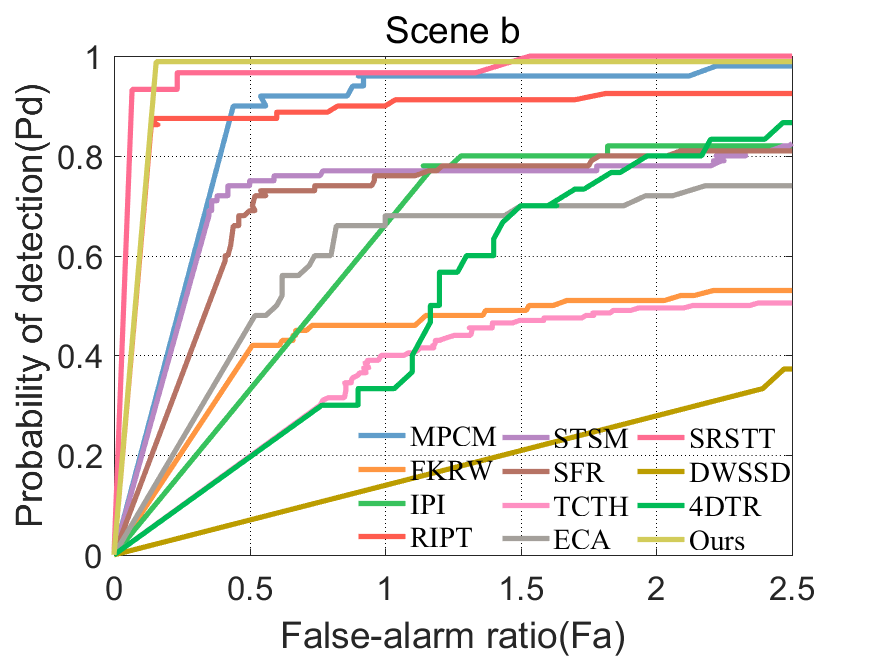}
    }
        \subfigure[]{
            \includegraphics[width=5.0cm]{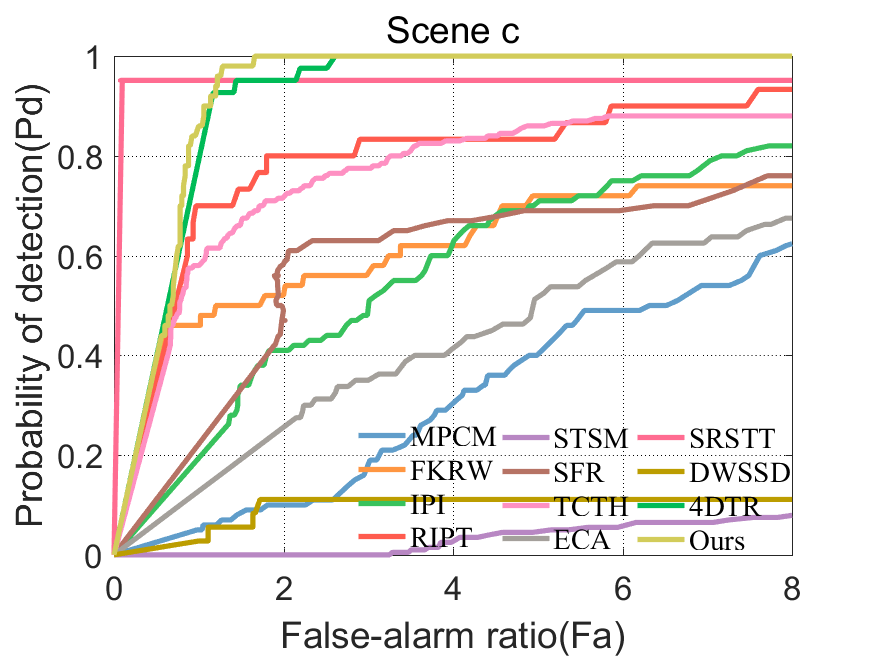}
    }
    
    \vskip -5pt
          \subfigure[]{
            \includegraphics[width=5.0cm]{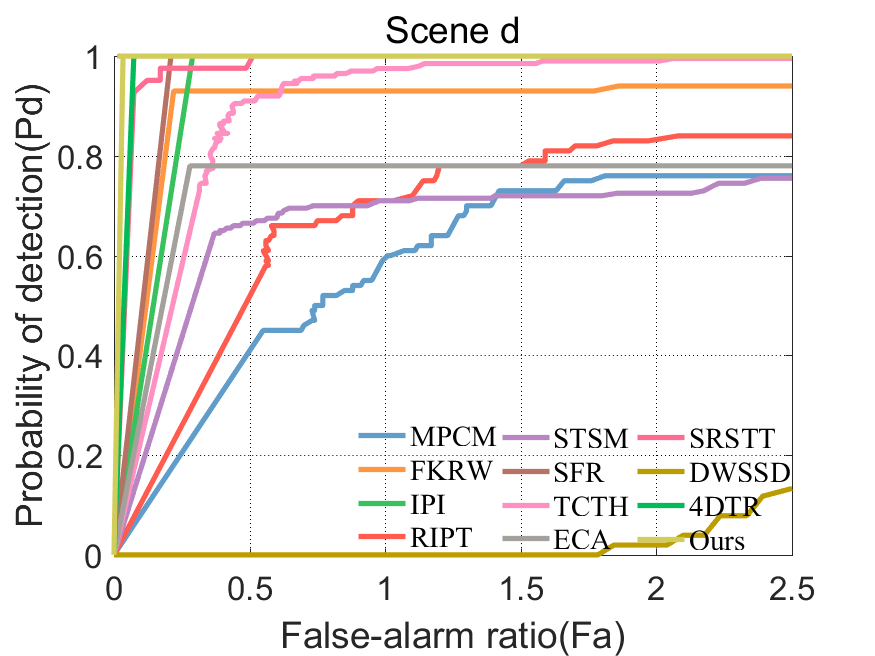}
    }%\hspace{-6mm}
          \subfigure[]{
            \includegraphics[width=5.0cm]{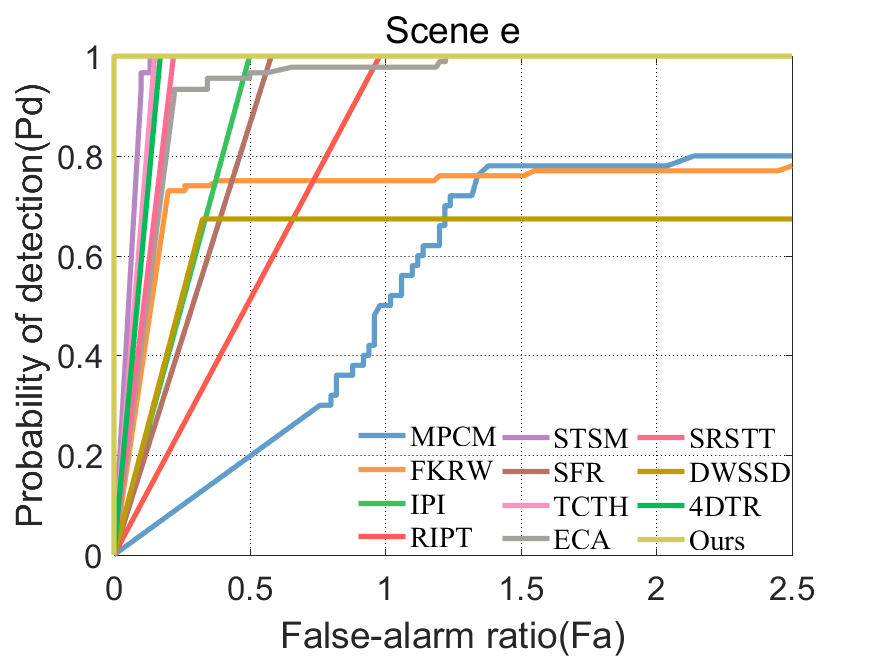}
    }%\hspace{-6mm}
        \subfigure[]{
            \includegraphics[width=5.0cm]{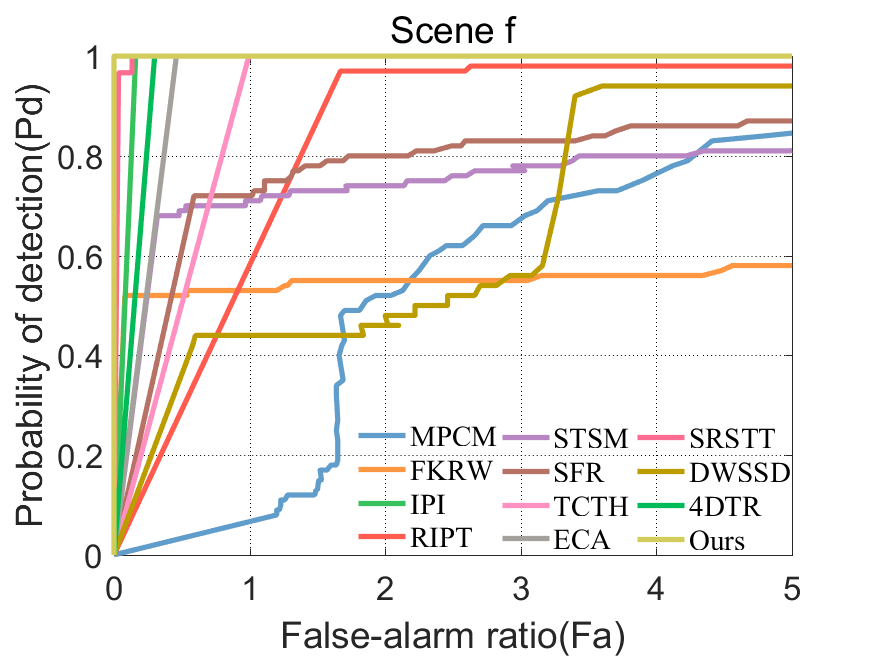}
    }
        \vskip -5pt
        \caption{ ROC curves of dynamic change of $P_d$ with $F_a$ compared with eight competitors on scene $a$-$f$.}
        \label{fig:comroc}
       
      \end{figure*}

  The three aforementioned indicators primarily assess the local background suppression performance of the evaluated methods. In this section, we present the global ROC curves illustrating the dynamic changes in detection probability with respect to the false alarm rate for all tested methods, as depicted in Fig. \ref{fig:comroc}. A steeper curve ascent indicates superior performance, signifying that higher detection performance can be attained with a reduced number of false alarms.
Among all methods, our proposed approach achieves the highest $P_d$ at lower $F_a$ values. It is important to note that other methods demonstrate inconsistent detection performance across different scenes. 
 % MPCM attains a satisfactory $P_d$ in scene $b$, but its performance in other scenes is less than optimal. 
For example, in scenes $c$ and $d$, RIPT's $P_d$ is 0.82 at an $F_a$ of 2.5, which is considerably lower than its performance in other scenes. While SFR achieves a $P_d$ of 1 in scenes $a$, $d$, and $e$, it does not maintain this level in the remaining tested scenes. 
SRSTT and 4DTR show promising results in scenes $d$-$f$, but its performance declines in scenes $a$-$c$.
Based on this analysis, we can conclude that our proposed method effectively balances $P_d$ and $F_a$ across different scenes, thereby confirming its stability.

\subsubsection{Computational Complexity and Time Consumption}
In this subsection, we examine the computational complexity of the PAM-based solution applied to an infrared sequence cube ${{\cal Y}^{{n_1} \times {n_2} \times {n_3}}}$, as outlined in Algorithm \ref{alg:modelsolutionPAM}.
The primary computational expense at each loop iteration arises from updating $A$, $\cal B$, ${{\cal Z}_1}$, ${{\cal Z}_2}$, ${{\cal P}_1}$, ${{\cal P}_2}$, $\cal T$, and the number of inner iterations $l$.
The cost of updating $A$ is ${\cal O}\left( {r{n_1}{n_2}{n_3} + {r^2}{n_3} + r{n_3}\log \left( {{n_3}} \right)} \right)$, attributed to the involvement of SVD, 1-D FFT, and multiple matrix multiplications.
Updating $\cal B$ incurs a computational expense of ${\cal O}\left( {\left( {r + {n_1} + {n_2}} \right){n_1}{n_2}{n_3} + {r^2}{n_3} + r{n_1}{n_2}\log \left( {{n_1}{n_2}} \right)} \right)$, resulting from SVD, 2-D FFT, and several tensor-matrix product operations.
For updates to ${{\cal Z}_1}$, ${{\cal Z}_2}$, and $\cal T$, simple threshold operators are employed, yielding a cost of ${\cal O}\left( {{n_1}{n_2}{n_3}} \right)$.
The computational complexity of updating ${{\cal P}_1}$ and ${{\cal P}_2}$ is ${\cal O}\left( {n_1^2{n_2}r} \right)$. Ultimately, the computational complexity of the proposed optimization is ${\cal O}\left( {r{n_3}\log \left( {{n_3}} \right) + l\left( {\left( {r + {n_1} + {n_2}} \right){n_1}{n_2}{n_3} +r{n_1}{n_2}\log \left( {{n_1}{n_2}} \right)} \right)} \right)$.

To assess the detection efficiency of our proposed algorithm, we compare the average time consumption per frame for all tested scenes with 14 competing methods, as detailed in Table \ref{tbl:timesum}. 
It can be observed that saliency-based approaches, such as WLDM, MPCM, FKRW, and STSM, exhibit faster performance than the rest of the low-rank decomposition-based methods.
% It can be observed that saliency-based approaches, such as WLDM, MPCM, FKRW, and STSM, exhibit faster performance than low-rank decomposition-based methods, including IPI, RIPT, SFR, TCTH, ECA, TSPK, 4DTR and SRSTT.
% This is primarily attributed to the fact that saliency-based methods involve only simple linear operations, devoid of time-consuming singular value decomposition. 
However, these methods demonstrate less stability across different scenes compared to their low-rank decomposition-based counterparts.
Among the low-rank decomposition-based methods, our algorithm exhibits competitive detection efficiency. Although the proposed method is marginally slower than RIPT, NLTA, TSPK, 4DTR and DWSSD. 
The proposed method improves detection efficiency by at least 50\% compared to similar algorithms SFR, TCTH, ECA, and SRSTT.
In summary, although our method does not achieve the fastest detection efficiency, it demonstrates superior target detection and background suppression capabilities.

\begin{table}[htb!]
 \tabcolsep=0.1cm
\centering
\caption{Average running time (/s) on per frame of all compared methods under different tested scenes.}
 \setlength{\tabcolsep}{1.5mm}{
\begin{tabular}{ccccccc}
\toprule
Methods & Scene $a$ & Scene $b$ & Scene $c$ & Scene $d$ & Scene $e$ & Scene $f$\\
\hline
\multicolumn{7}{l}{\textit{Saliency-based Method}}  \\ 
\hline
 WLDM\cite{deng2016small} & 1.5740	&1.5322	&1.6121	&1.5985	&1.5735	&1.5624 \\
 MPCM\cite{Wei2016Multiscale}  &0.1533	&0.1581	&0.1566	&0.1676	&0.1471	&0.1265   \\
 FKRW\cite{Qin2019Infrared}  & 0.9349	&0.9794	&1.056	&0.9446	&0.8350	&0.8754   \\
 STSM\cite{Li2016A}  &  0.3409	&0.3527	&0.3141	&0.3050	&0.3053	&0.304   \\
 \hline
 \multicolumn{7}{l}{\textit{Low-rank Sparse Decomposition Method}}  \\ 
 \hline
 IPI\cite{Cao2013Infrared}   & 8.634	&8.646	&9.449	&8.893	&8.101	&8.423   \\
 RIPT\cite{Dai2017Reweighted}  &2.511	&2.117	&2.485	&1.815	&1.943	&2.155   \\
 SFR\cite{Pang2022STTM}  &  11.001	&12.432	&12.013	&10.561	&12.141	&11.205   \\
 TCTH\cite{Zhu2020Infrared}  & 9.751	&9.864	&9.7519&	9.573	&9.704	&9.67   \\
 ECA\cite{Zhang2020Edge} &9.7297	&9.7422	&9.7673	&9.753	&9.6521	&10.636 \\
 NLTA\cite{liu2022Nonconvex} & 3.0719	&3.3283	&3.3271	&3.3699	&3.0874	&3.0773 \\
 % \rowcolor{yellow}
  TSPK\cite{zhou2023infrared}  & 1.8916	&1.9233	&1.7931	&1.6670	&1.5580	&1.7500 \\
 % \rowcolor{yellow}
  4DTR\cite{Wu2023Infrared} & 2.4384	&2.4796	& 2.5130 &2.5820	&2.4374	&2.4893 \\
% % \rowcolor{yellow}
  SRSTT\cite{Li2023Sparse} & 14.3105	&13.3884	&14.7158	&13.3774	&13.1602	&12.8718 \\
% \rowcolor{yellow}
  DWSSD\cite{Dan2024Dynamic} & 0.3616	&0.3373	&0.3756	&0.3204	&0.3008	&0.3018 \\
 SDD  & \textbf{4.0272} & \textbf {4.1055}  & \textbf{4.0528} & \textbf{4.1315}  & \textbf{4.0279}  & \textbf{4.1326}  \\
  \bottomrule
  \end{tabular}}
  \label{tbl:timesum}
\end{table}

%% file: contents/conclusion.tex
% !TEX root = ../main.tex

\section{Conclusion} \label{sec:conclusion}
% In this study, to effectively distinguish targets from background interference, we first present a new potential prior: sparse differential directionality (SDD).
% It is integrated by the strategy of the mixed sparse constraint on the directional difference images of spatial factors into tensor decomposition.
% And the temporal difference factor is regularized by a continuity constraint to exploit inter-frame correlations.
In this study, we introduce a new potential prior, SDD, to distinguish targets from background interference. We integrate it into tensor decomposition using a mixed sparse constraint on the spatial factors' directional difference images. The temporal difference factor is regularized by a continuity constraint to leverage inter-frame correlations.
% Furthermore, a saliency enhancement coherence map is devised, and it is integrated into the decomposition process using a reweighting strategy to enhance the saliency of the target and weaken the saliency coherence of background interference.
% The proposed model is then efficiently solved using a Proximal Alternating Minimization (PAM)-based algorithm, ensuring numerically guaranteed convergence. 
Additionally, we devise a saliency enhancement coherence map, which we integrate using a reweighting strategy to enhance target saliency and reduce background interference. The proposed model is solved using a PAM-based algorithm for guaranteed numerical convergence.
% Through extensive experimentation, we compare our method with ten state-of-the-art approaches, demonstrating its superior performance in small target detection and background interference suppression. 
% \hl{Our method outperforms ten state-of-the-art approaches in small target detection and background interference suppression, as demonstrated through extensive experimentation. In particular, our model achieves full detection in testing scenarios and surpasses the second-best algorithm by 10.6\% in $G_{SCR}$ and 75.1\% in $CG$ under a false alarm rate of 1.}
Extensive experiments compare our method with ten state-of-the-art approaches, demonstrating its superior performance in small target detection and background interference suppression. Remarkably, the proposed model consistently detects all targets in testing scenarios and surpasses other algorithms when allowing a false alarm rate of 1, outperforming the second-best algorithm by 10.6\% in $G_{SCR}$ and 75.1\% in $CG$.
% \hl{Notably, the proposed model achieves a 100\% detection rate in testing scenarios and outperforms other algorithms under the false alarm rate of 1. Specifically, it surpasses the second-best algorithm in $G_{SCR}$ by 10.6\% and in $CG$ by 75.1\%.}
% As a future objective, we aim to incorporate conventional neural networks within the tensor decomposition model, thereby enabling the automatic learning of a more rational regularization in lieu of handcrafted priori constraints.

\section*{Acknowledgments}
The authors would like to thank the editor and anonymous reviewers for their help comments and suggestions.

%% file: contents/biography.tex
% !TEX root = ../main.tex

% Xiang Li
% \begin{IEEEbiography}[{\includegraphics[width=1.1in,height=1.35in,clip,keepaspectratio]{./figs/biography/Xiang_Li}}]{Xiang Li} is an Associate Professor in College of Computer Science, Nankai University. He received the PhD degree from the Department of Computer Science and Technology, Nanjing University of Science and Technology (NJUST) in 2020. There, he started the postdoctoral career in NJUST as a candidate for the 2020 Postdoctoral Innovative Talent Program. In 2016, he spent 8 months as a research intern in Microsoft Research Asia, supervised by Prof. Tao Qin and Prof. Tie-Yan Liu. He was a visiting scholar at Momenta, mainly focusing on monocular perception algorithm. His recent works are mainly on: neural architecture design, CNN/Transformer, object detection/recognition, unsupervised learning, and knowledge distillation. He has published 20+ papers in top journals and conferences such as TPAMI, CVPR, NeurIPS, etc.
% \end{IEEEbiography}

% Fei Zhou
\begin{IEEEbiography}[{\includegraphics[width=1in,height=1.25in,clip,keepaspectratio]{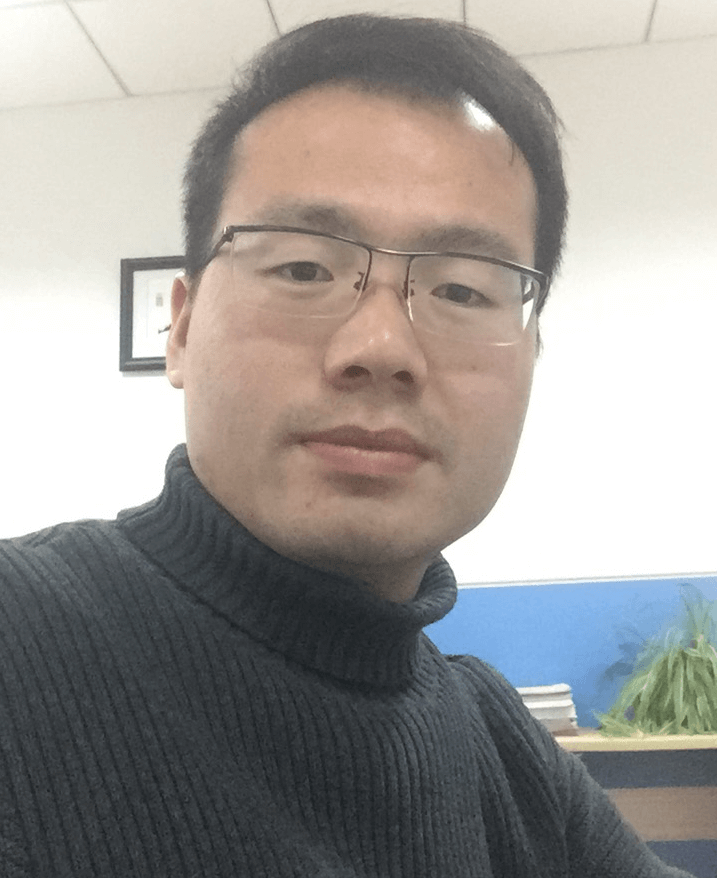}}]{Fei Zhou}
  received his M.S. degree in electronic engineering from Xinjiang University in 2017, and received Ph.D. degree from College of Electronic and Information Engineering, Nanjing University of Aeronautics and Astronautics. He is currently a lecturer with College of Information Science and Engineering, Henan University of Technology, Zhengzhou, China. His main research interests are signal processing, target detection and image processing. He has published 10+ papers in journals and conferences, such as TGRS, TAES, WACV, etc.
\end{IEEEbiography}

% Maixia Fu
\begin{IEEEbiography}[{\includegraphics[width=1in,height=1.25in,clip,keepaspectratio]{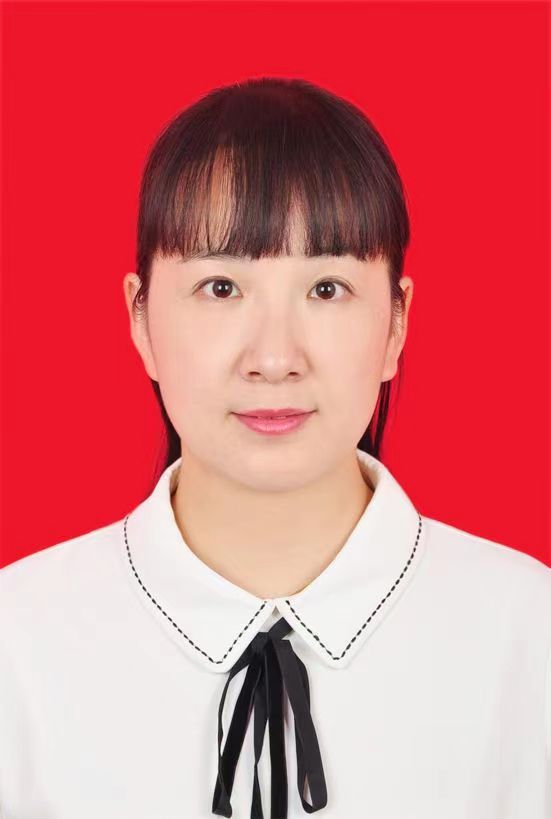}}]{Maixia Fu}
received the B.S. degree in electronic
science and technology from Information Engineering University, Zhengzhou, China, in 2003, the M.S. degree in information and communication engineering from Zhengzhou University, Zhengzhou, China, in 2009, and the Ph.D. degree in optical engineering from Capital Normal University, Beijing, China, in 2017.
She is currently an Associate Professor and
master’s Supervisor with the College of Information Science and Engineering, Henan University of Technology, Zhengzhou, China. Her research interests include signal
processing, spectrum analysis, optoelectronic devices designing, and remote sensing
\end{IEEEbiography}

% Yulei Qian
\begin{IEEEbiography}[{\includegraphics[width=1.1in,height=1.35in,clip,keepaspectratio]{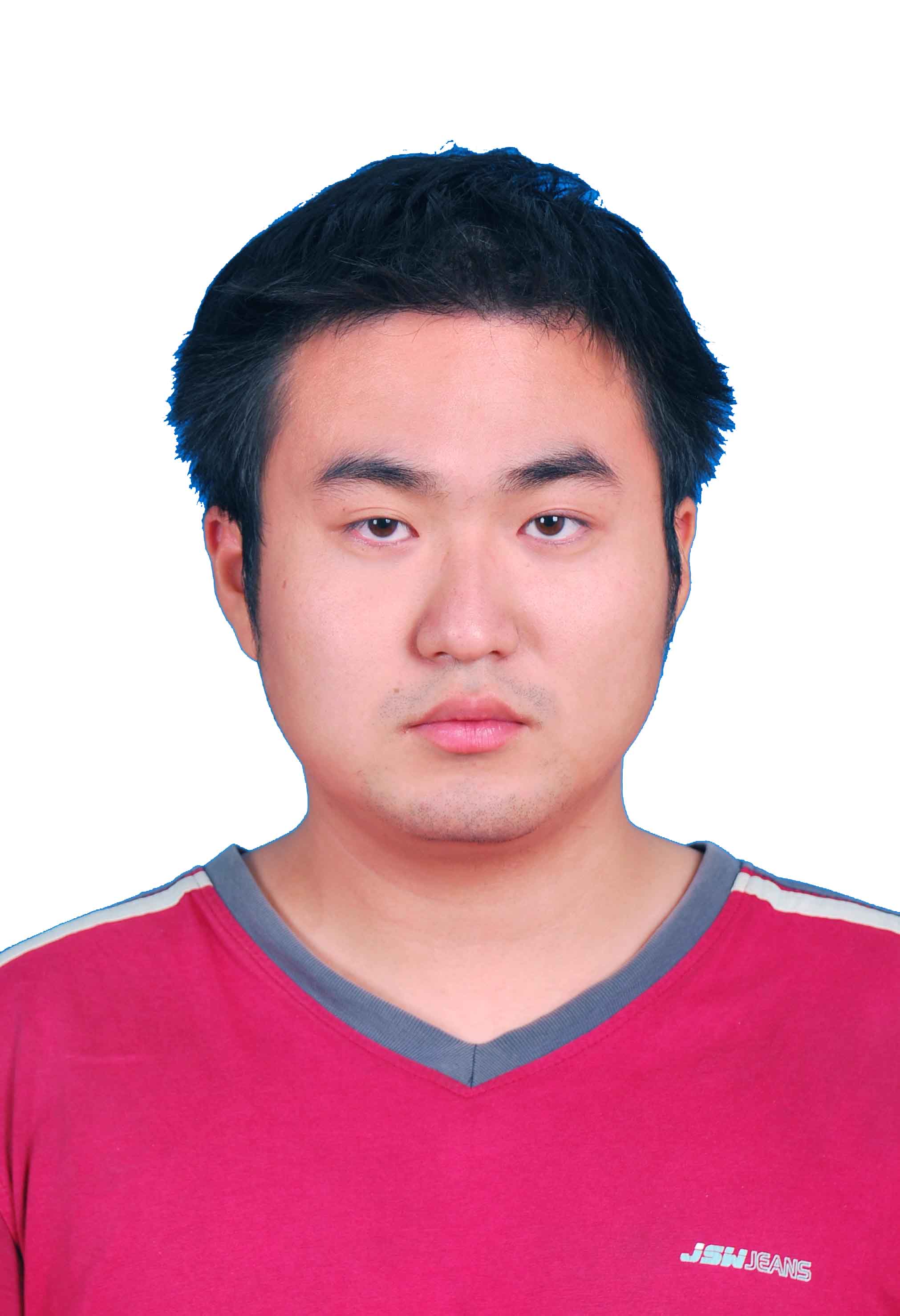}}]{Yulei Qian} received his PhD degree from College of Electronic and Information Engineering, Nanjing University of Aeronautics and Astronautics in 2020. He is now an engineer in Nanjing Marine Radar Institute. His main research interests include target detection, radar imaging, satellite synthetic aperture radar, sparse recovery, deconvolution and radar waveform design. He has published 10+ paper in remote sensing journals and conferences.
\end{IEEEbiography}

\begin{IEEEbiography}[{
\includegraphics[width=1.45in,height=1.3in,clip,keepaspectratio]{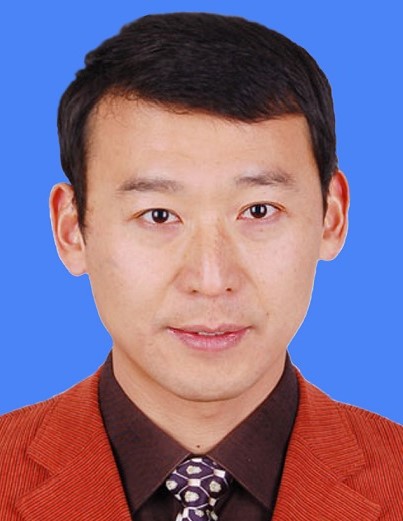}}]{Jian Yang} received the PhD degree from Nanjing University of Science and Technology (NJUST) in 2002, majoring in pattern recognition and intelligence systems. From 2003 to 2007, he was a Postdoctoral Fellow at the University of Zaragoza, Hong Kong Polytechnic University and New Jersey Institute of Technology, respectively. From 2007 to present, he is a professor in the School of Computer Science and Technology of NJUST. Currently, he is also a visiting distinguished professor in the College of Computer Science of Nankai University. He is the author of more than 300 scientific papers in pattern recognition and computer vision. His papers have been cited over 40000 times in the Scholar Google. His research interests include pattern recognition and computer vision. Currently, he is/was an associate editor of Pattern Recognition, Pattern Recognition Letters, IEEE Trans. Neural Networks and Learning Systems, and Neurocomputing. He is a Fellow of IAPR. 
\end{IEEEbiography}

% Yimian Dai
\begin{IEEEbiography}[{\includegraphics[width=1in,height=1.25in,clip,keepaspectratio]{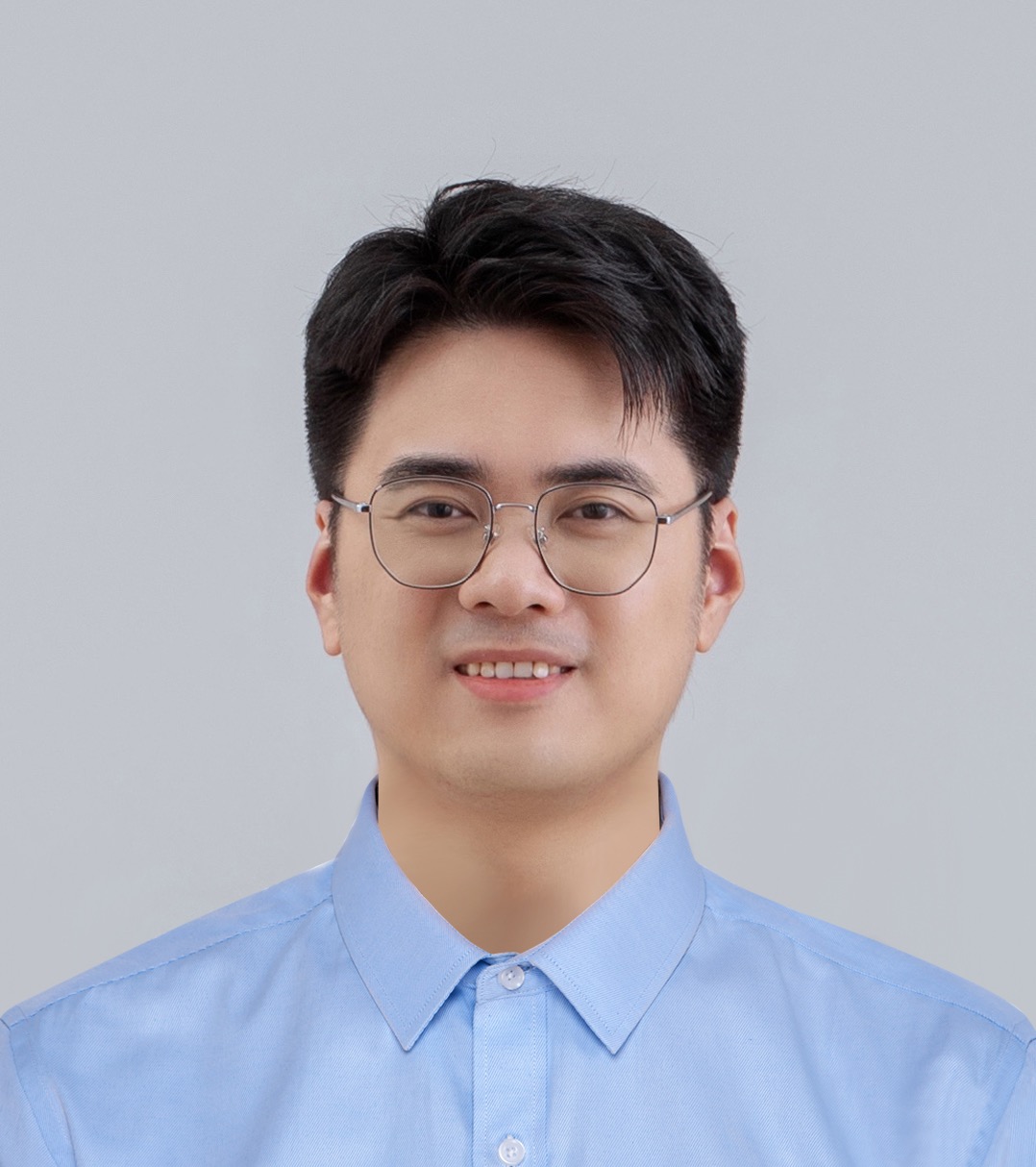}}]{Yimian Dai} received the Ph.D. degree in signal and information processing from Nanjing University of Aeronautics and Astronautics, Nanjing, China, in 2021, during which he honed his research skills as a visiting Ph.D. student at the University of Copenhagen and the University of Arizona between March 2018 and October 2020. Since 2021, he has been with the School of Computer Science and Engineering, Nanjing University of Science and Technology (NJUST), Nanjing, where he is currently an Assistant Researcher. His research interests include remote sensing, computer vision, and deep learning, with a focus on developing algorithms for object detection, data assimilation, and computational imaging to tackle real-world challenges. He has authored more than 20 peer-reviewed journal and conference papers such as IJCV, TGRS, TAES, etc.
\end{IEEEbiography}